%% file: paper.tex
\documentclass[10pt,twocolumn,twoside]{IEEEtran}

\usepackage{graphicx}
\usepackage{amsmath}
\usepackage{subfigure}
\usepackage{indentfirst}
\usepackage{url}
\usepackage{booktabs,multirow}
\usepackage{stfloats}
\usepackage{lettrine}
\usepackage{mathabx}
\usepackage{color}

\renewcommand\paragraph[1]{\noindent \textbf{#1}}

\def\etal{\emph{et al}.~}
\def\ie{\emph{i.e.}~}
\def\eg{\emph{e.g.}~}
\def\l2{$\ell_2$}

\def\sssp{\hspace{2pt}}
\def\ssp{\hspace{3pt}}

\begin{document}

\title{A comparison of dense region detectors for image search and fine-grained classification}

\author{Ahmet~Iscen$^{\star}$, Giorgos~Tolias, Philippe-Henri~Gosselin and~Herv\'e~J\'egou
\thanks{INRIA, Campus Universitaire de Beaulieu 35042 Rennes Cedex FRANCE}   
\thanks{1057-7149 (c) 2015 IEEE. Personal use is permitted, but republication/redistribution requires IEEE permission.} }

\maketitle

\begin{abstract}
We consider a pipeline for image classification or search based on coding approaches like Bag of Words or Fisher vectors. In this context, the most common approach is to extract the image patches regularly in a dense manner on several scales. This paper proposes and evaluates alternative choices to extract patches densely. Beyond simple strategies derived from regular interest region detectors, we propose approaches based on super-pixels, edges, and a bank of Zernike filters used as detectors. 

The different approaches are evaluated on recent image retrieval and fine-grain classification benchmarks. Our results show that the regular dense detector is outperformed by other methods in most situations, leading us to improve the state of the art in comparable setups on standard retrieval and fined-grain benchmarks. 
As a byproduct of our study, we show that existing methods for blob and super-pixel extraction  achieve high accuracy if the patches are extracted along the edges and not around the detected regions.  
\end{abstract}


\IEEEpeerreviewmaketitle

\input{intro.tex}
\input{related.tex}

\input{background.tex}

\input{methods.tex}
\input{experiments.tex}

\section{Conclusions}

We investigated dense keypoint detection solutions that lie in between sparse interest points and dense sampling on a uniform grid. We propose to modify existing interest point detectors by relaxing the cornerness criterion and the local maxima selection.
We introduce a new detection method using Zernike filters, which provides dense, yet localized image patches.
We also show that sampling patches on the borders of a region of interest performs better than the standard choice of fitting an ellipse and describing it by a single descriptor.
Finally, we propose to detect dense patches by using the \l2-norm of the descriptor instead of low-level pixel information. To our knowledge, this is the first detection strategy which focuses on descriptors, and the results seem very promising.

Interestingly, solutions employing patches of multiple fixed scales perform better than patches detected as local maxima in the scale space. Albeit not scale invariant, apparently this option provides enough scale tolerance for the tasks of image retrieval and fine-grained classification.

Compared with the existing studies, Zernike patches encoded with a standard technique, such as Fisher vectors, appear to outperform state of the art approaches for some of the fine-grained classification datasets. An exception is the Oxford-Flowers dataset that Zernike seem to perform poorly. 

\section*{Acknowledgment}
This work was supported by ERC grant VIAMASS no. 336054 and ANR project FIRE-ID.

{\small
\bibliographystyle{IEEEtranS}
\bibliography{egbib}
}
\input{bios.tex}
\end{document}

%% file: intro.tex
\section{Introduction}
\label{sec:intro}

\lettrine{L}{ocal} image description is a popular research topic in computer vision, as it is involved in many applications such as image classification and particular object detection. 
Extracting local descriptors from an image consists of two steps. The \emph{detection} step selects regions of interest, which are normalized into fixed-size patches. 
The \emph{description} step produces a vector representation for each of the detected patches. 
The SIFT descriptor~\cite{L04} and its RootSIFT extension~\cite{AZ12} have been shown to perform very well for most applications. Many descriptors have been introduced in the last years to improve the description speed or descriptor compactness, such as SURF~\cite{BETV08}, CHOG~\cite{CTCTGG09} or BRIEF~\cite{CLSF10}. The matching accuracy is improved by learning the descriptor design~\cite{WB07}\cite{WHB09}\cite{SVZ14}. In this paper, we are solely interested in the detection stage and therefore adopt the gold-standard SIFT and RootSIFT. 

The early works in this line of research have focused on the detection of sparse interest points and regions, typically producing a few thousand descriptors per image. These approaches aim at extracting distinctive and repeatable image parts, such as blobs and corners~\cite{L04}\cite{MTSZMSKG05}\cite{MCMP02}\cite{BETV08}, offering covariance properties: the same regions should be detected under some geometrical transformations. From a historical perspective, the choice of sparse representations was arguably motivated by the lack computational and memory resources. 
Although such methods perform well and are still widely used for image matching, they are not competitive in other application scenarios such as image classification. 

Local feature detection has recently shifted towards denser techniques. Dense sampling is an easy way to provide a large number of patches and a better coverage of the objects of interest. Fei-Fei and Perona~\cite{FP05} were the first to show that dense patches leads to better classification accuracy. Nowak \etal~\cite{NJT06} argue that the key parameter for classification is the number of extracted patches. 
Similar conclusions hold for other tasks, like fine-grained classification~\cite{GMJP14} or action recognition in videos~\cite{WKSC11}\cite{JJB13}. 
Recent works~\cite{ZJG13}\cite{DGJP13}\cite{TAJ13}\cite{TJ14} also evidence that methods for image and particular object retrieval, which traditionally rely on sparse regions typically extracted with the Hessian-Affine detector~\cite{MiS04}, are improved when using a larger set of descriptors. 

Nevertheless, regular dense sampling has serious limitations. Uniform sampling of patches ignores the image structure and extracts many uniform and uninformative patches. Additionally, the position of the features is less or not repeatable. This prevents the image engine from employing a spatial verification method, such as RANSAC~\cite{bolles-ransac}\cite{PCISZ07}, which typically filters out many outliers by enforcing the spatial consistency of the detected regions. 

\begin{figure*}[b]
    \centering
    \begin{tabular}{c c c}
        \subfigure{\includegraphics[width=0.31\textwidth]{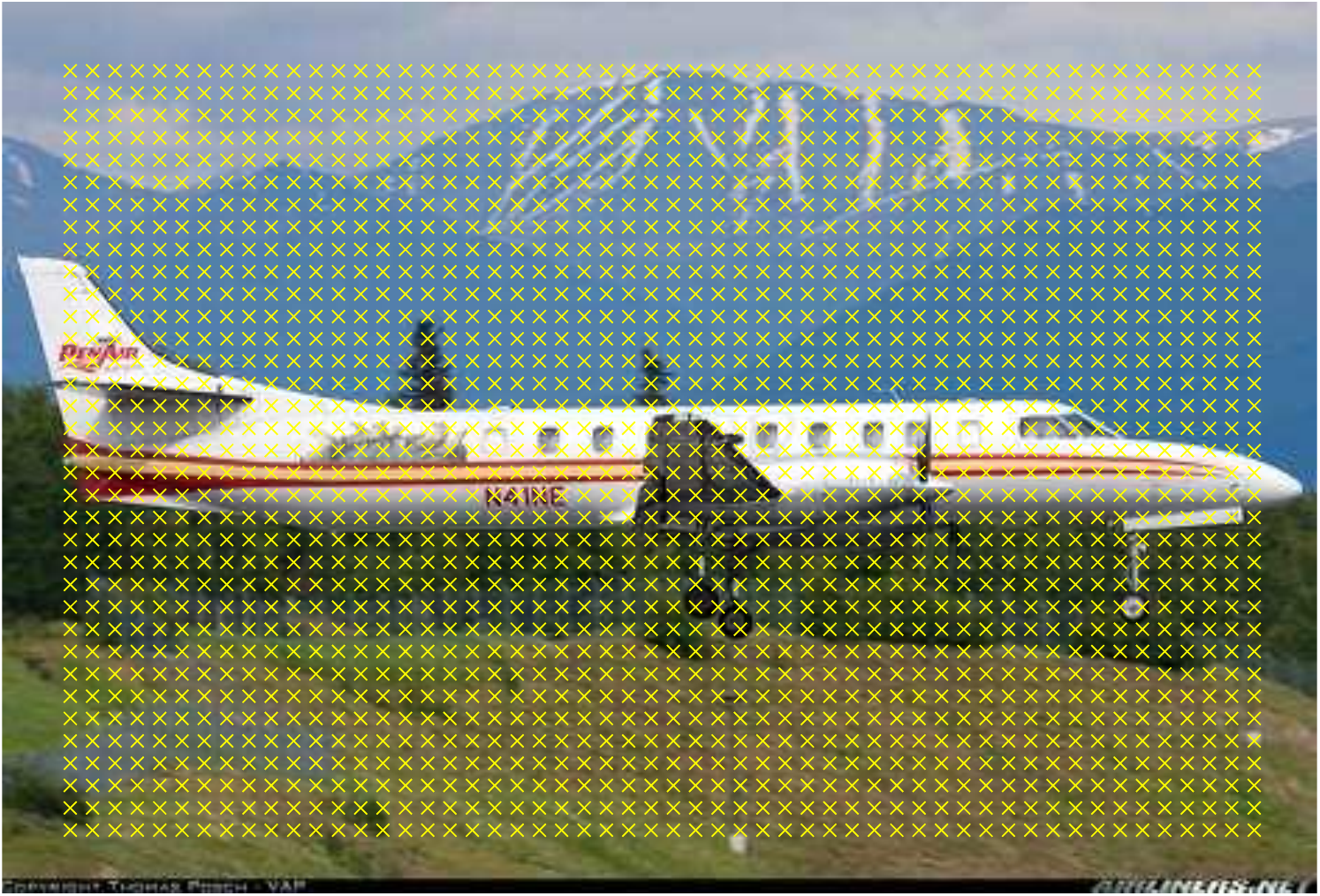}}
        &
        \subfigure{\includegraphics[width=0.31\textwidth]{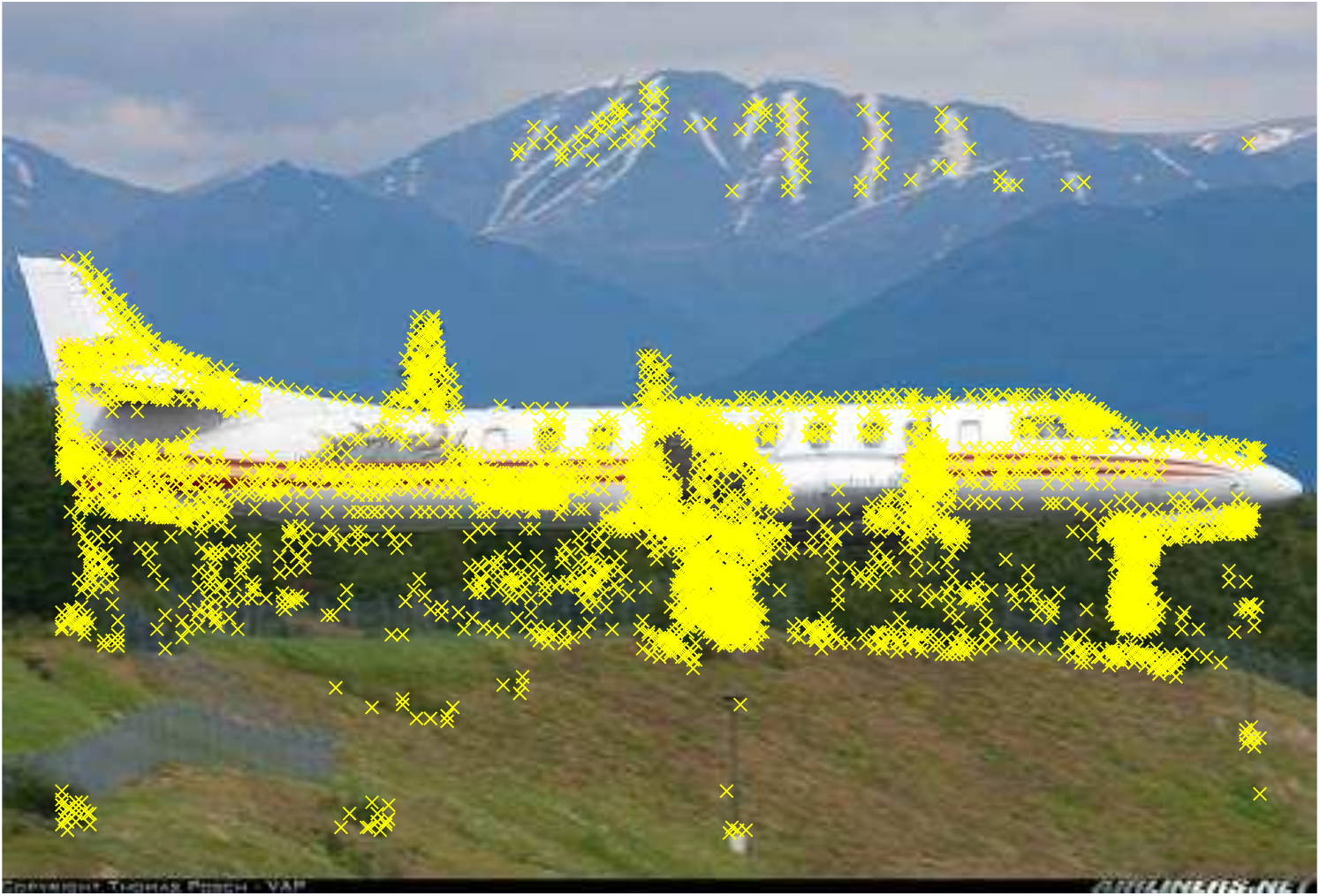}}
        &
        \subfigure{\includegraphics[width=0.31\textwidth]{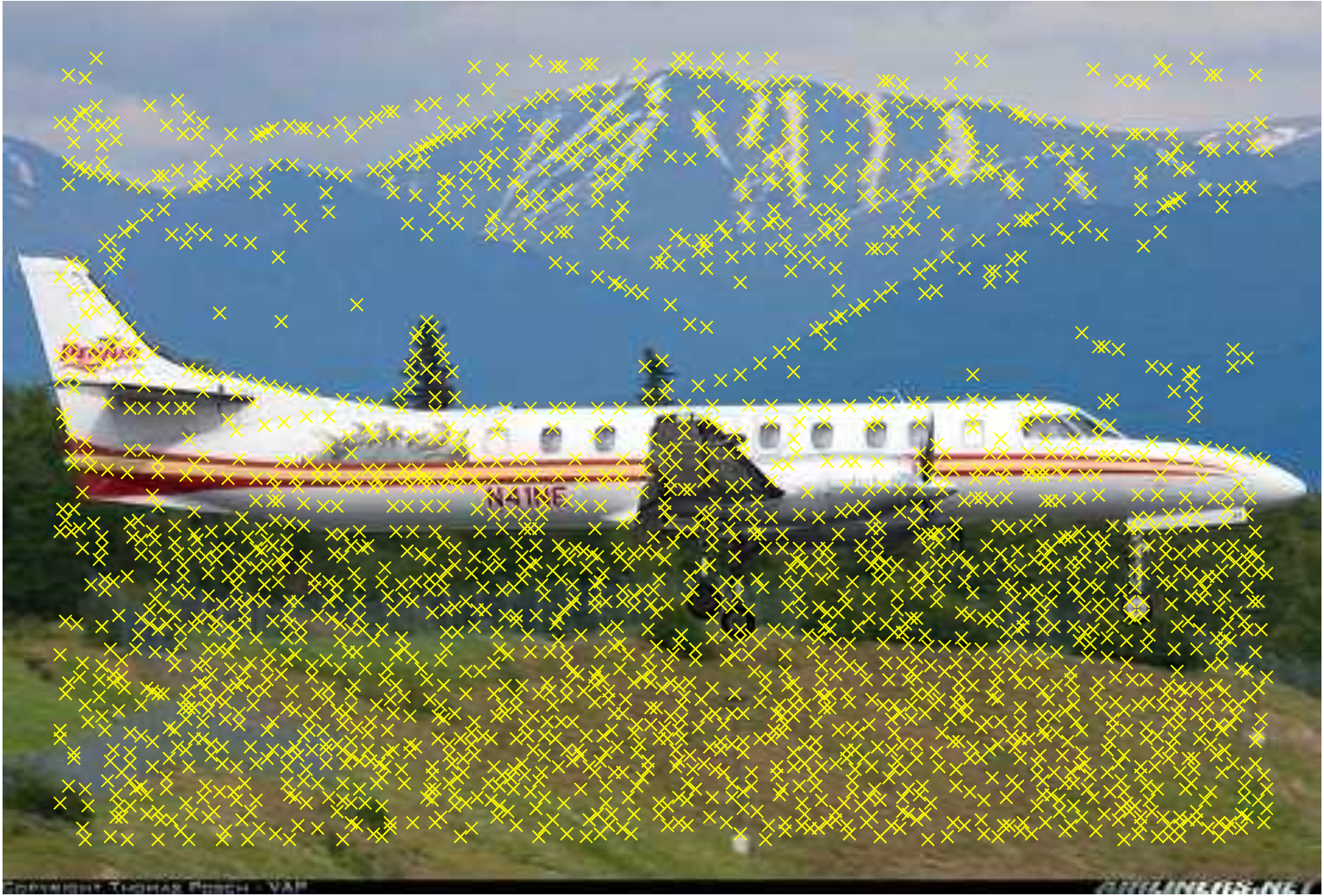}}
    \end{tabular}
\caption{Locations of patches detected by regular dense sampling (left), by Zernike detector (middle), and by dense \l2-norm based detector (right). We visualize the patches only for the first scale.}
\label{fig:intro}
\end{figure*}

In this paper, our goal is to develop and evaluate dense detection strategies for image retrieval and fine-grained image classification. Our motivation is similar to that of Tuytelaars when she introduced ``dense interest points''~\cite{Tu10}: we consider solutions in between localized sparse interest points and dense strategies, in order to produce a large number of localized regions. 
We depart from using traditional evaluation metrics for detector evaluation, which do not reflect the final goal. For instance, the repeatability score reflects the effectiveness in detecting inliers, but is not directly related to the determination of class membership.  
Instead, we evaluate the  performance with the metrics employed for the target application scenario: mean average precision for image retrieval and accuracy for fine-grained classification. 
We stress that better localized dense patches (see Figure~\ref{fig:intro}) are more important for these tasks than in traditional image classification: the objects are more repeatable and distinguishing between two classes often rely on tiny details that suffer from being loosely localized. 

We make the following contributions in this context of dense detection for image retrieval and fine-grained classification:

\begin{enumerate}
\item We propose strategies derived from standard interest point detectors (Harris, Hessian and DoG) to extract patches densely. In particular, we modify the detection process by relaxing the standard local maxima criterion so that it focuses on edges and not only corners. This, jointly with the optimization of the scaling factor, is shown to be a key to achieve higher performance.

\item We depart from the typical choice of fitting an ellipse to describe a region of interest extracted by blob detectors (MSER or super-pixels). Instead, we sample patches at several scales along the region's borders. This increases the performance significantly when considering a large number of patches per image. We also show that sampling on the edges produced by a state-of-the-art edge detector~\cite{DZ13} offers competitive performance. 

\item Finally, we propose two novel response filters to select the patch locations.  First, we propose to use a bank of Zernike polynomials~\cite{Zern34} as detectors. These filters have been proposed to construct descriptors~\cite{RLB09}, but to our knowledge not as detectors. Our second strategy is descriptor-oriented: the response for each pixel is simply the norm of local descriptor associated with the patch centered at this location. These two new approaches appear to perform best in most cases of our experiments.
\end{enumerate}

This paper is organized as follows. Section \ref{sec:related} discusses previous work related to the purpose of our study. Section \ref{sec:background} describes existing approaches which are part of our evaluation, while in section \ref{sec:methods} we present modified schemes of existing detectors and the use of pseudo-Zernike polynomials for dense feature detection. Finally in section \ref{sec:experiments} we present the outcome of this study on 2 datasets on image retrieval and 4 datasets on fine-grained classification.

%% file: related.tex
\section{Related work}
\label{sec:related}

A great deal of works has focused on the detection of local interest points~\cite{L04}\cite{BETV08}\cite{MiS04}\cite{MCMP02}. They are rather designed to be appropriate for image matching applications. The standard evaluation metrics for such detectors~\cite{MTSZMSKG05}, \eg repeatability, are designed to reflect sufficiency for matching applications. However, this could be far from the final application tasks we focus in this paper, \ie fine-grained classification.

Typically, local interest points are tuned to detect image structures, such as corners and blobs, highly distinctive and repeatable. One of the exceptions is the work of Mikolajczyk \etal~\cite{MZS03} where they focus on edges to represent objects. Similarly, in this study we argue that such a choice is beneficial for image retrieval and fine-grained classification. 

Tuytelaars's work on dense interest points~\cite{Tu10} has a motivation close to ours. Initially, a dense grid of patches is considered, and then for each point, local refinement of its location and scale parameters is performed. In particular, the point of maximum \emph{interestingness} is selected within a bounded area. This choice suffers from quantization artifacts, since a true local maximum might not exist in this search area. Apparently, a large number of patches are located on smooth and uninformative regions.

A previous survey closely related to this study is the work of Nowak \etal~\cite{NJT06}. They are also not interested in the repeatability of patches, but directly measure classification accuracy. According to them, detectors designed to obtain high repeatability perform the same as randomly selected patches. They observe that performance is, up to some extent, an increasing function of the number of points per image. Inspired by their finding, we conduct a similar evaluation and try to study different ways to control the average number of patches. Similarly, Avrithis and Rapantzikos~\cite{AR11} do not restrict comparison to repeatability, but further compared image retrieval performance, while considering the average number of points as a crucial parameter.

One of the parameters we consider in our study is the size of the measurement region for local descriptor extraction. This has been given particular attention by the recent work of Simonyan \etal~\cite{SVZ14}, who shows that introducing a scaling factor increases retrieval performance. Similarly, the improved Hessian-Affine detector by Perdoch \etal~\cite{PCM09} uses a larger measurement region to improve the performance.

In regular sampling local descriptors are agnostically extracted on a dense grid, and as a consequence many uninformative descriptors are derived from smooth regions. A significant improvement is achieved by filtering out descriptors with low \l2-norm~\cite{GMJP14}. We combine this approach with examined detectors and study its effect on different setups.
\medskip

\paragraph{Convolutional neural networks} are very effective for image classification~\cite{KSH12}\cite{DJVH14} and detection~\cite{GDDM14}. However, for the tasks addressed in this paper, namely fine-grained classification, image and particular object, this is not (yet) the case. Fisher vectors based on a dense representation and employing medium-sized codebooks and spatial coding achieved much better results in the fine-grained challenge~\cite{GMJP14}. For image retrieval, recent work demonstrates that CNNs~\cite{BSCL14} achieve excellent performance for aggressive operating points for compact representations. However, their best reported performance is lower than that of the state-of-the-art Fisher vectors, and outperformed by a large margin by state-of-the-art methods like the selective match kernels~\cite{TAJ13}.

%% file: background.tex
\begin{figure*}[t]
    \centering
    \begin{tabular}{c c c c}
        \subfigure[Dense]{\includegraphics[width=0.22\textwidth]{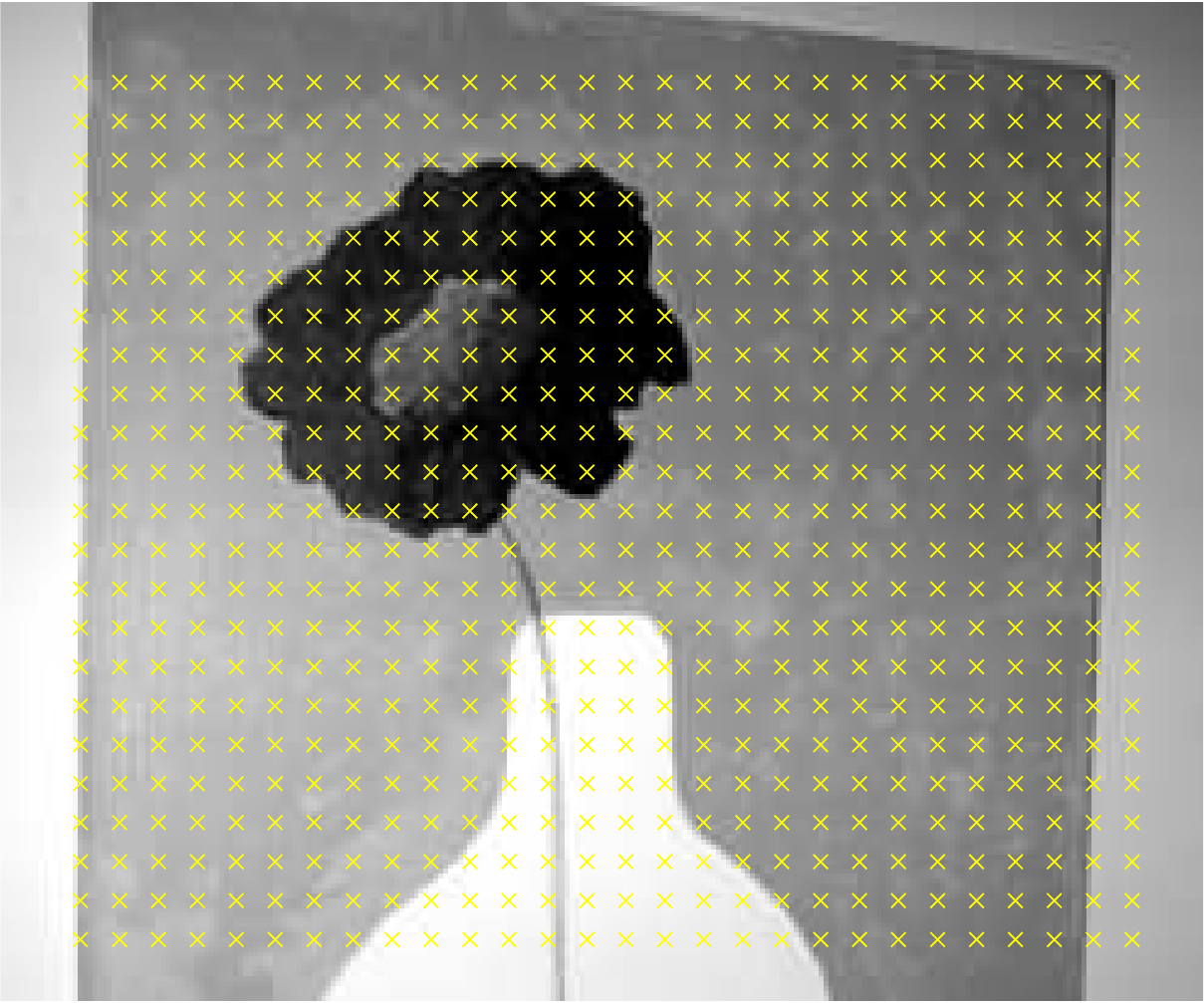}\label{capt:dense}}
        &\subfigure[Dense-IP]{\includegraphics[width=0.22\textwidth]{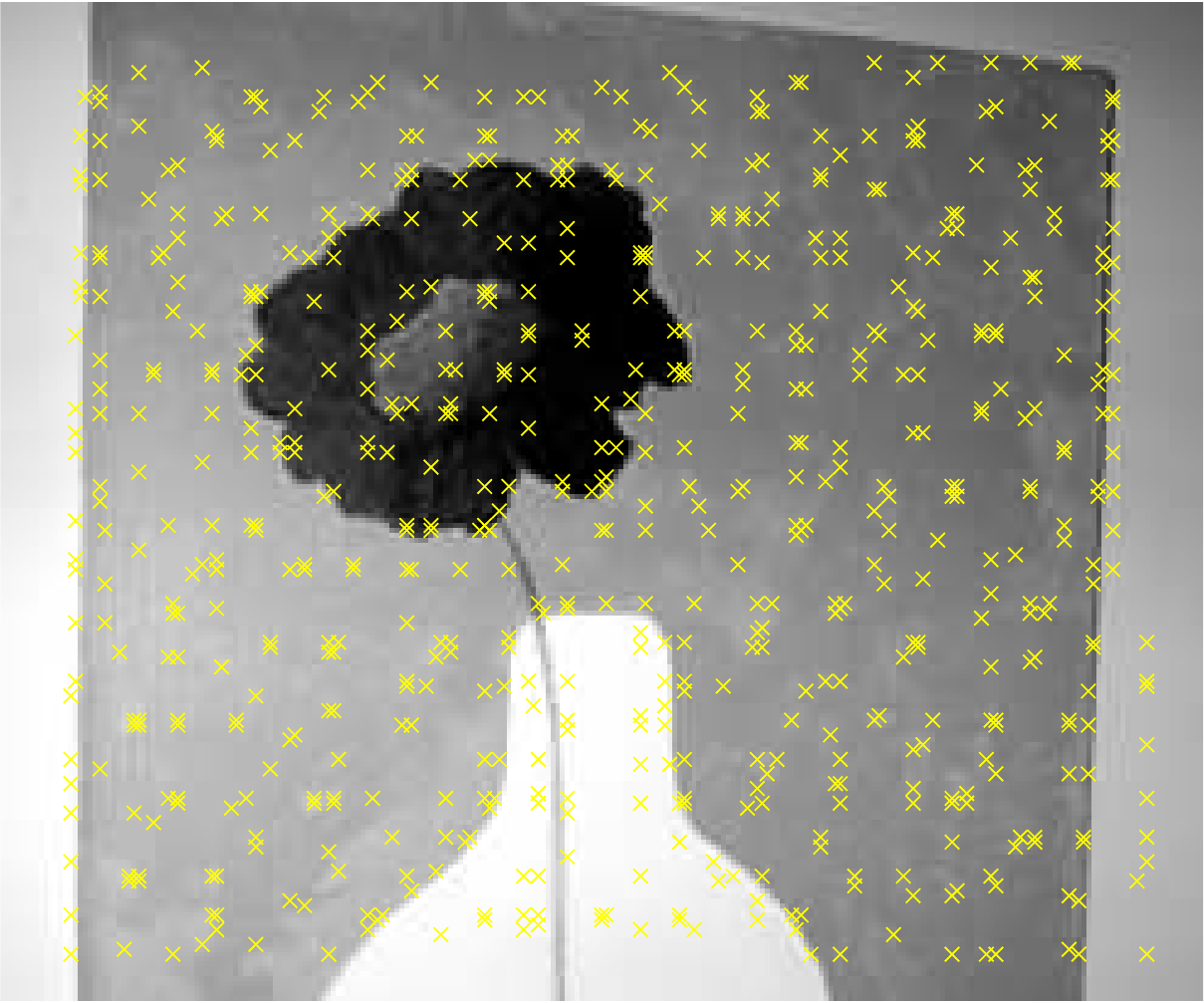}\label{capt:denseip}}
        &\subfigure[Zernike]{\includegraphics[width=0.22\textwidth]{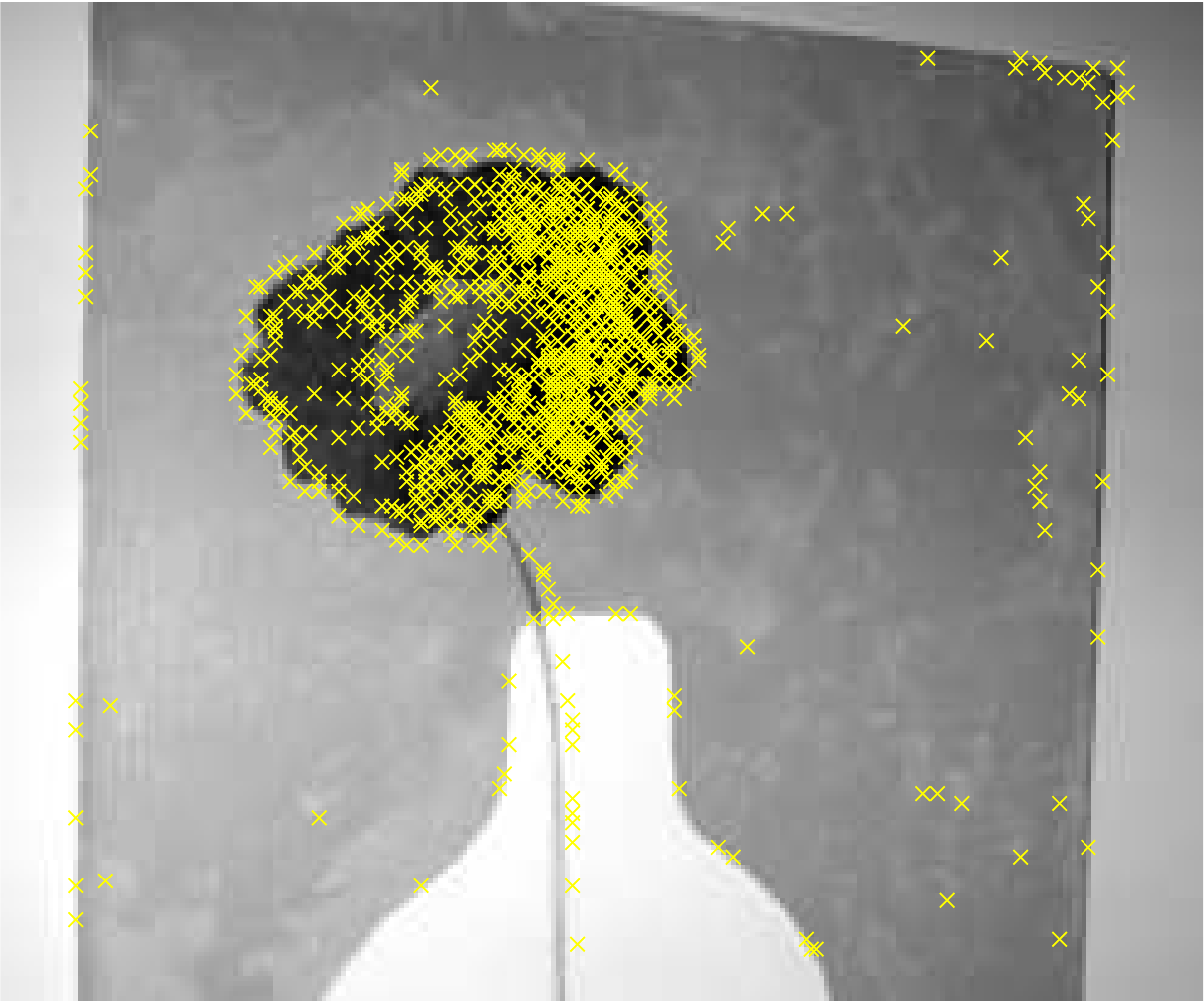}\label{capt:zernike}}
        &\subfigure[Dense \l2-norm]{\includegraphics[width=0.22\textwidth]{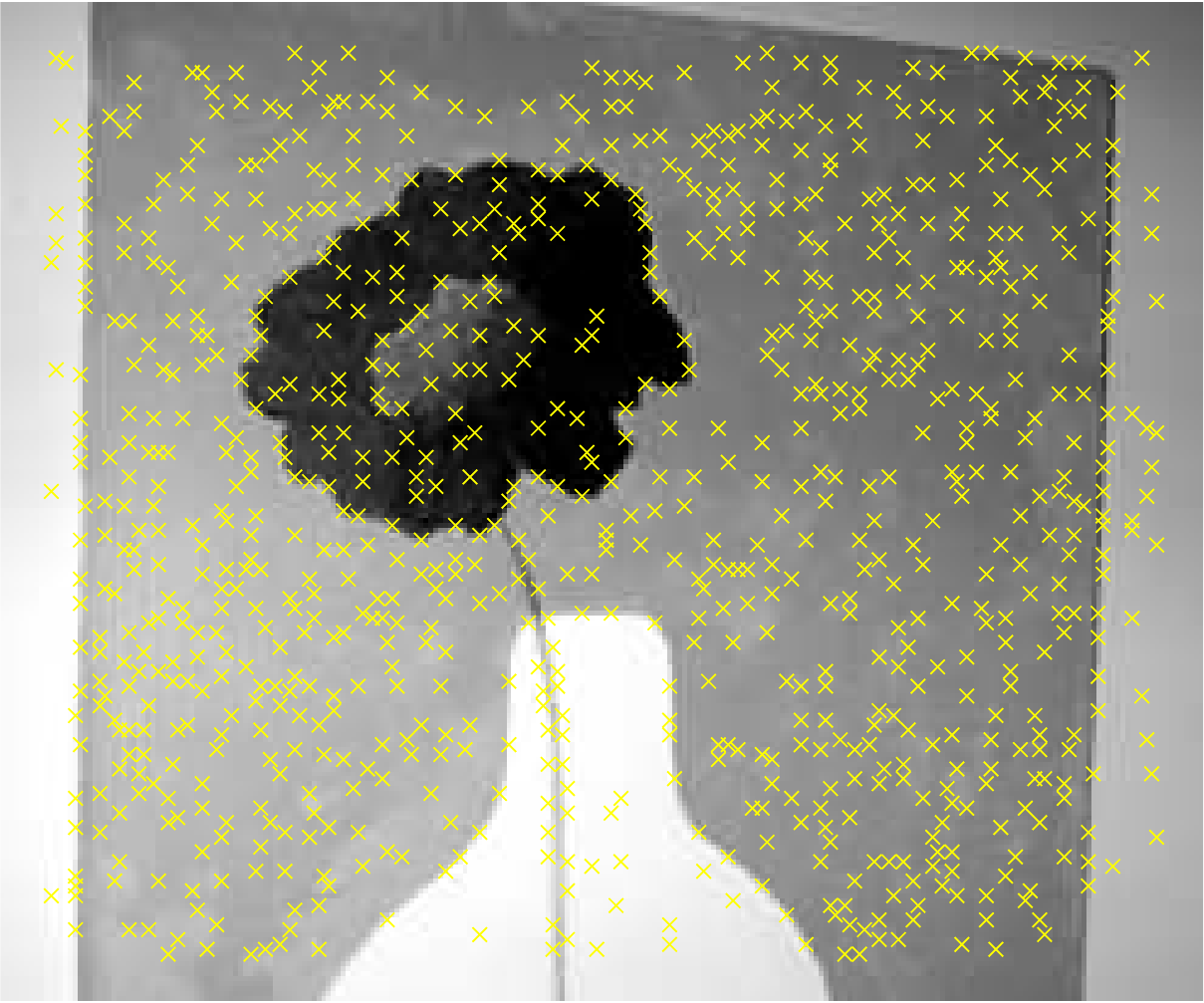}\label{capt:densel2}}\\
        \subfigure[MSER-edge]{\includegraphics[width=0.22\textwidth]{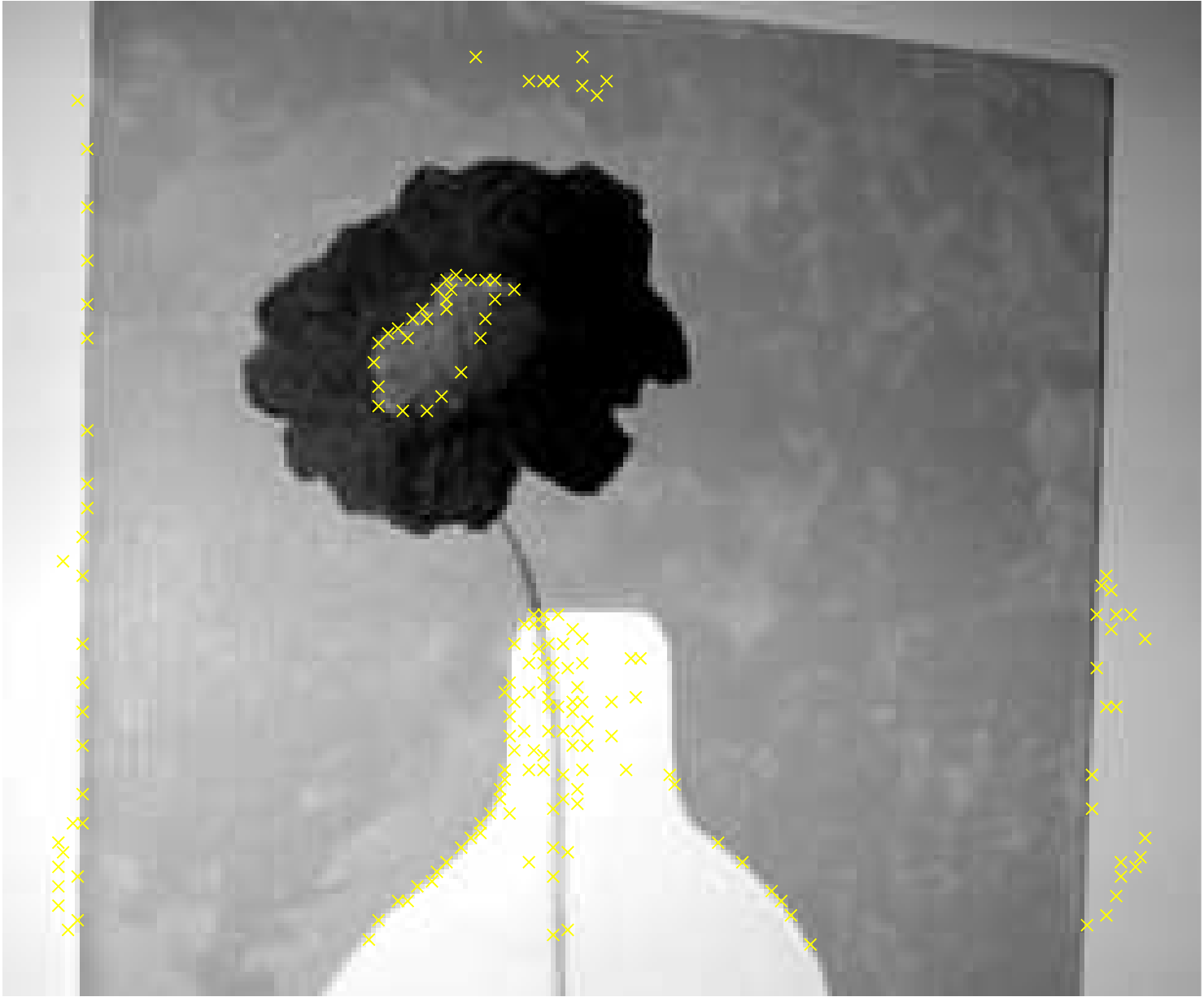}\label{capt:mseredge}}
        &\subfigure[SSR-edge]{\includegraphics[width=0.22\textwidth]{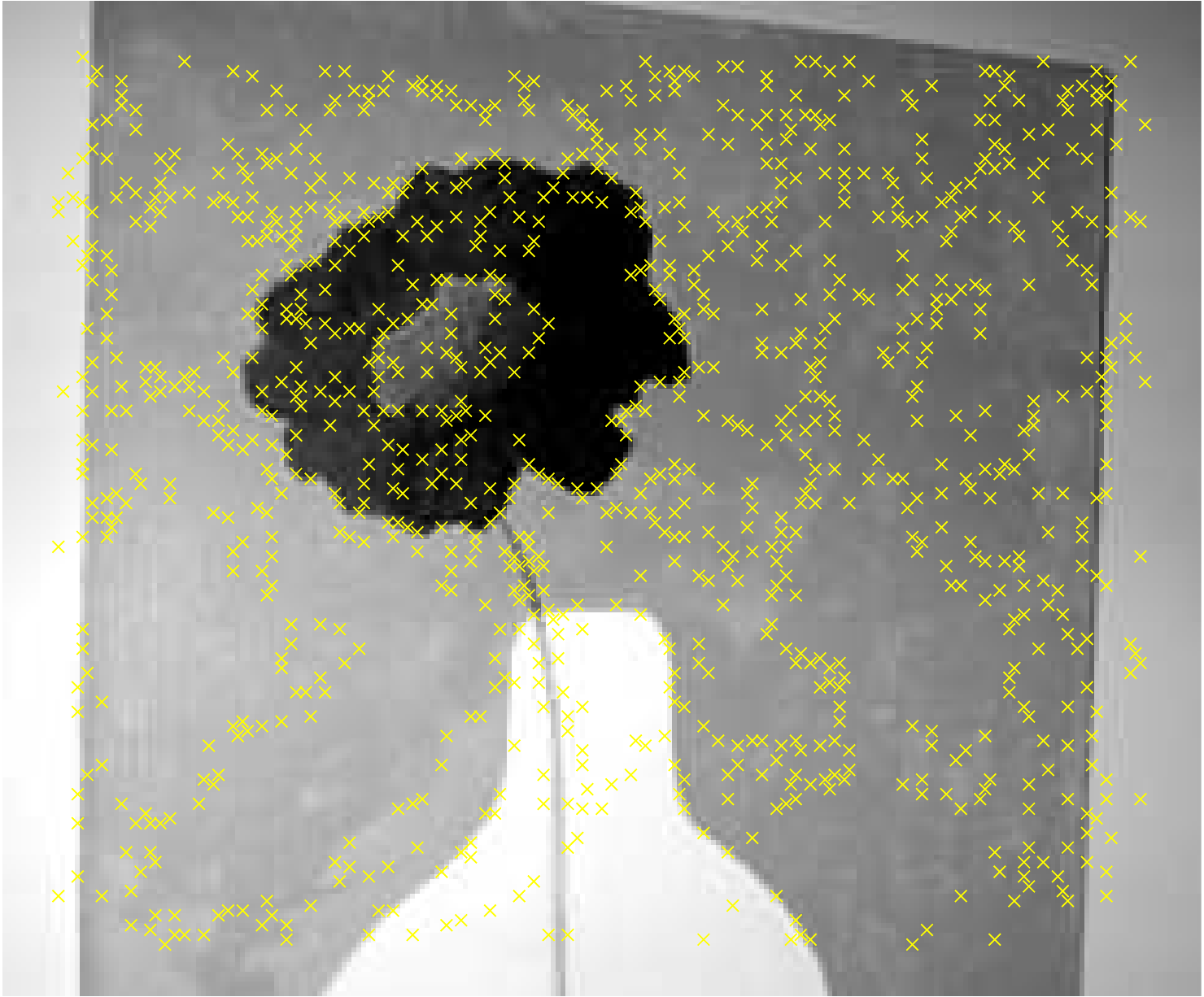}\label{capt:ssredge}}
        &\subfigure[fast-edge]{\includegraphics[width=0.22\textwidth]{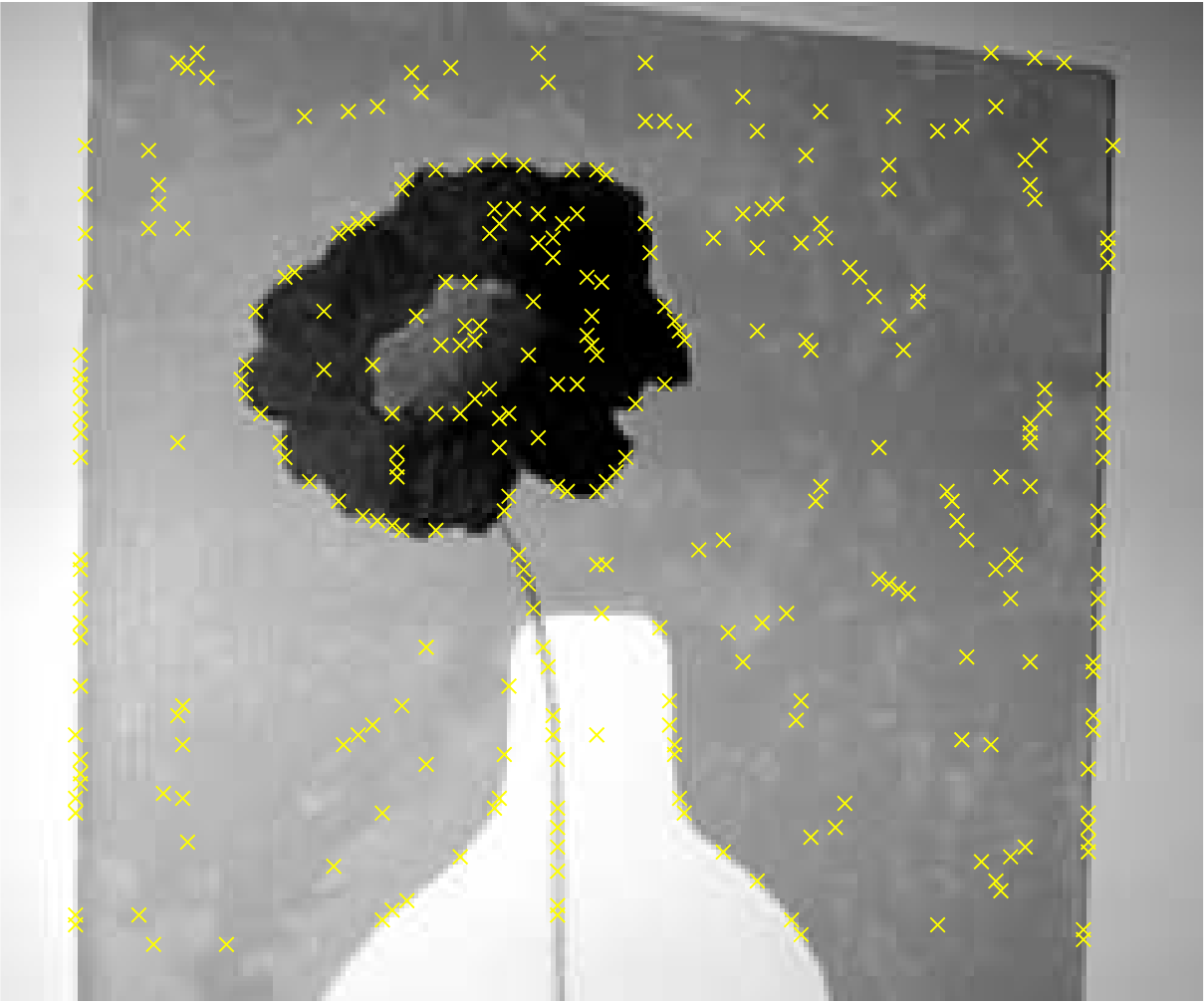}\label{capt:fastedge}}
        &\subfigure[DoG]{\includegraphics[width=0.22\textwidth]{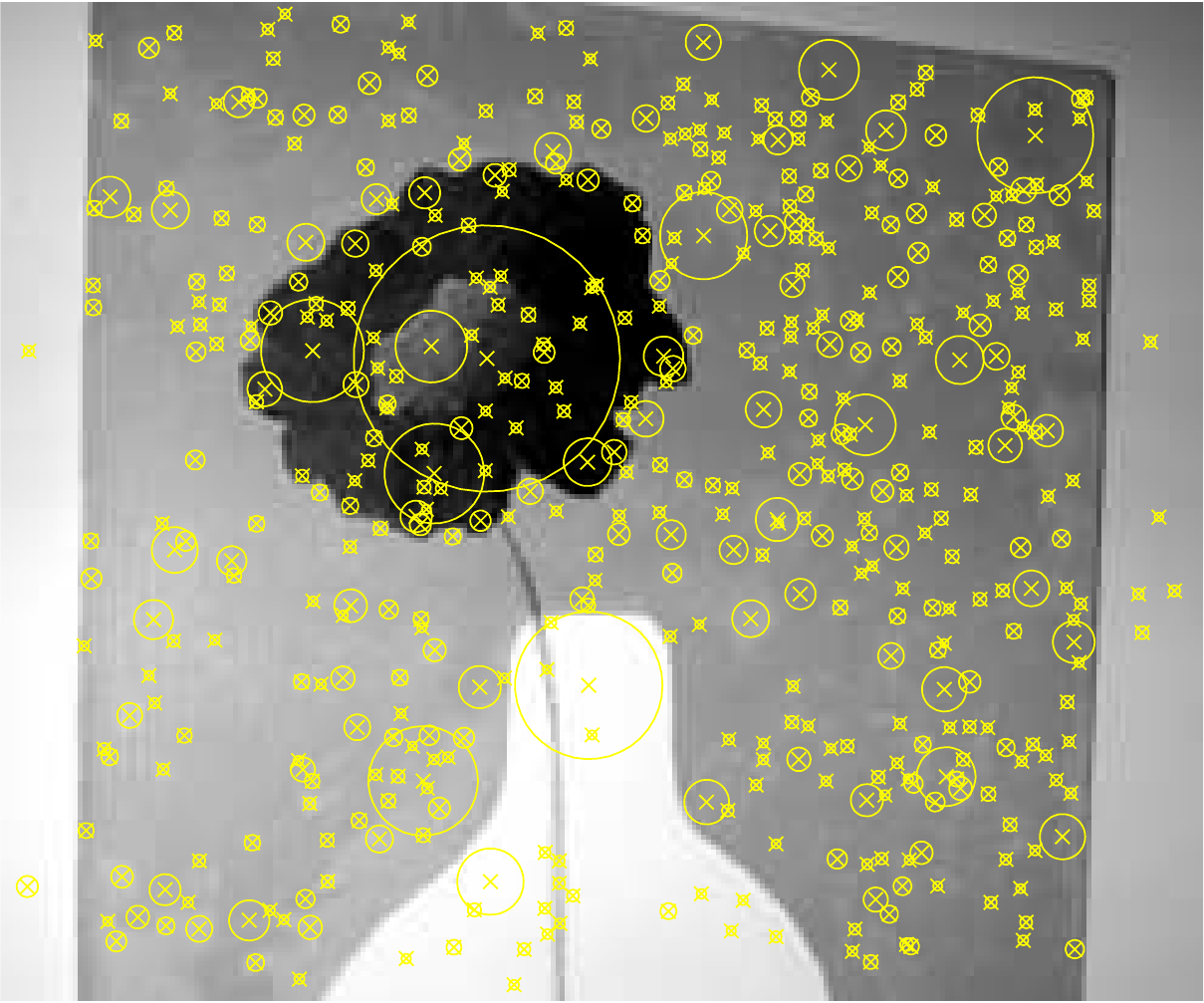}\label{capt:dog}}\\
        \subfigure[MSER]{\includegraphics[width=0.22\textwidth]{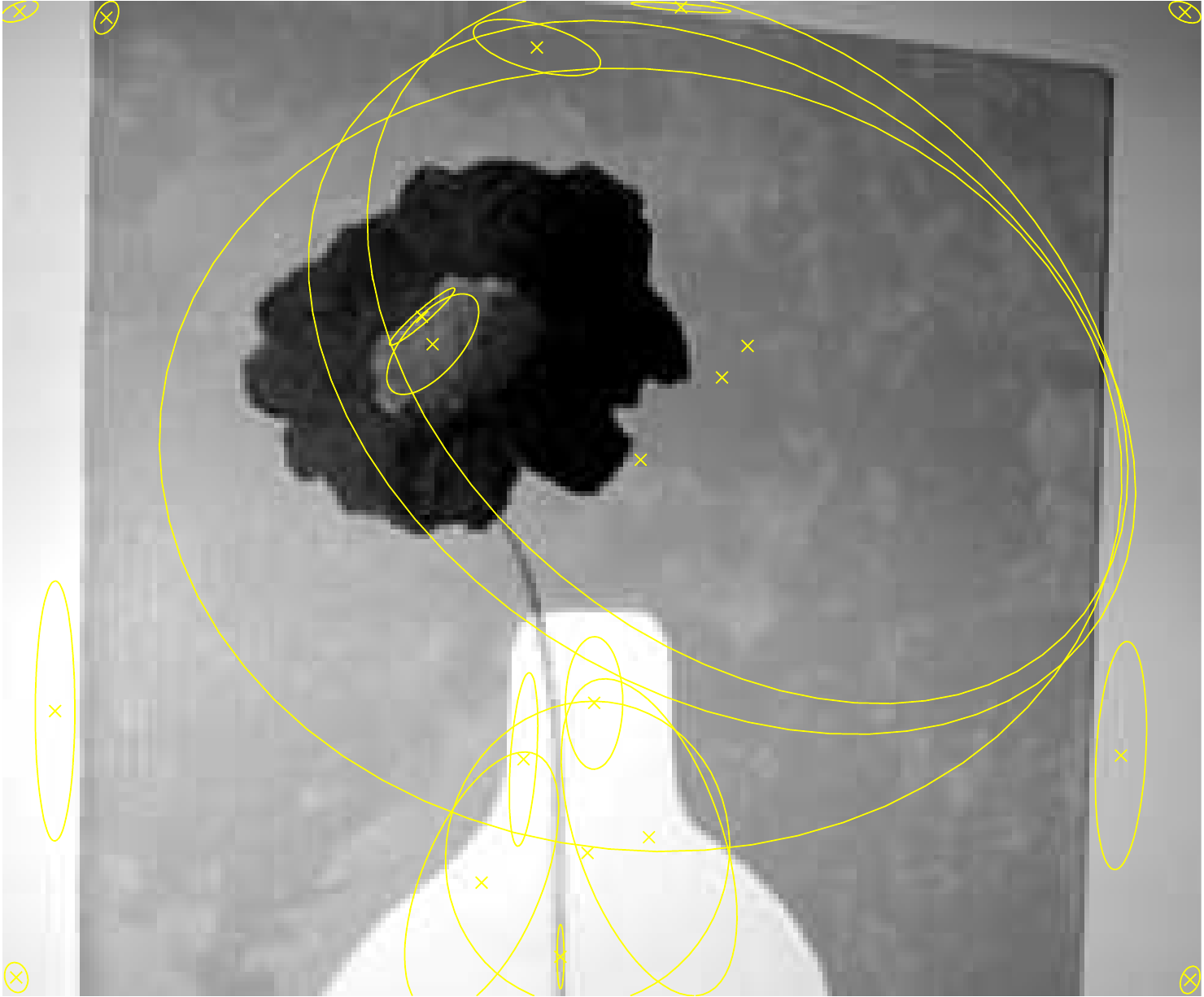}\label{capt:mser}}
        &\subfigure[SSR]{\includegraphics[width=0.22\textwidth]{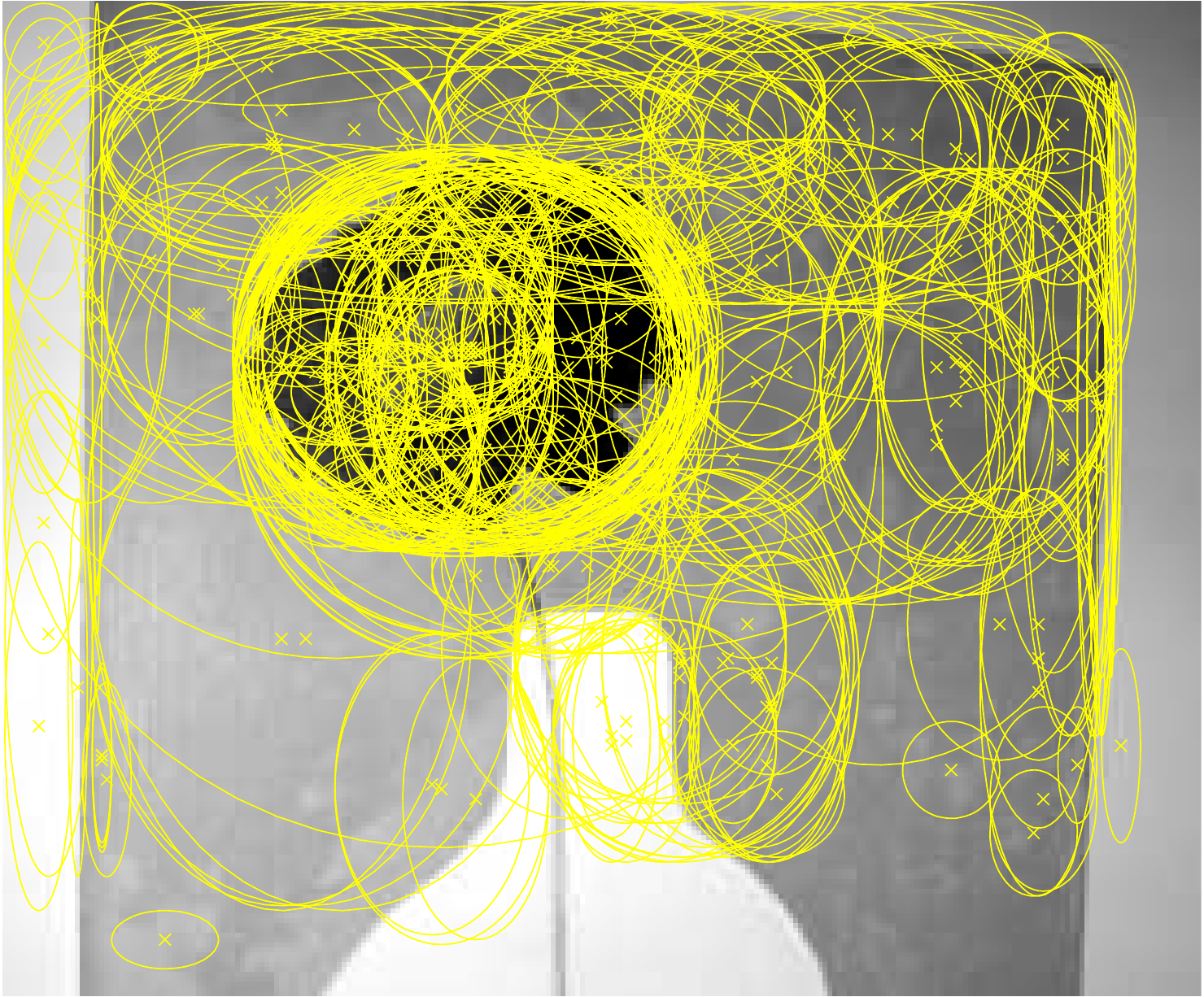}\label{capt:ssr}}
        &\subfigure[Harris]{\includegraphics[width=0.22\textwidth]{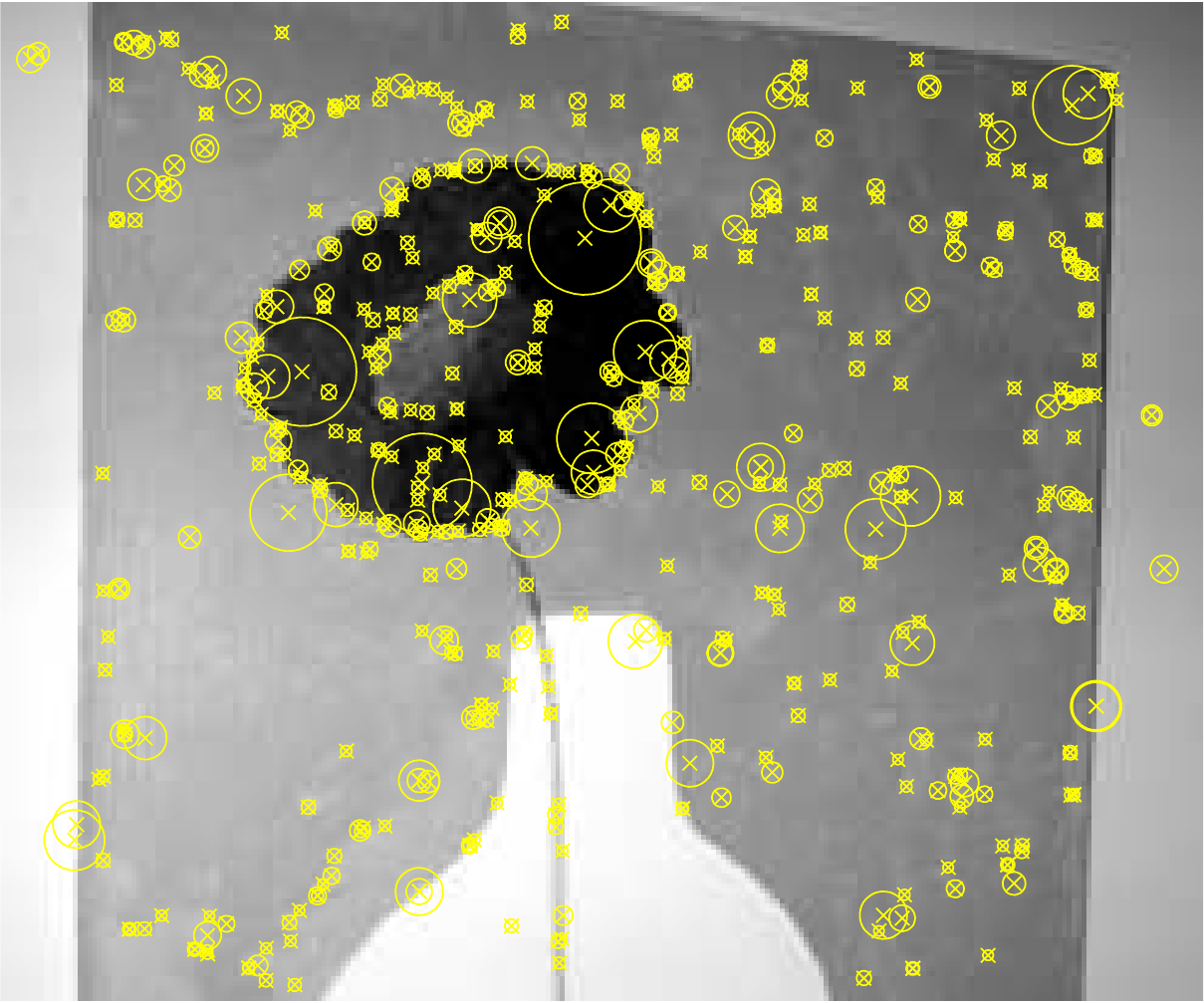}\label{capt:harris}}
        &\subfigure[HesAff]{\includegraphics[width=0.22\textwidth]{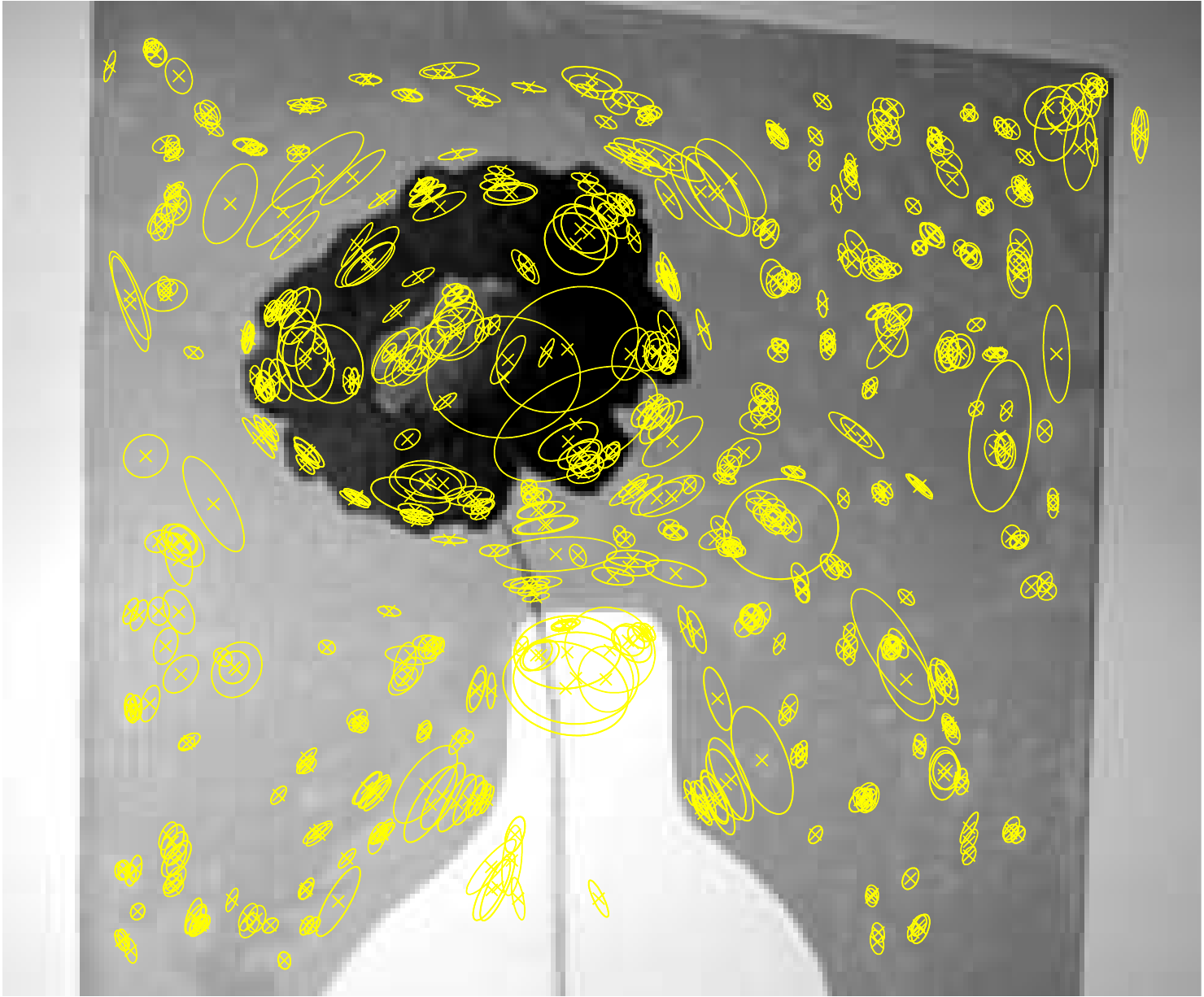}\label{capt:hesaff}}
    \end{tabular}
\caption{Visualization of regions detected by each method. For methods with predefined scales, we visualize the center of points detected on the first scale \subref{capt:dense}-\subref{capt:fastedge}, while for the ones that perform scale selection we draw the corresponding ellipses or circles \subref{capt:dog}-\subref{capt:hesaff}.}
\label{fig:detectors}
\end{figure*}

\section{Background}
\label{sec:background}

In this section we briefly describe existing methods for interest point detection or dense patch sampling, that are part of our study. In all those methods, a local descriptor is extracted from each region of interest and an image is represented by a set of such descriptors. We visualize detected features for all the detectors examined or proposed in this work on Figure~\ref{fig:detectors}.

\subsection{Interest points}
\label{sec:intpoints}

\paragraph{Harris-Laplace detector.} Harris detector localizes corners, based on the fact that gradient values will change in multiple directions around a corner.
It uses a scale adapted version of the second moment matrix, known as Harris matrix~\cite{HarM88}:
\begin{equation}
M =
 \sigma_D^2 \cdot g(\sigma_I) \ast
\begin{bmatrix} 
L_x^2(x,\sigma_D) & L_xL_y(x,\sigma_D) \\
L_xL_y(x,\sigma_D) & L_y^2(x,\sigma_D)
\end{bmatrix},
\end{equation}
where $\sigma_D$ is the differentiation scale, $\sigma_I$ is the integration scale and $L_z$ is the derivative computed in $z$ direction. 
Differentiation scale $\sigma_D$ is used to compute the local derivatives with Gaussian kernels, and a Gaussian window with a size $\sigma_I$ is used to smooth and average the neighborhood around the point. 
The eigenvalues of this matrix represent the gradient changes in two directions. 
Consequently, if one eigenvalue is large while the other is small, there exists an edge, whereas if both eigenvalues are large, there exists a corner. 
The \textit{interestingness} of a point is captured by the \textit{cornerness} function, defined as $cornerness = \text{det}(M) - \alpha \text{trace}^2(M)$, where $\alpha$ is usually set to $0.05$. 

Extending Harris detector to be scale invariant, Mi\-kolajczyk \etal~\cite{MiS04} use Laplacian-of-Gaussian (LoG) response and detect local extrema over multiple scales to perform scale selection~\cite{Lin98}. This is the well known Harris-Laplace detector. Only points with cornerness value higher than threshold $\tau$ are retained and final point locations are chosen by a local maxima search procedure. 
\medskip

\paragraph{Hessian-Affine detector}~\cite{MTSZMSKG05}, similar to Harris detector, detects image locations that have large derivatives in both directions. 
Point locations are selected as local maxima of the Hessian matrix determinant, which now constitutes the interestingness measure.
The Hessian matrix is defined as
\begin{equation}
H = 
\begin{bmatrix} 
L_{xx}(x,\sigma_D) & L_{xy}(x,\sigma_D) \\
L_{xy}(x,\sigma_D) & L_{yy}(x,\sigma_D)
\end{bmatrix},
\end{equation}
where $L_{zz}$ is the second order partial derivative.
All points with interestingness below threshold $\tau$ are discarded. 
Although Hessian and Harris detectors are quite similar, the detected points may be slightly different. 
In particular, unlike Harris, Hessian detector tends to select locations with texture variations, in addition to corners.

The affine shape of the point is estimated by the eigenvalues of the second moment matrix $M$. An iterative procedure modifies the point's location, scale and shape until the estimated affine transform is able to map the detected region into one that has equal eigenvalues of its second moment matrix.
\medskip

\begin{figure*}[t]
    \centering
    \begin{tabular}{c c c c}
        \subfigure[Harris]{\includegraphics[width=0.21\textwidth]{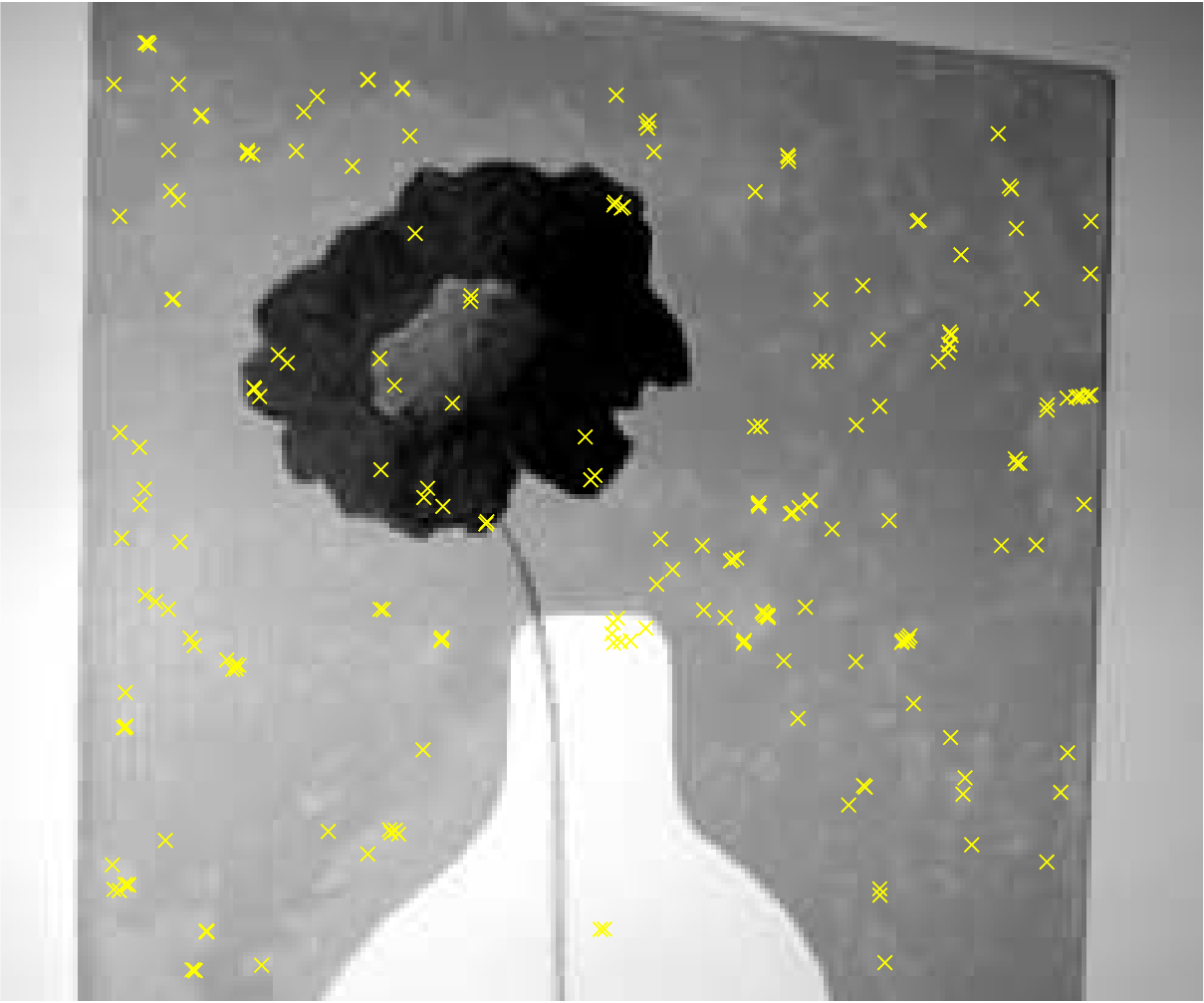}}
        &
        \subfigure[Frobenius]{\includegraphics[width=0.21\textwidth]{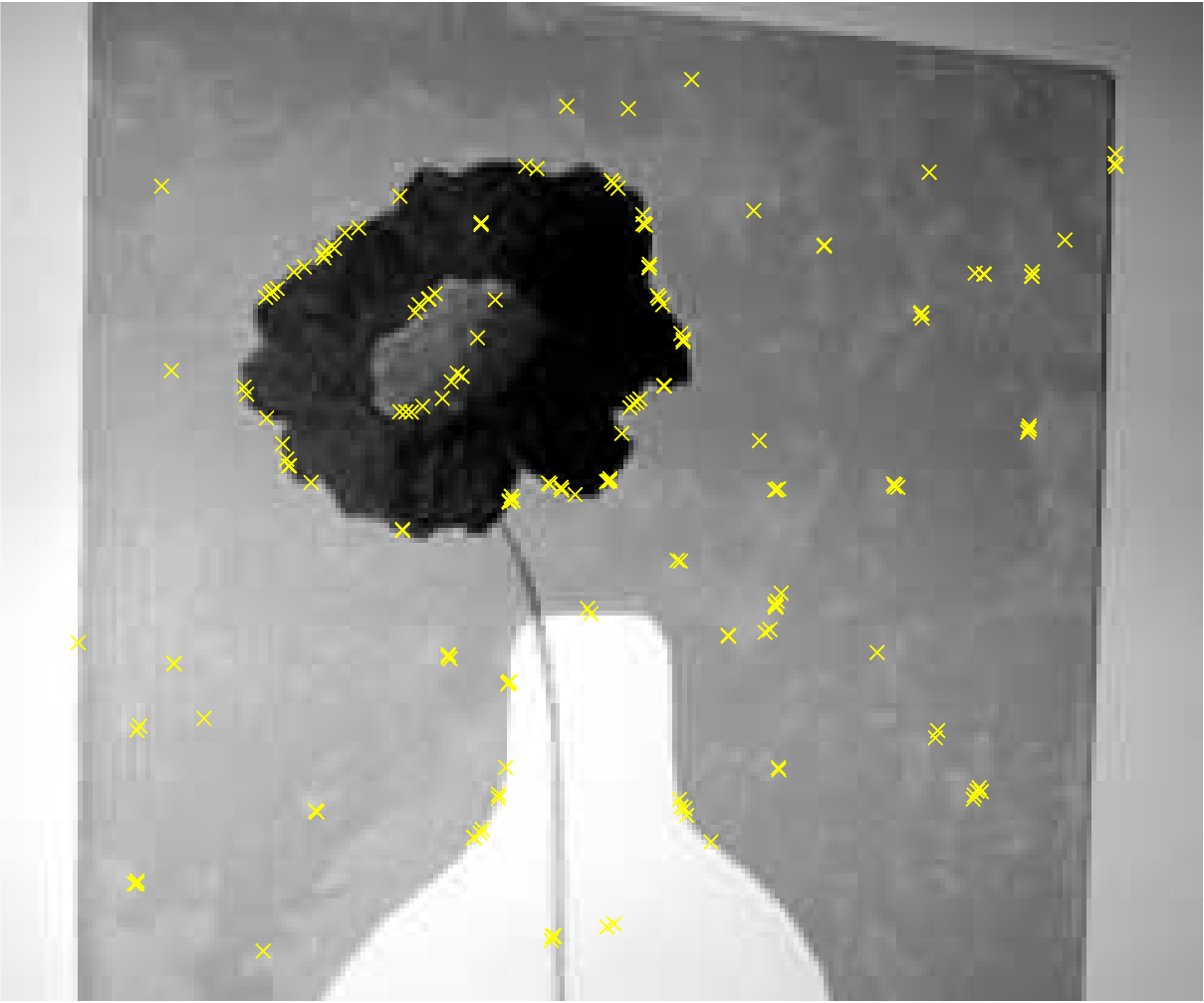}}
        &
        \subfigure[relaxed-Harris]{\includegraphics[width=0.21\textwidth]{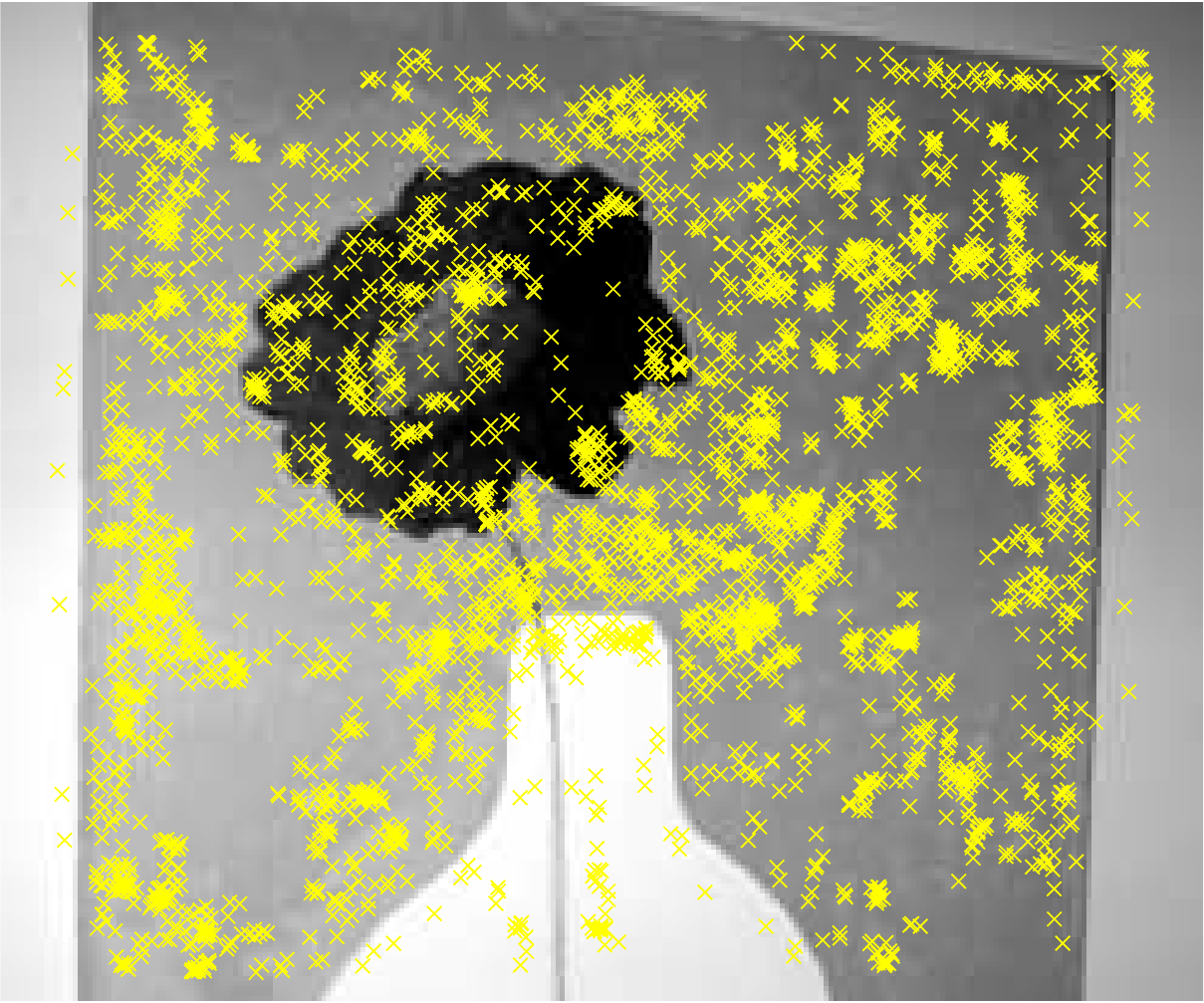}}
        &
        \subfigure[relaxed-Frobenius]{\includegraphics[width=0.21\textwidth]{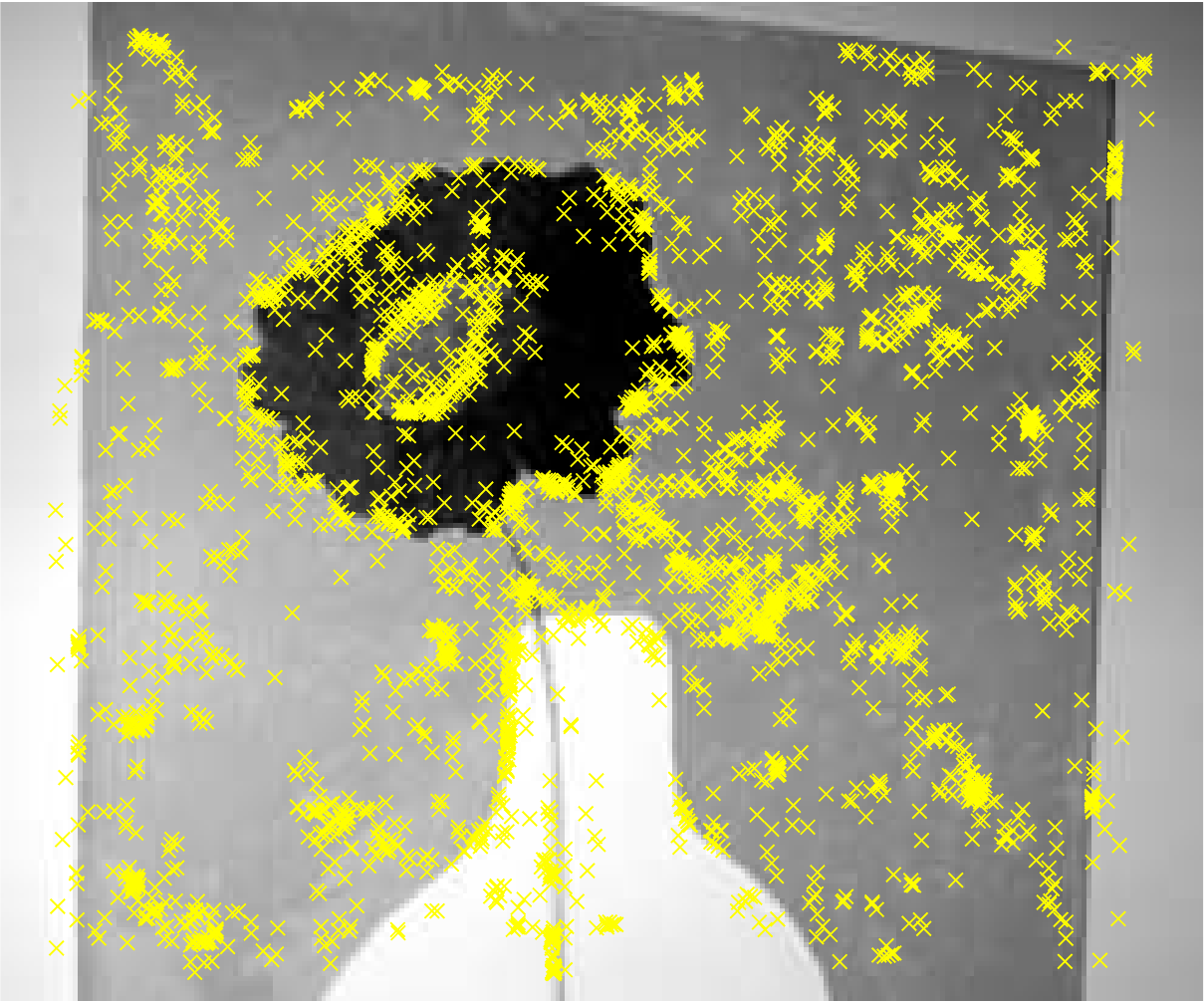}}
    \end{tabular}
\caption{The detected keypoints for our Harris Laplace modifications. We visualize the centers of detected points.}
\label{fig:harrisdets}
\end{figure*}

\paragraph{MSER} \emph{Maximally Stable Extremal Regions} were introduced by Matas \etal ~\cite{MCMP02} as a feature detector that tends to localize blobs. 
An \textit{extremal region} has all its intensity values greater (or less) than the outer region boundary pixels.
Then, a sequence of nested extremal regions is considered. Along this sequence scale change between neighboring regions is estimated. Maximally stable are the local minima of this quantity.
In this fashion, nested regions are also likely to appear.
The detected regions can have very irregular shapes, therefore an ellipse is fitted to detected regions in order to extract local descriptors. A parameter $\Delta$ controls the locality of the scale change computation.
\medskip

\paragraph{Difference of Gaussians (DoG)} was originally used for local feature detection by Lowe~\cite{L04}.
DoG is an efficient way to approximate the Laplacian of Gaussian and to detect edges at various image scales.
A Gaussian kernel is used to create multiple blurred versions of the image per octave.
Simple subtraction of two consecutive blurred images produces the DoG response. 
Interest points are located in the scale-space as local-maxima in a 3D search area of size $3$.

The original SIFT detection algorithm~\cite{L04} employs DoG, but further applies additional steps to filter out points belonging to edges or low-contrast regions.
However, in our experiments with DoG detector, we keep all the points for a denser representation.
\medskip

\subsection{Dense patch sampling}
\label{sec:densesample}

\paragraph{Regular grid dense sampling.} In contrast to interest points, dense sampling methods give less importance to high repeatability and try to provide a dense coverage of the depicted objects. The most popular method is to sample points on a regular grid, every $\delta_{xy}$ pixels. Depending on the application $\delta_{xy}$ can be really small, such as $3$, or quite larger, such as $16$. In order to provide some scale tolerance, different scales are considered by following the same procedure at $n_\mathrm{\sigma}$ multiple scales of the image. All the patches from different scales are pooled together in the end. A typical value for $n_\mathrm{\sigma}$ is 5, with 2 scales per octave. We adopt this choice in our experiments. 
\medskip

\paragraph{Dense interest points} were introduced by Tuytelaars~\cite{Tu10} as a hybrid solution to trade-off between sparse interest point detection and dense sampling.
Instead of selecting the center pixel of grid cell, as in regular dense sampling, they conduct local search inside each cell over spatial and scale space.
The point with maximum response is  kept per cell. Selected points are not necessarily local maxima.
Since a single interest point is selected from each cell, patches are very likely to be localized on smooth regions.
Moreover, quite a few patches are localized on the cell borders. 
In the original work, a large local search area of $16\times16$ pixels $\times$ $8$ scale levels is used, while in our study we experiment with smaller search areas in order to supply more points.
\medskip

\begin{figure}[b]
    \centering
    \begin{tabular}{c|c}
        \includegraphics[height=0.165\columnwidth]{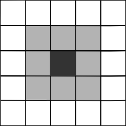}
        &
        \includegraphics[height=0.165\columnwidth]{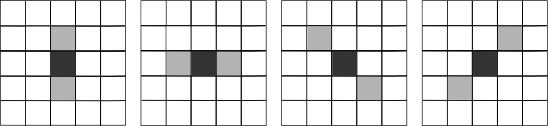}
    \end{tabular}
\caption{The search neighborhood typically used to detect  local maxima is shown on the left. We relax it by taking the union of extrema obtained from multiple neighborhoods shown on the right.}
\label{fig:localmax}
\end{figure}

%% file: methods.tex
\section{Proposed Methods}
\label{sec:methods}

Our purpose is to provide a dense set of patches without targeting high repeatability. We rather focus on accuracy in retrieval and fine-grained classification.
The proposed methods try to provide a nice coverage of the depicted objects, while focusing on edges in addition to corners and blobs. 

In this section we present the proposed approaches for dense patch representation. 
We relax the Harris-Laplace detector to detect mostly edges. We propose to use convolution by Zernike polynomials~\cite{Zern34} as a response function to select point locations. Moreover, we modify the well-known region detectors, such as MSER, to sample points on edge maps, or the edges of the detected regions.
Finally, we propose a detector that exploits the \l2-norm of already extracted local descriptors to select the patches.
In the following, whenever we mention that local maxima search is performed, a $3\times3$ neighborhood is used.

\subsection{Relaxing Harris-Laplace detector}
\label{sec:harlapext}
Harris-Laplace detector is designed to detect corners. We propose to replace Harris-Laplace cornerness function with Frobenius norm in order to select other points in addition to corners.
We also propose to adjust the local extrema selection criterion to sample denser points for classification context.

The first modification is to adjust the standard cornerness function to sample denser points. We estimate the energy of the Harris matrix by its \emph{Frobenius norm}. 
It is defined as the square root of the sum of the absolute squares of its elements.
This offers the ability to produce high response for structures with high energy other than corners.

As our second modification, we propose to modify the local maxima criterion. 
Regularly, keypoints are selected as local maxima of some  interestingness measure. 
Instead of using the 8-neighborhood of a pixel as local maxima criterion as the standard procedure, we propose to use multiple 2-neighborhoods in different directions, as shown in Figure~\ref{fig:localmax}. We select the points that are local maxima in any of those neighborhoods.
In this fashion, we retrieve more points along the edges of objects or object parts. 

In order to investigate the contribution of different modifications, we end up with four different methods. These are, 
the regular Harris-Laplace (Harris), Frobenius norm based detector (Frobenius), Harris-Laplace with relaxed local maxima selection (relaxed-Harris) and Frobenius norm based detector with relaxed local maxima selection (relaxed-Frobenius).

In the example of Figure~\ref{fig:harrisdets}, we observe that points are located on edges as well as corners using the Frobenius norm.
The relaxed local maxima criterion also provides more points outlining the shapes on the image.

\subsection{Zernike}
\label{sec:zernikefilt}

\begin{figure}
\begin{tabular}{@{\sssp}c@{\sssp}c@{\sssp}c@{\sssp}c@{\sssp}c@{\sssp}c@{\sssp}c@{\sssp}c@{\sssp}c@{\sssp}}

&&&$v_{1}^{-1}$ &$v_{1}^{0}$ & $v_{1}^{1}$ &&&\\
&&&\includegraphics[height=26pt]{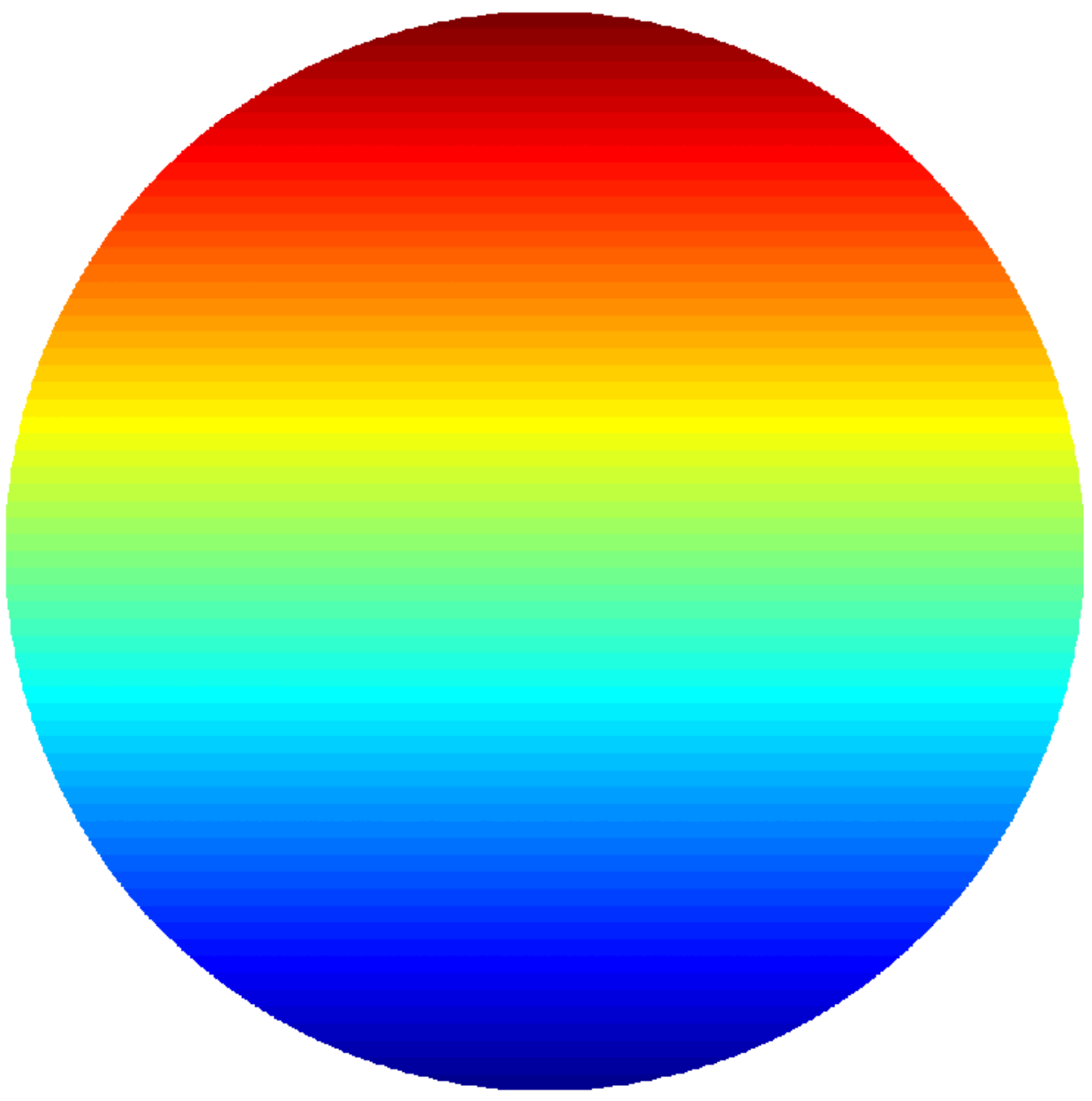} & \includegraphics[height=26pt]{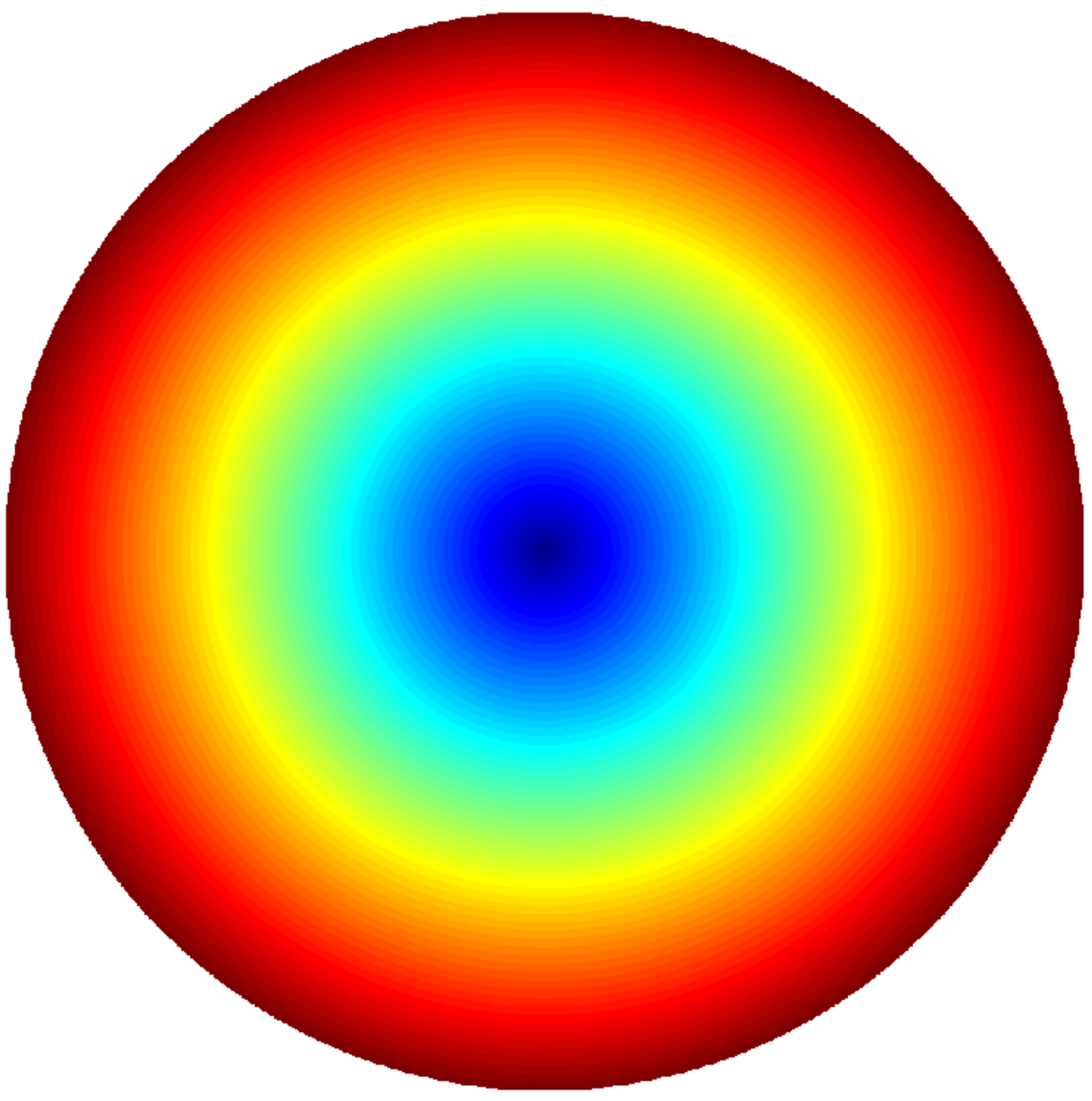} & \includegraphics[height=26pt]{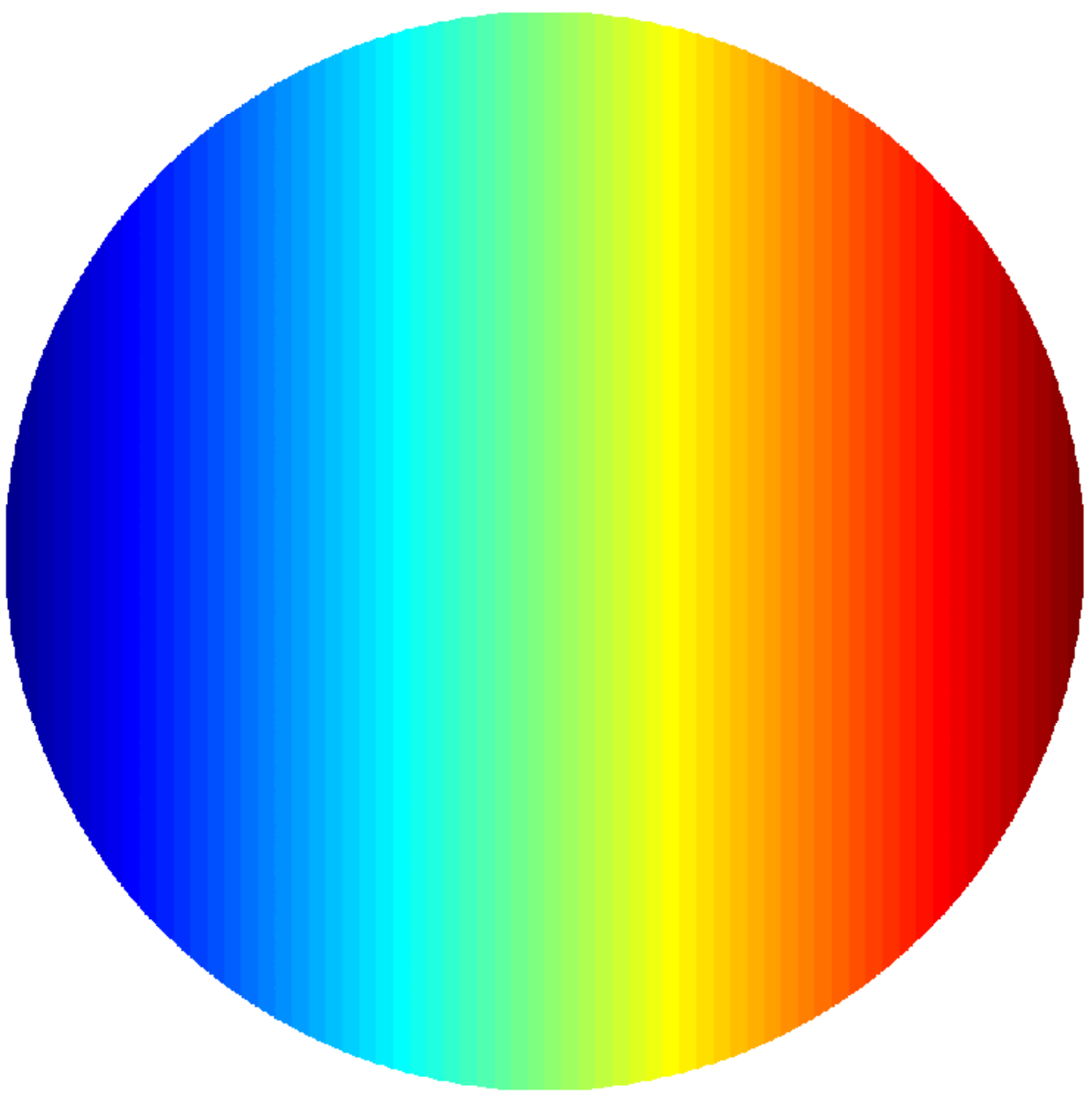} &&&\\
&&$v_{2}^{-2}$&$v_{2}^{-1}$ &$v_{2}^{0}$ & $v_{2}^{1}$ &$v_{2}^{2}$&&\\
&&\includegraphics[height=26pt]{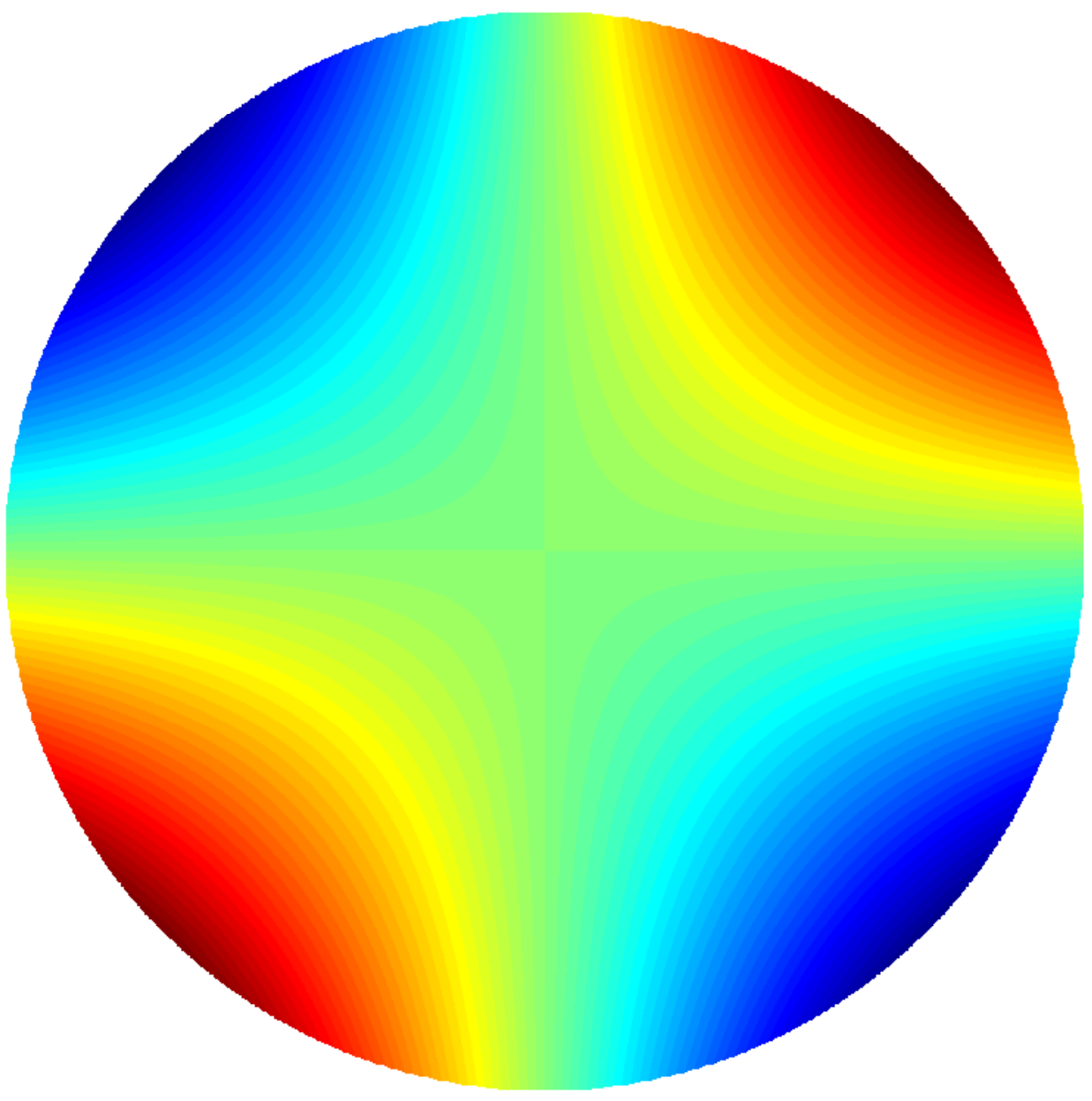}&\includegraphics[height=26pt]{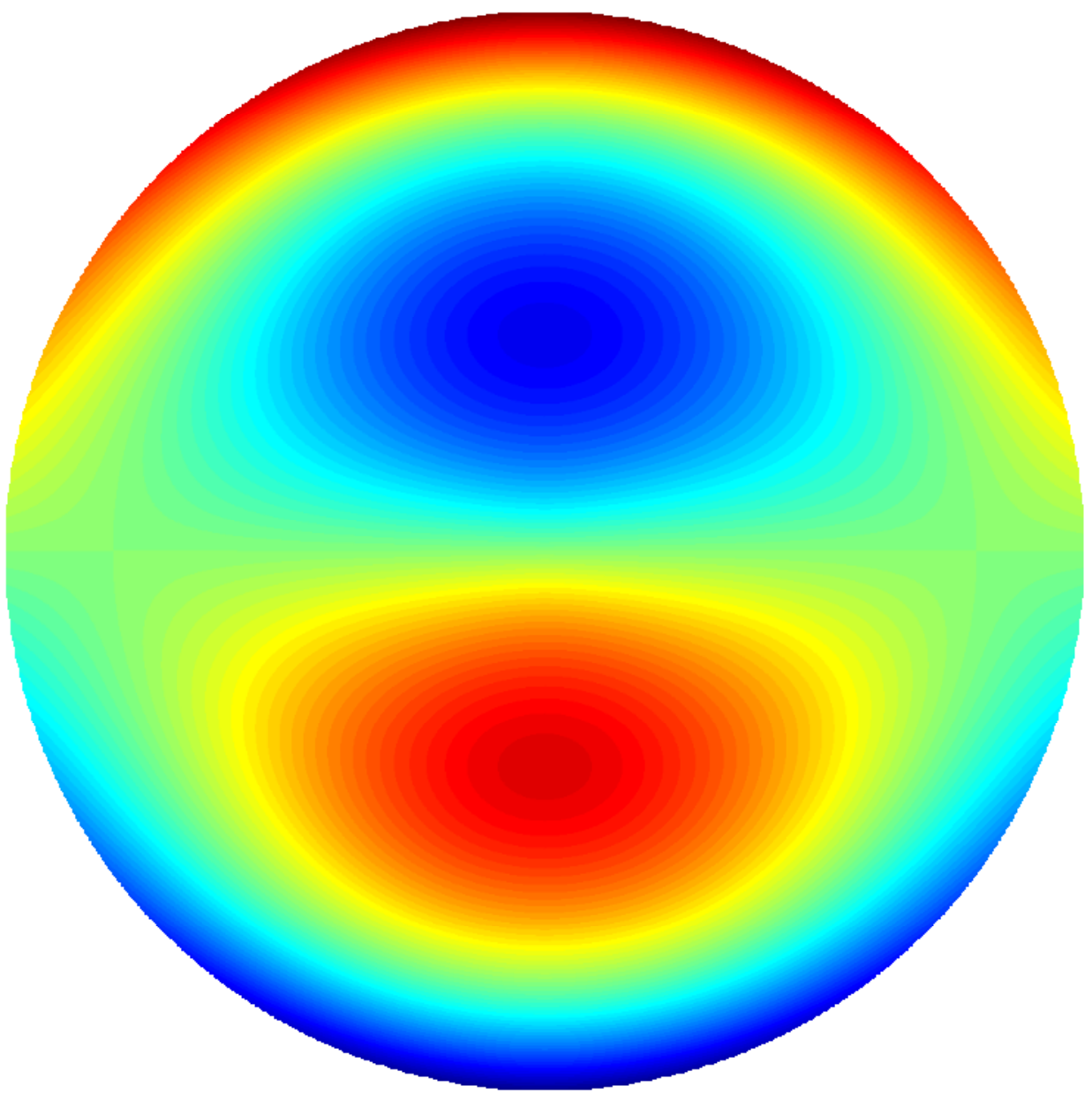} & \includegraphics[height=26pt]{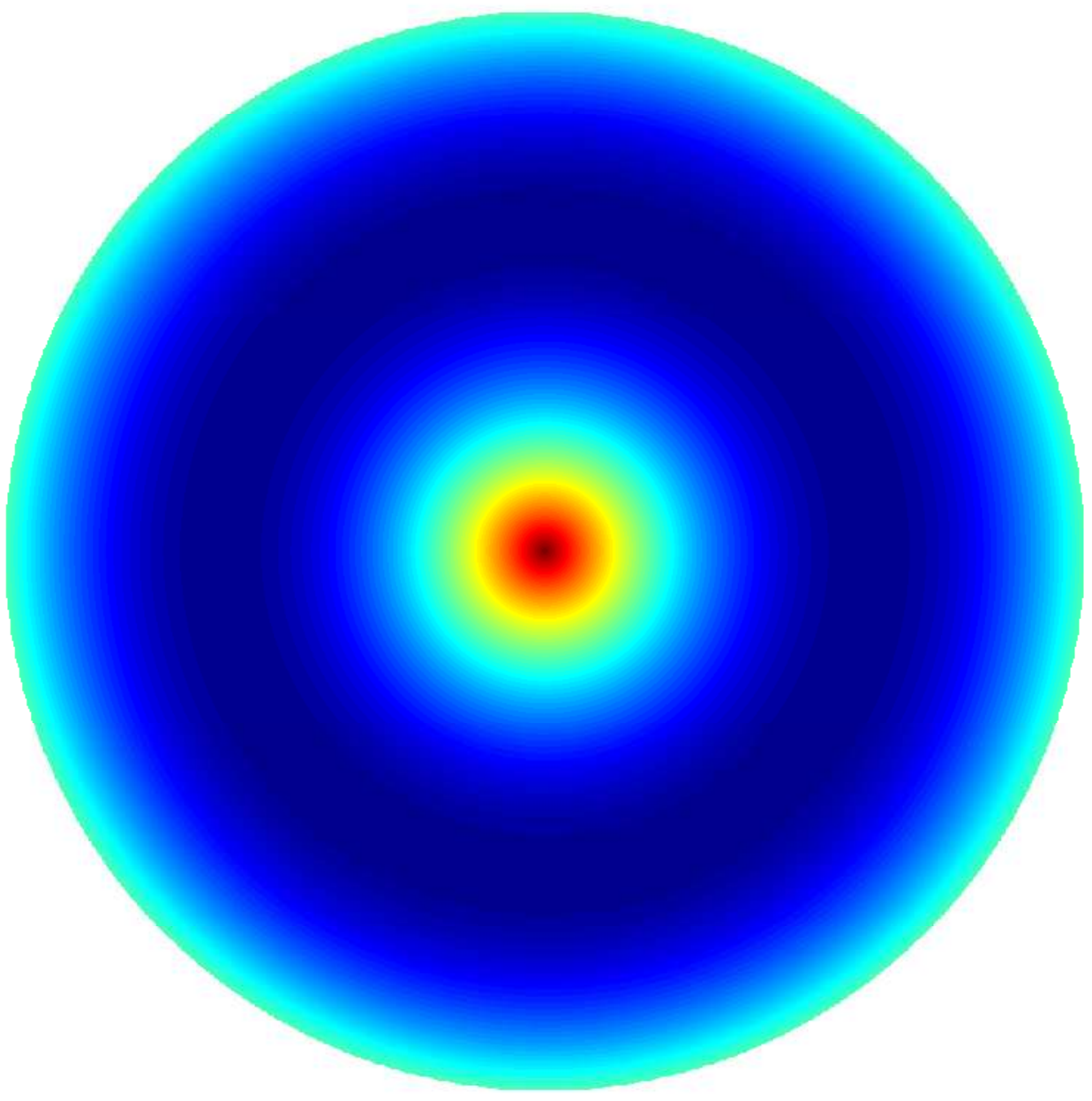} & \includegraphics[height=26pt]{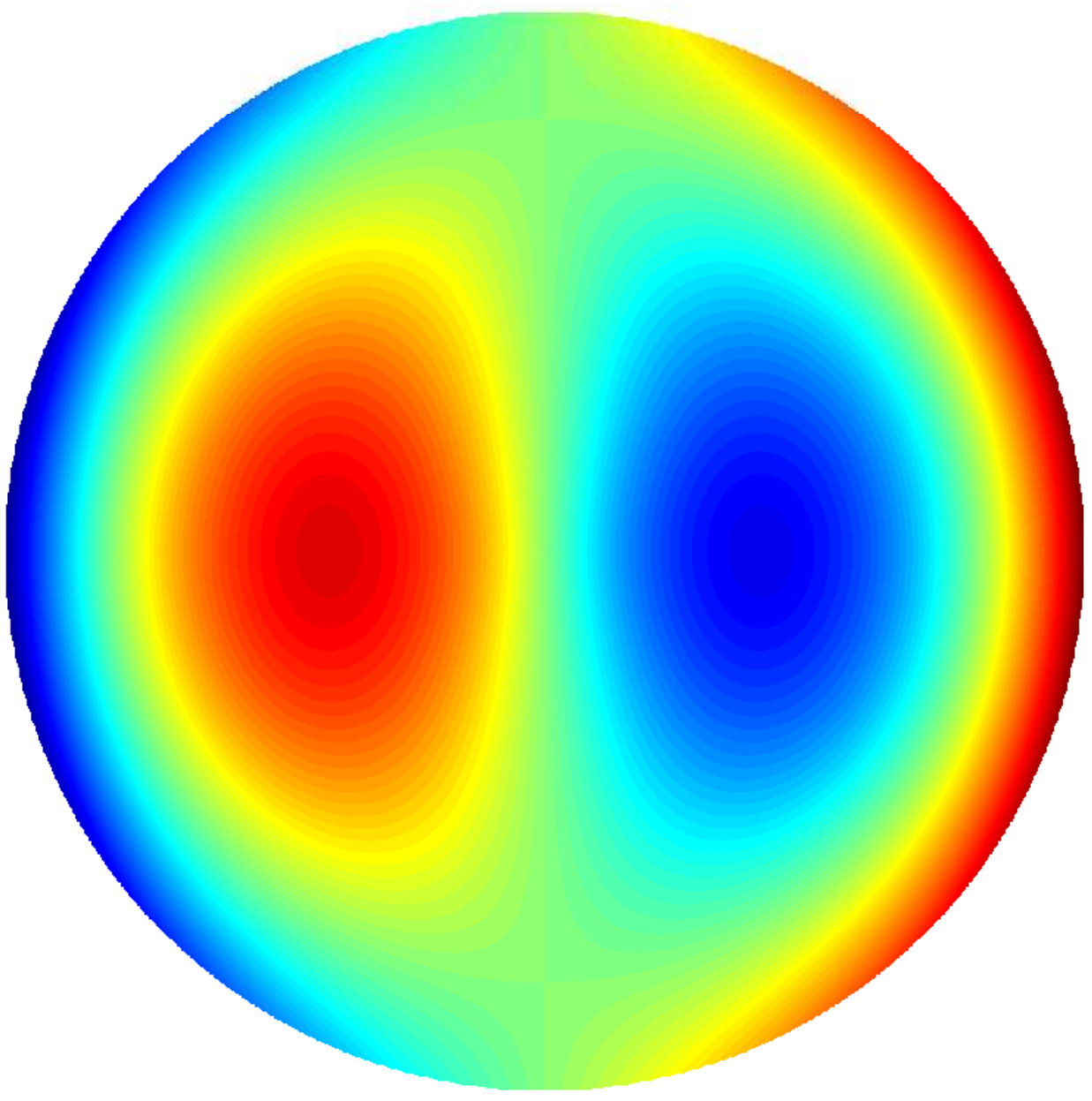} &\includegraphics[height=26pt]{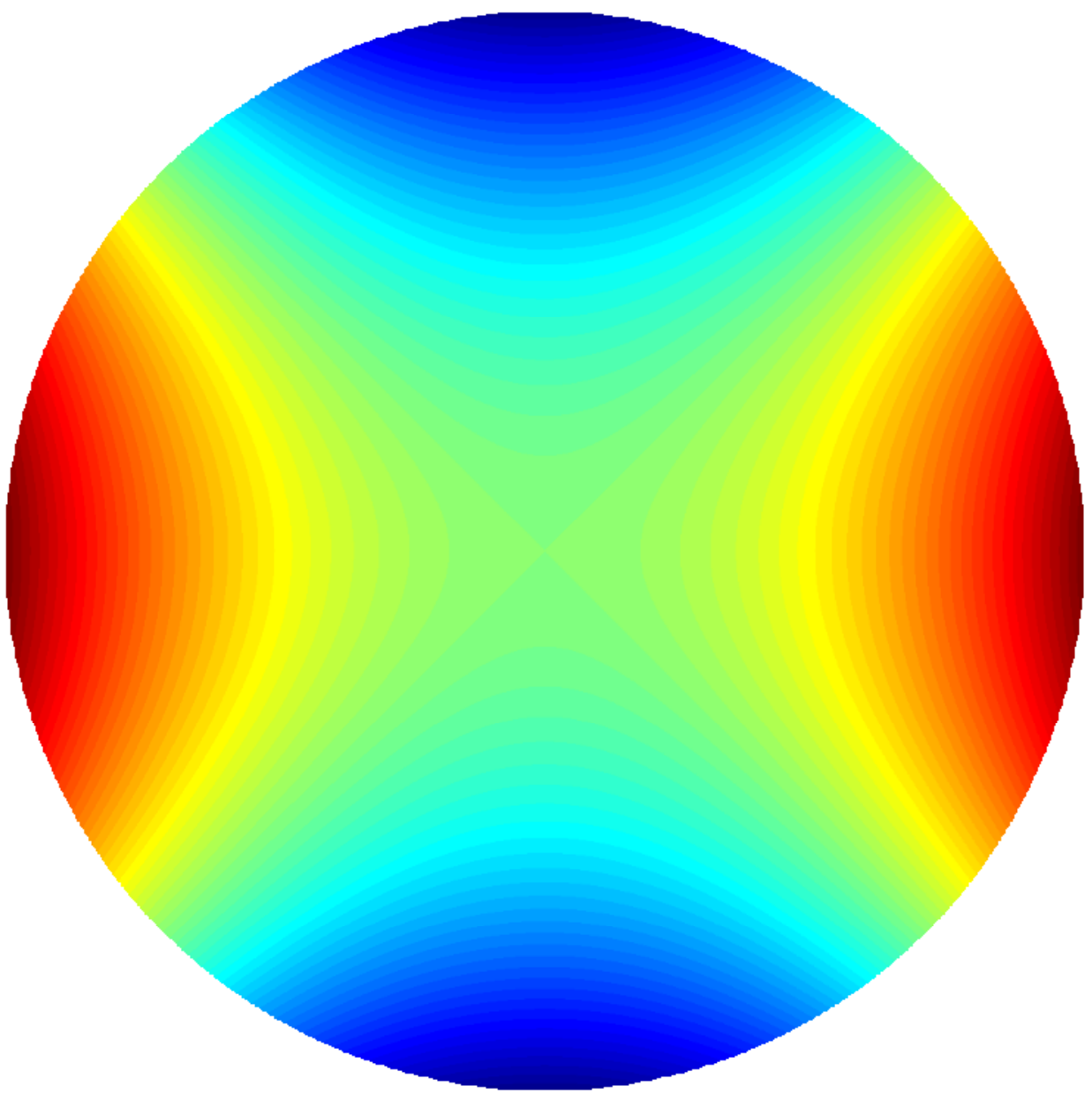}&&\\
&$v_{3}^{-3}$&$v_{3}^{-2}$&$v_{3}^{-1}$ &$v_{3}^{0}$ & $v_{3}^{1}$ &$v_{3}^{2}$&$v_{3}^{3}$&\\
&\includegraphics[height=26pt]{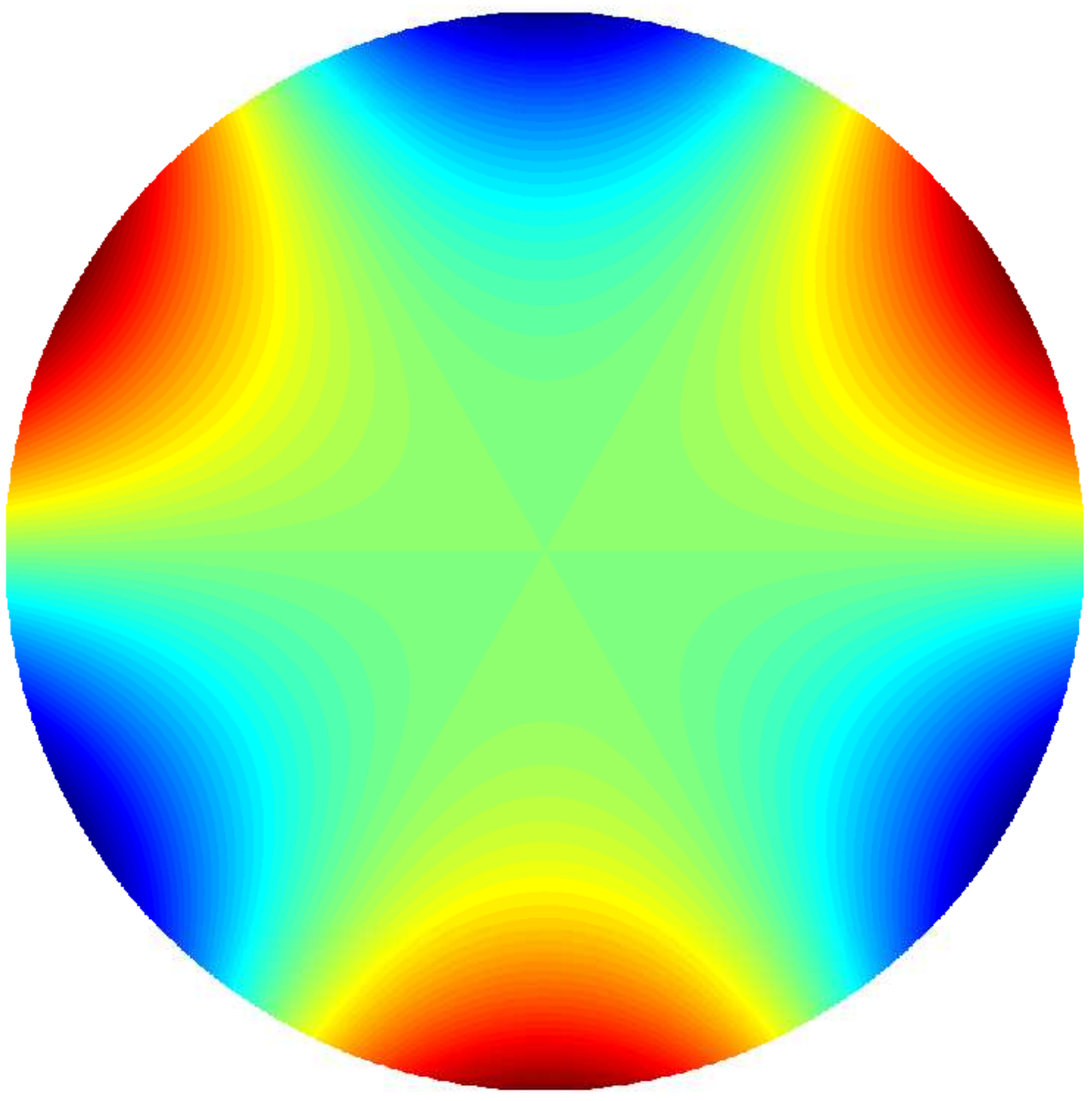}&\includegraphics[height=26pt]{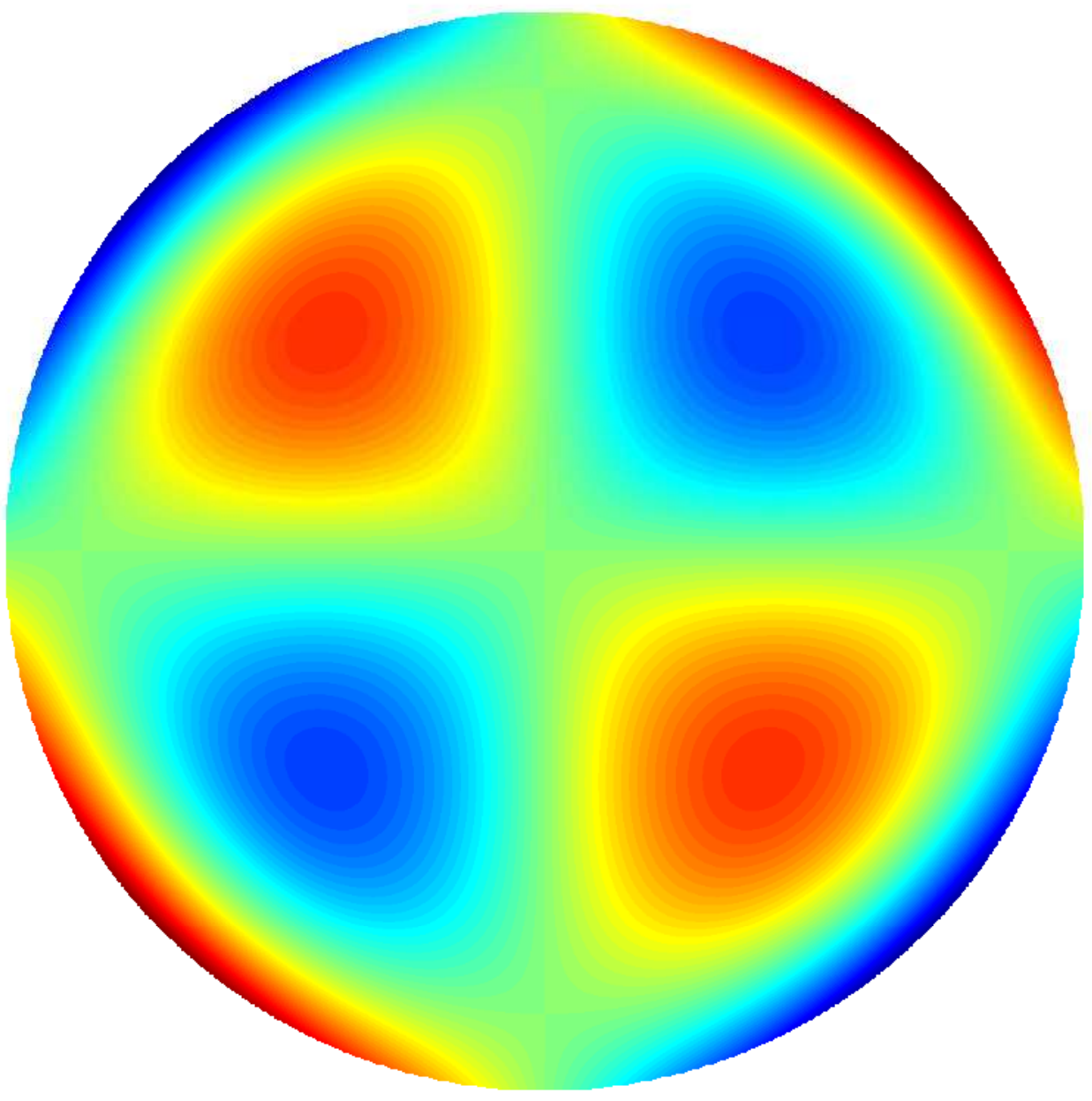}&\includegraphics[height=26pt]{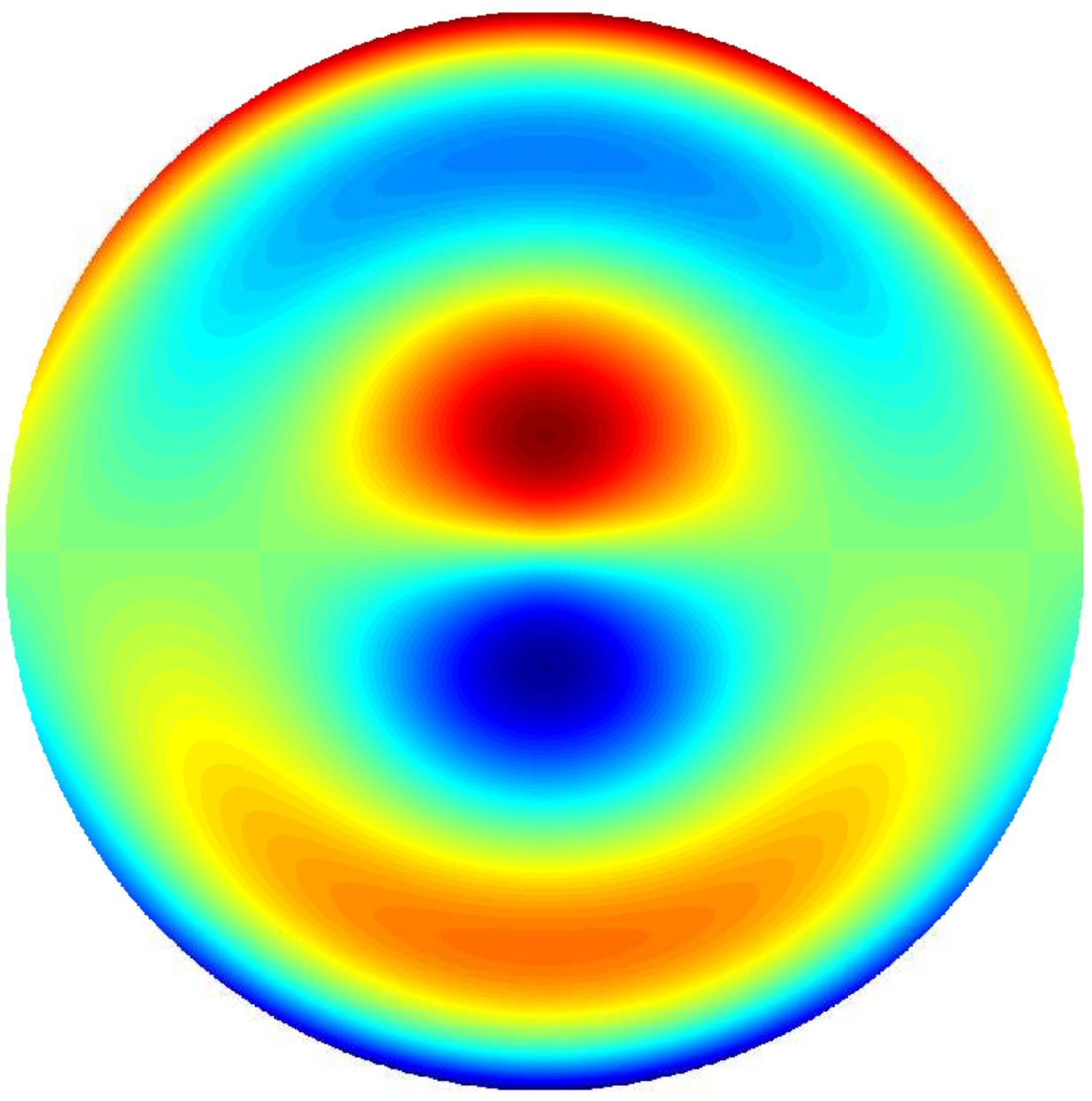} & \includegraphics[height=26pt]{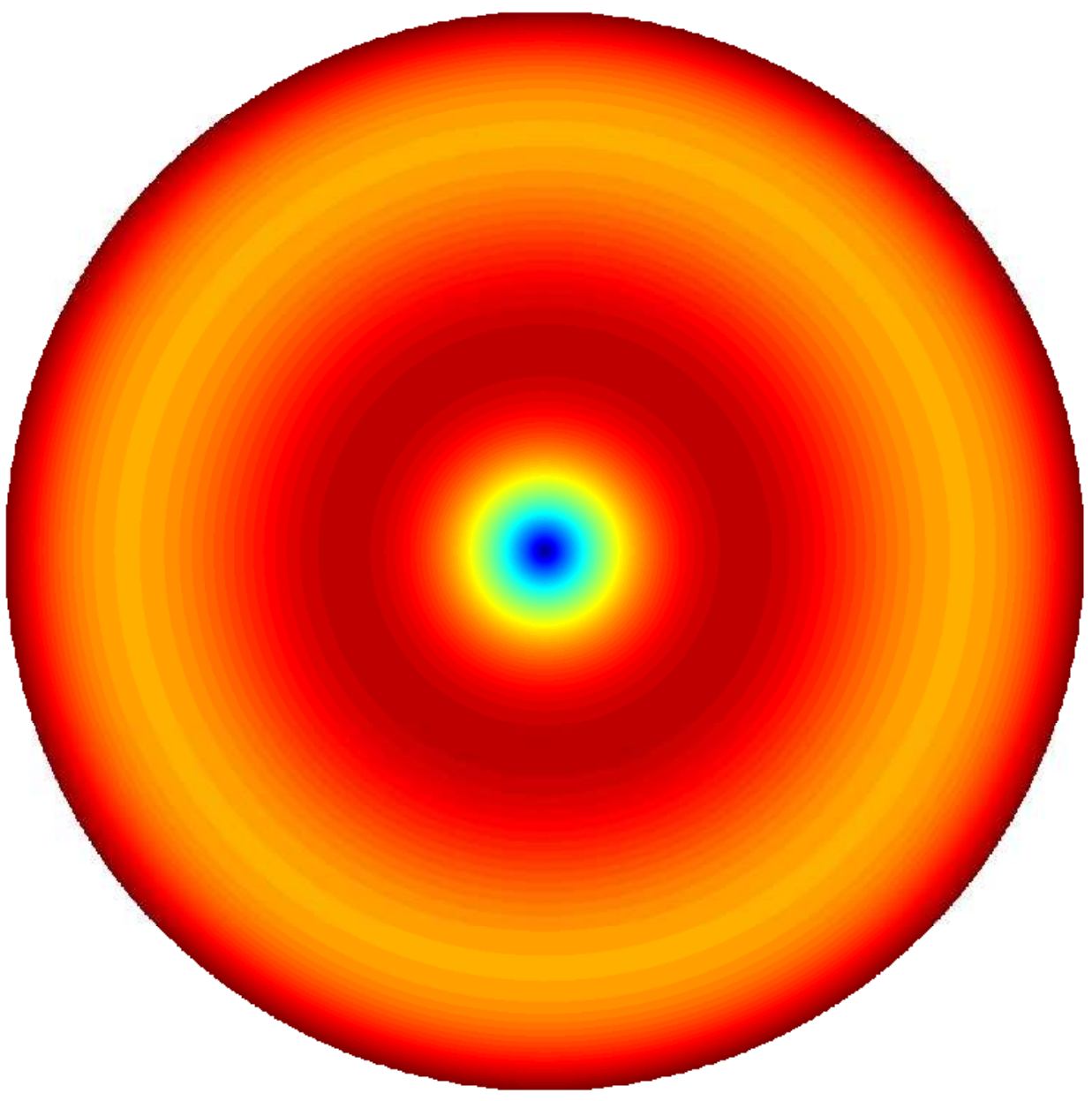} & \includegraphics[height=26pt]{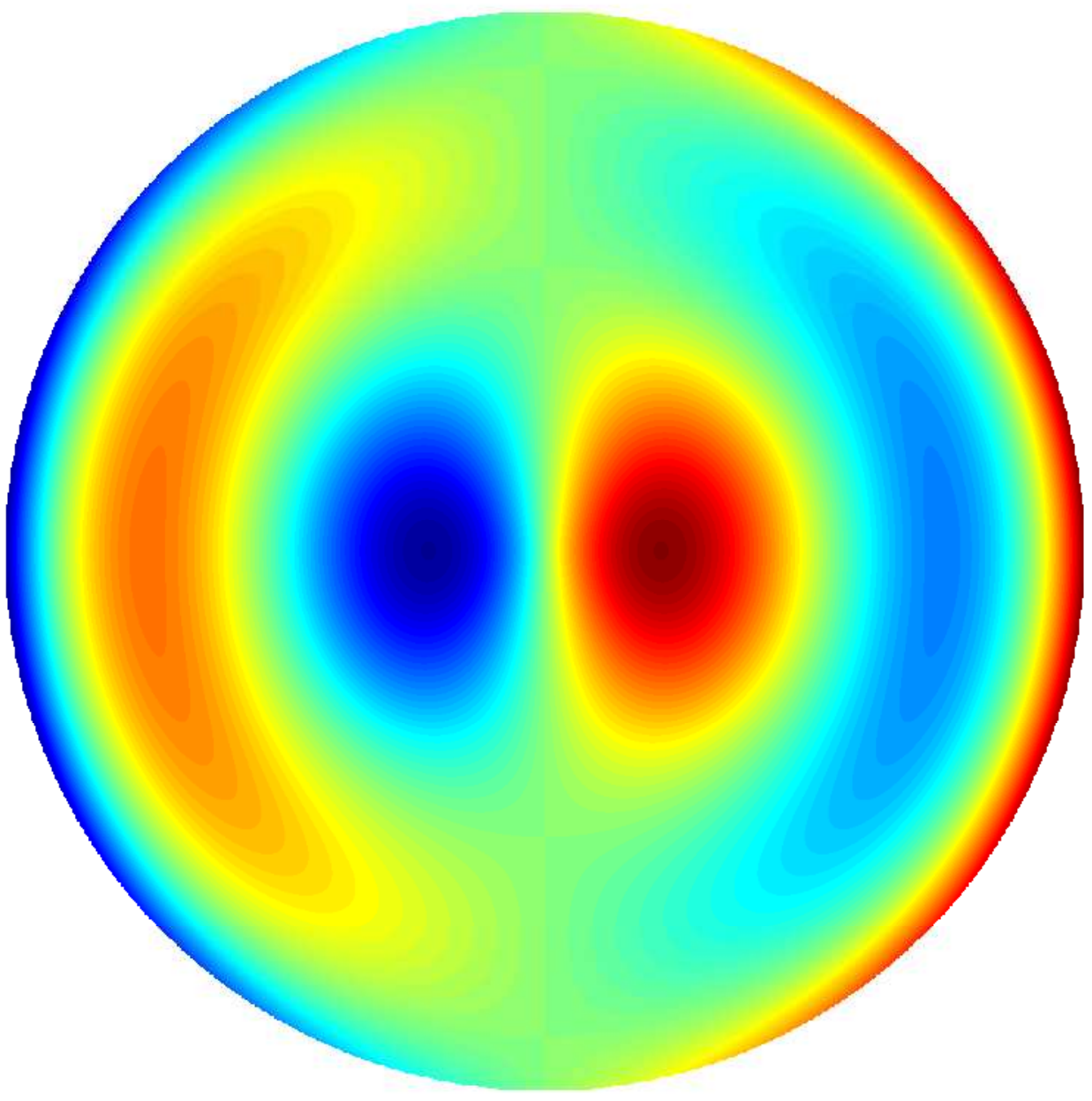} &\includegraphics[height=26pt]{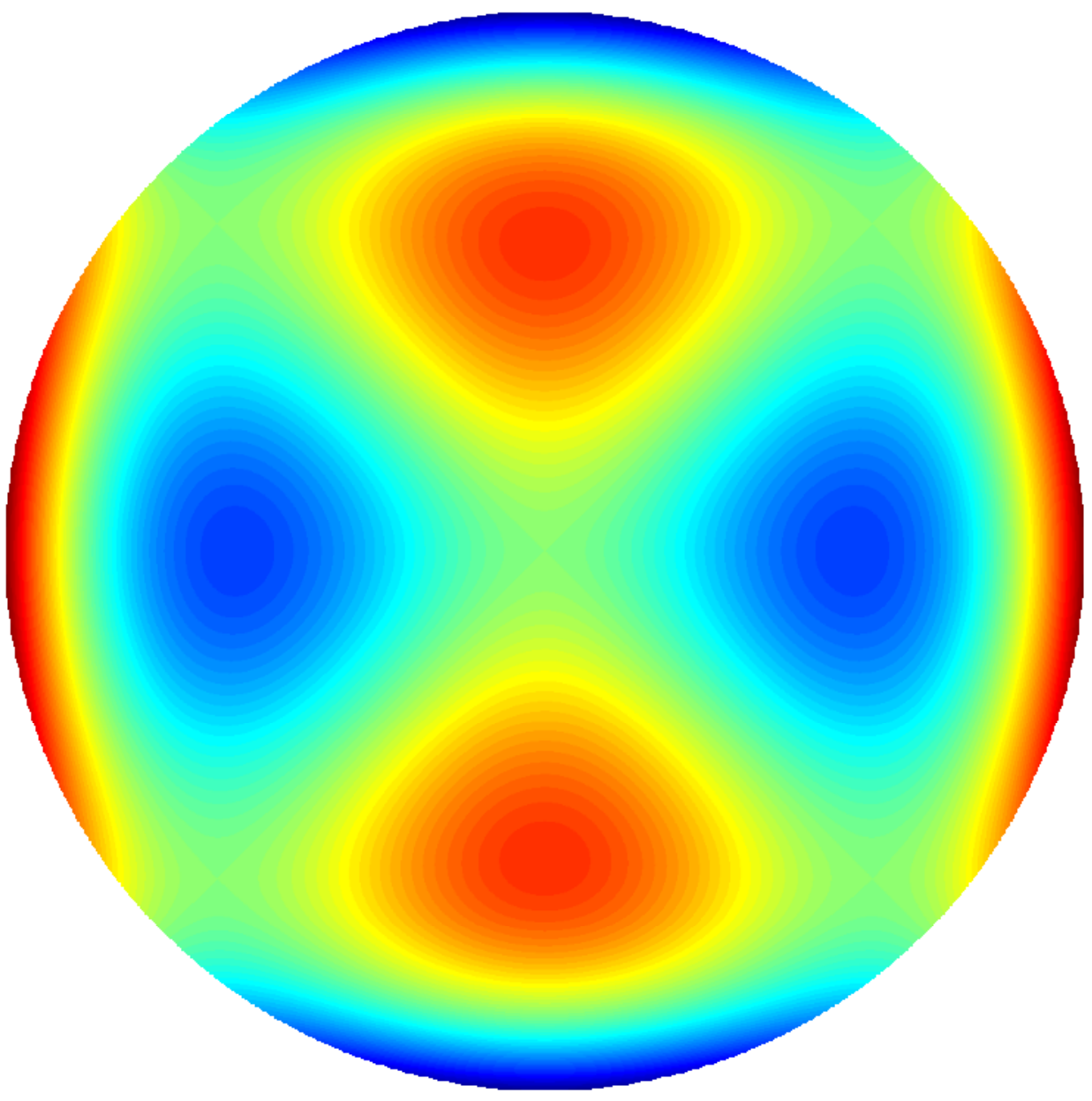}&\includegraphics[height=26pt]{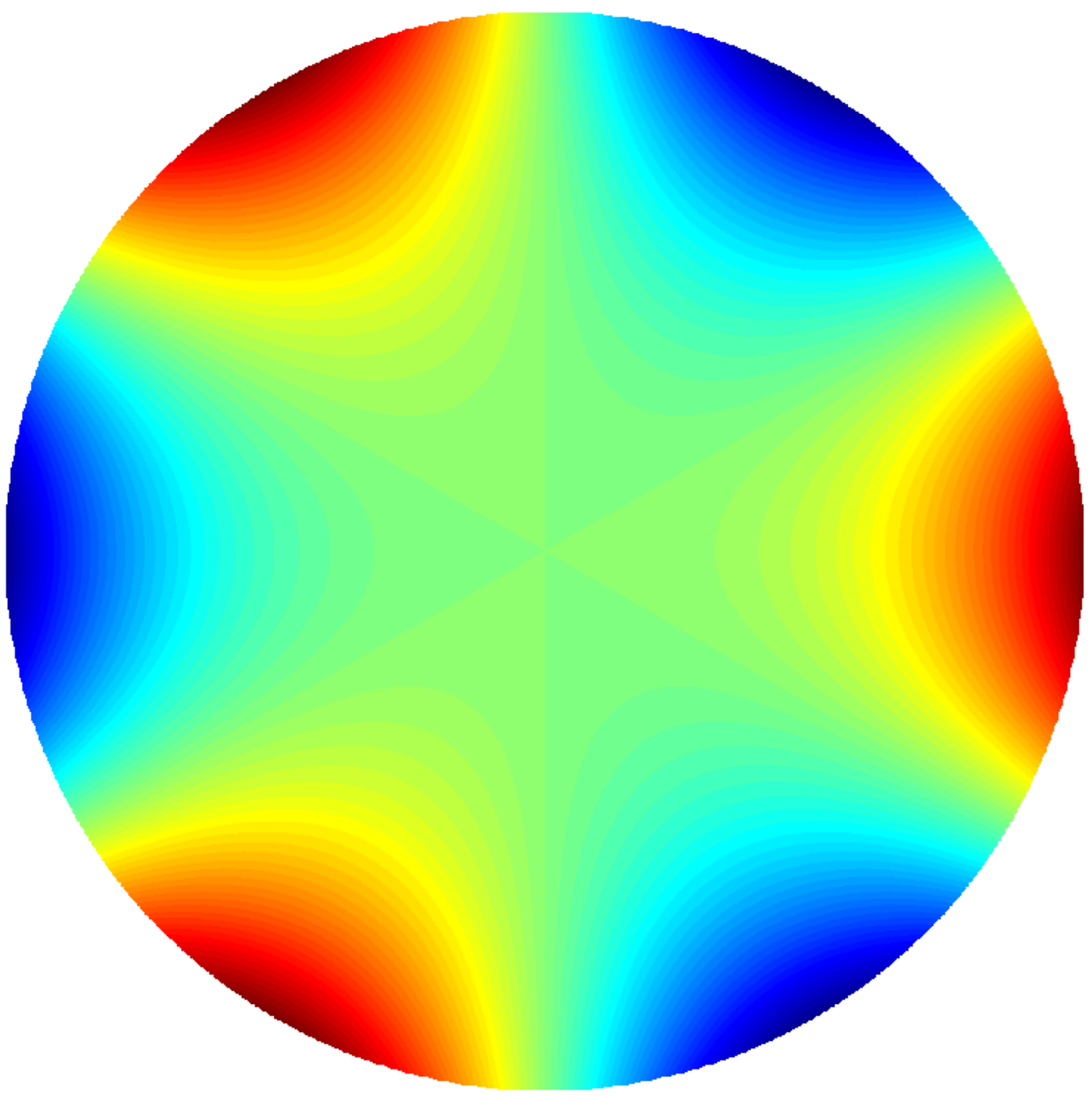}&\\
$v_{4}^{-4}$&$v_{4}^{-3}$&$v_{4}^{-2}$&$v_{4}^{-1}$ &$v_{4}^{0}$ & $v_{4}^{1}$ &$v_{4}^{2}$&$v_{4}^{3}$&$v_{4}^{4}$\\
\includegraphics[height=26pt]{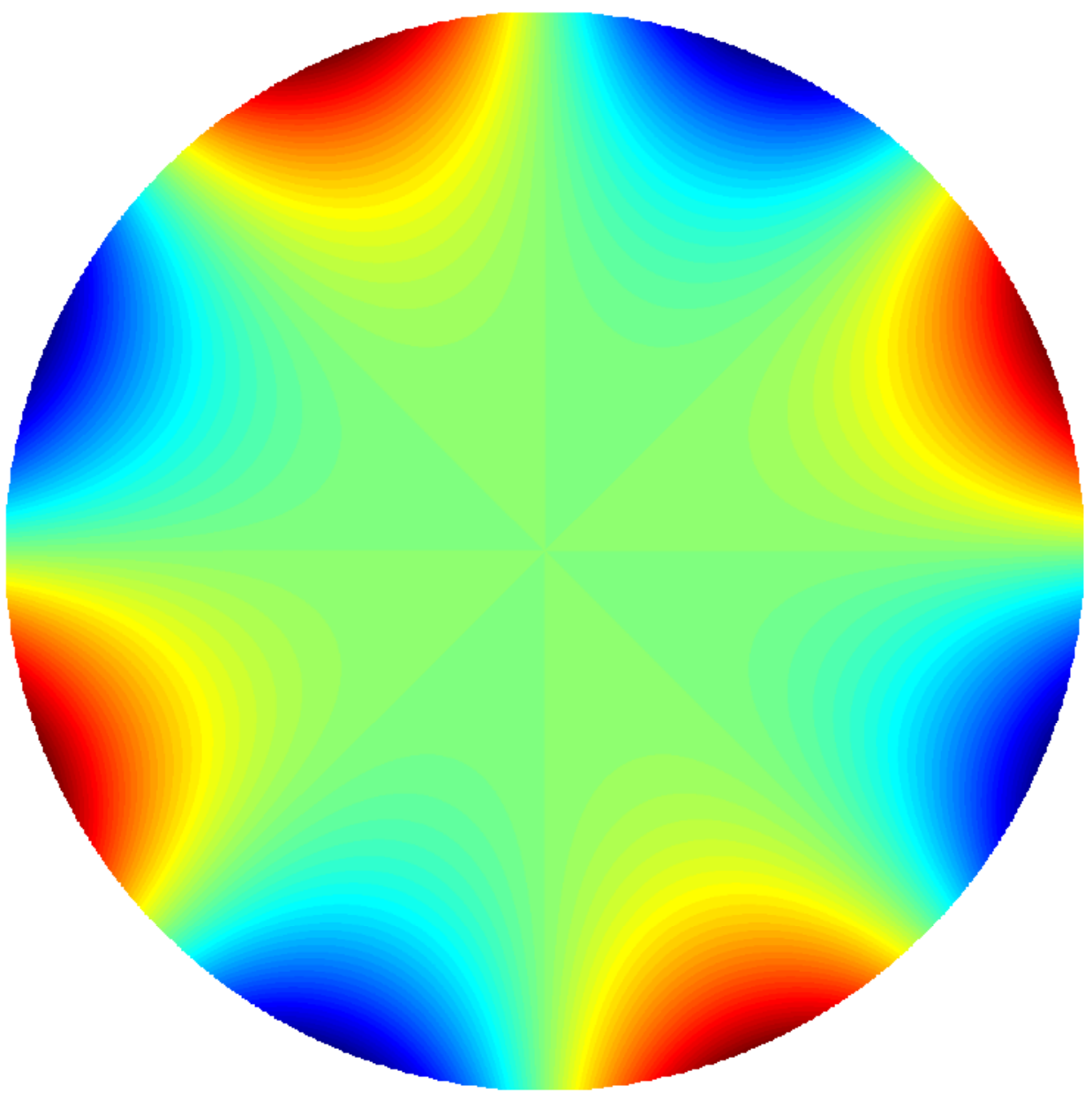}&\includegraphics[height=26pt]{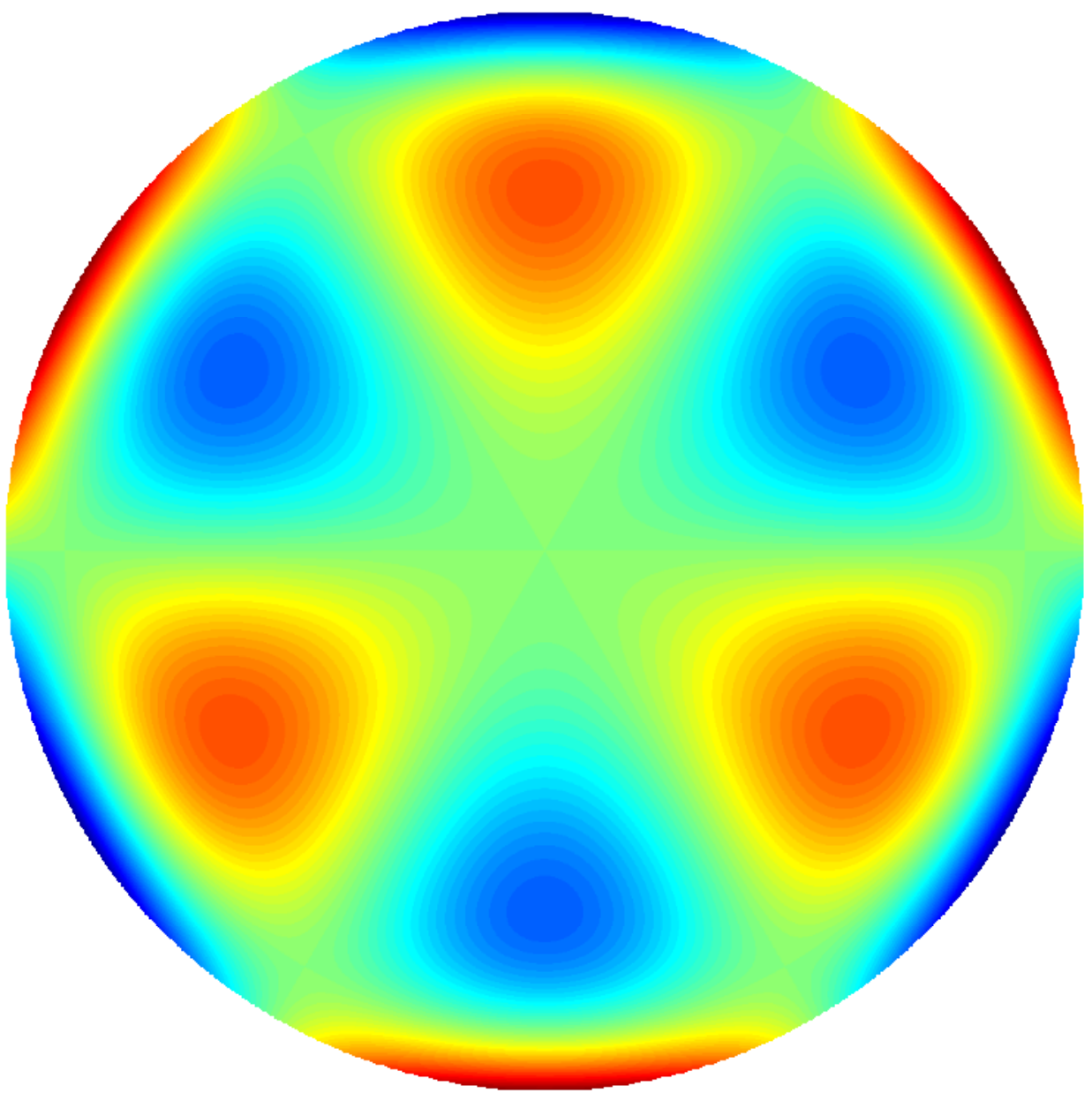}&\includegraphics[height=26pt]{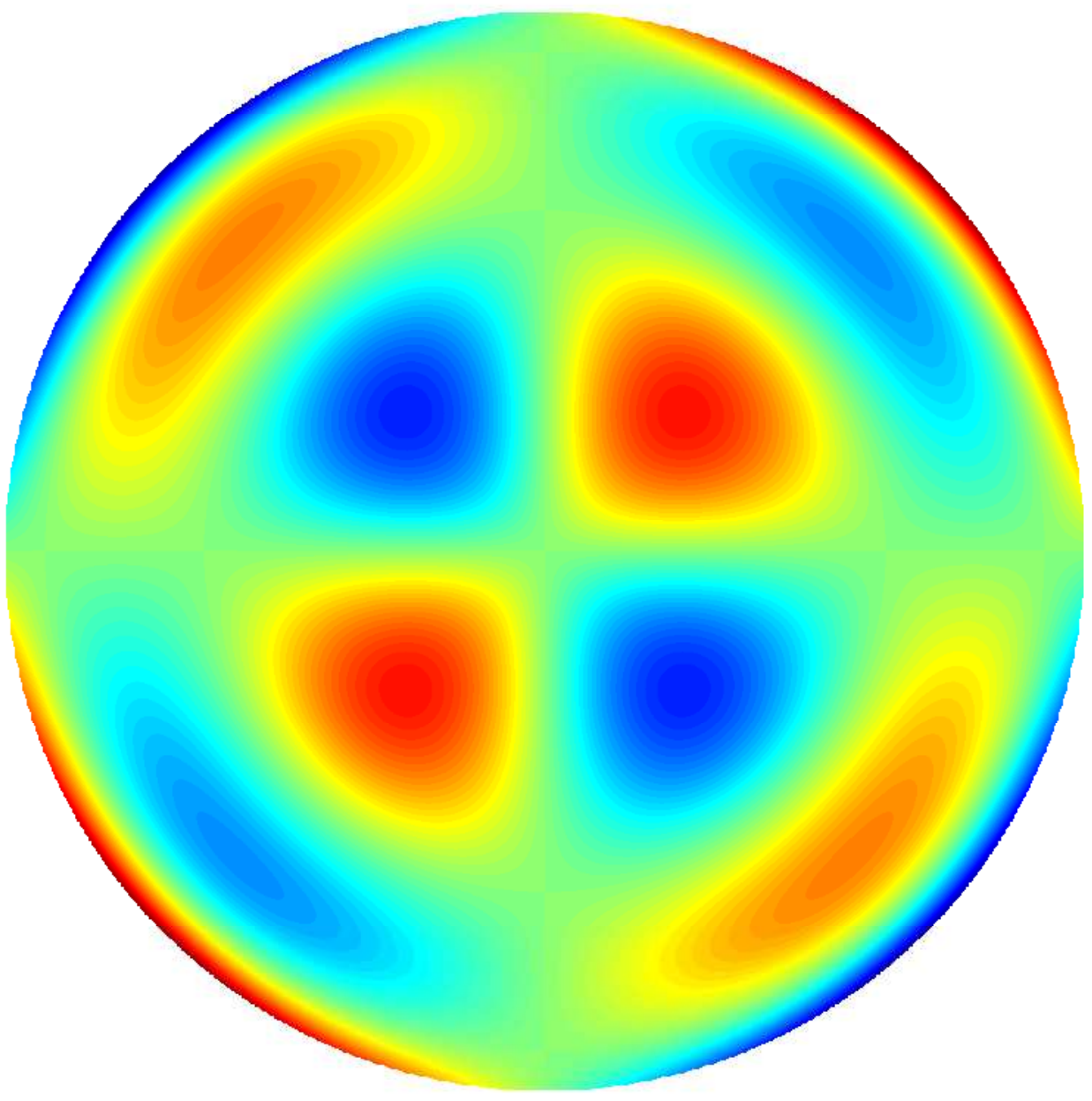}&\includegraphics[height=26pt]{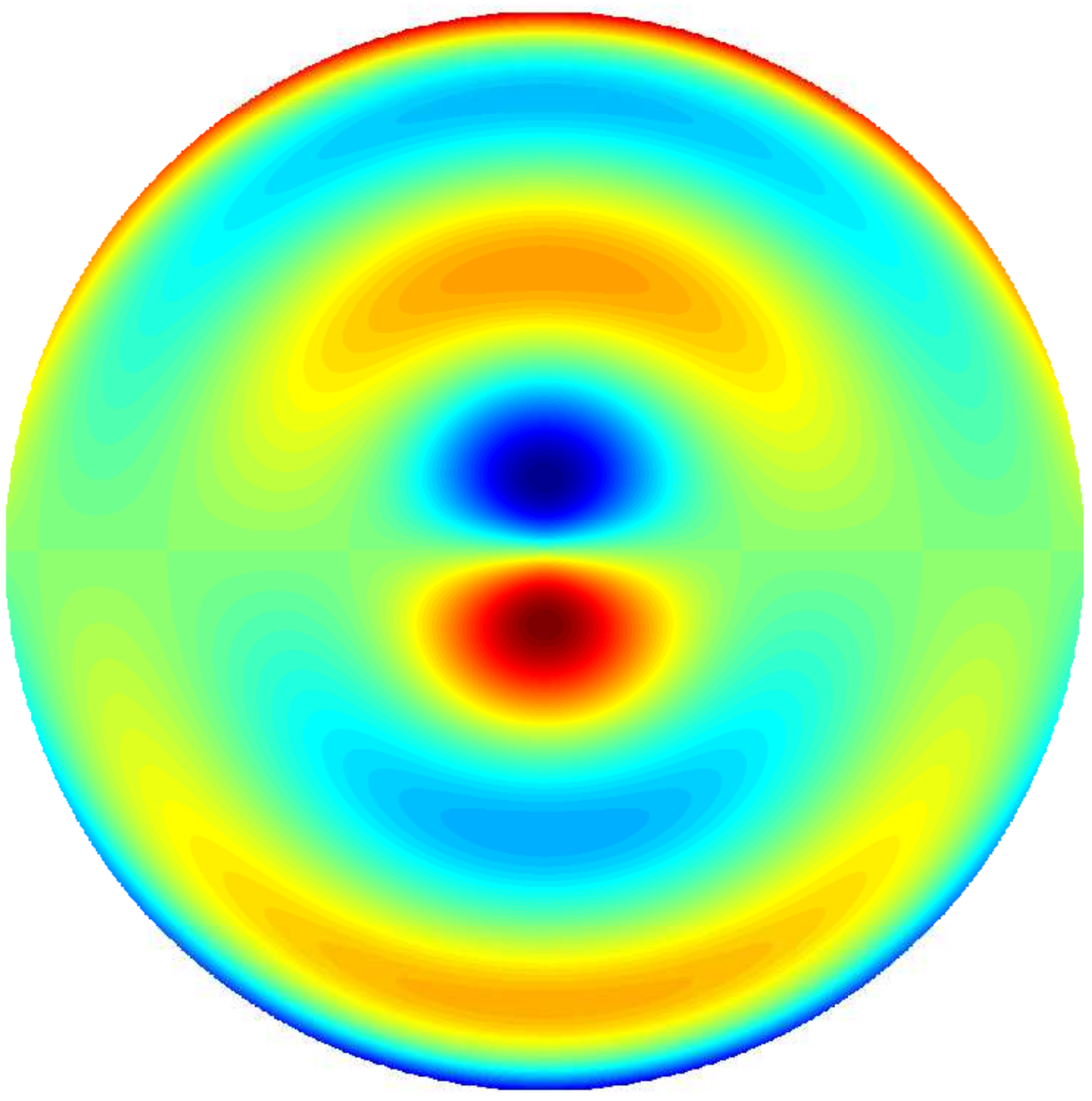} & \includegraphics[height=26pt]{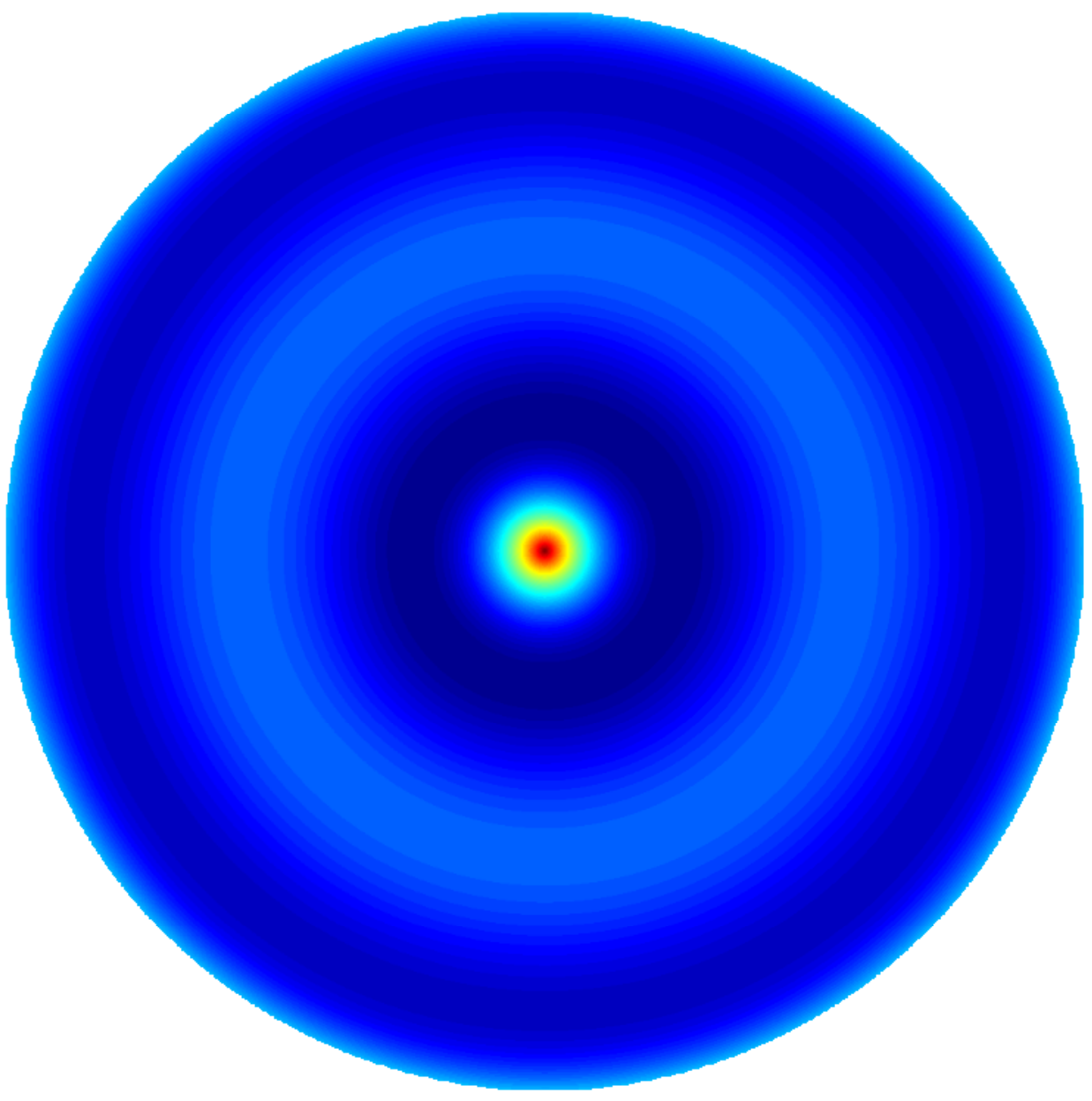} & \includegraphics[height=26pt]{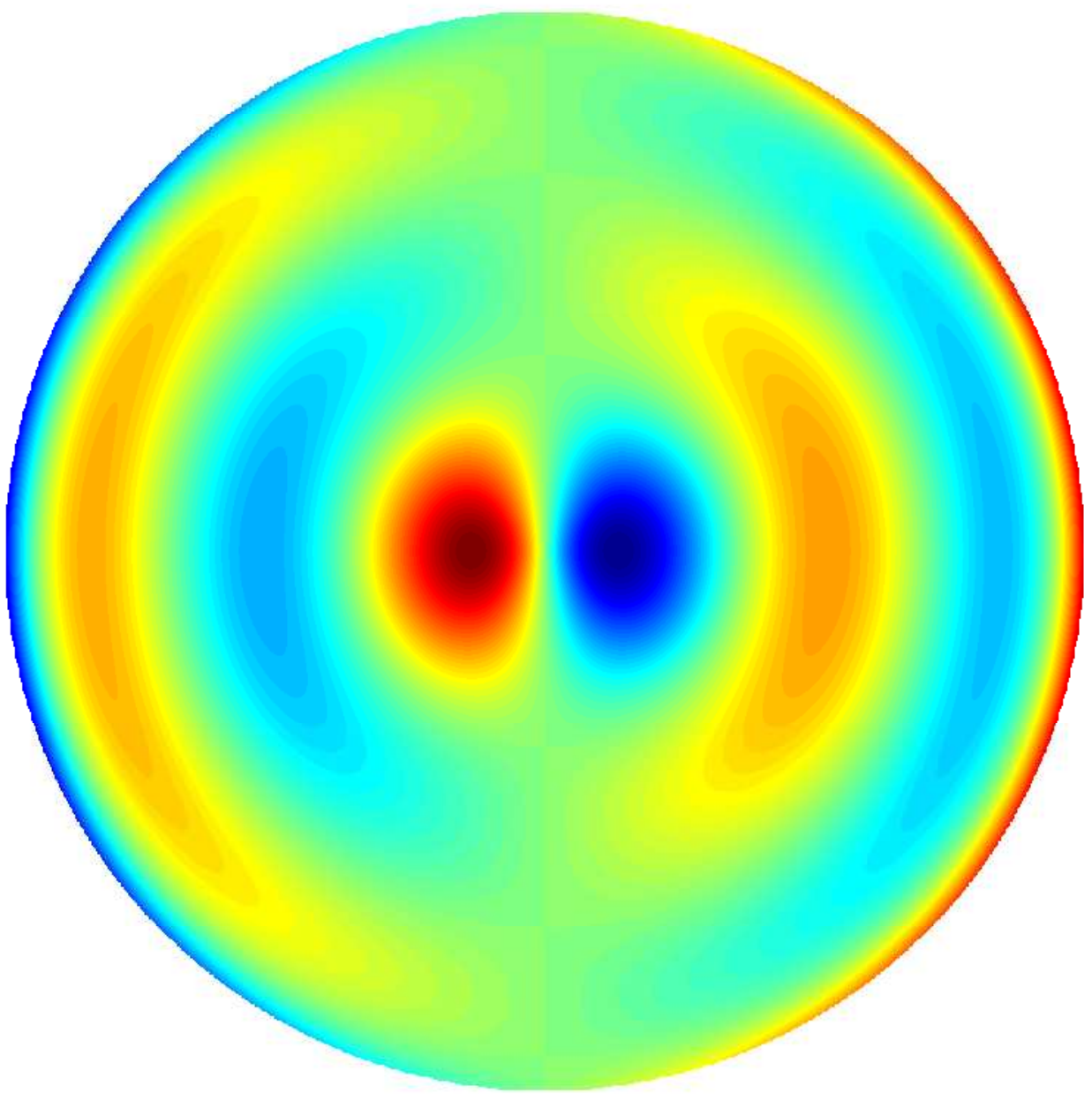} &\includegraphics[height=26pt]{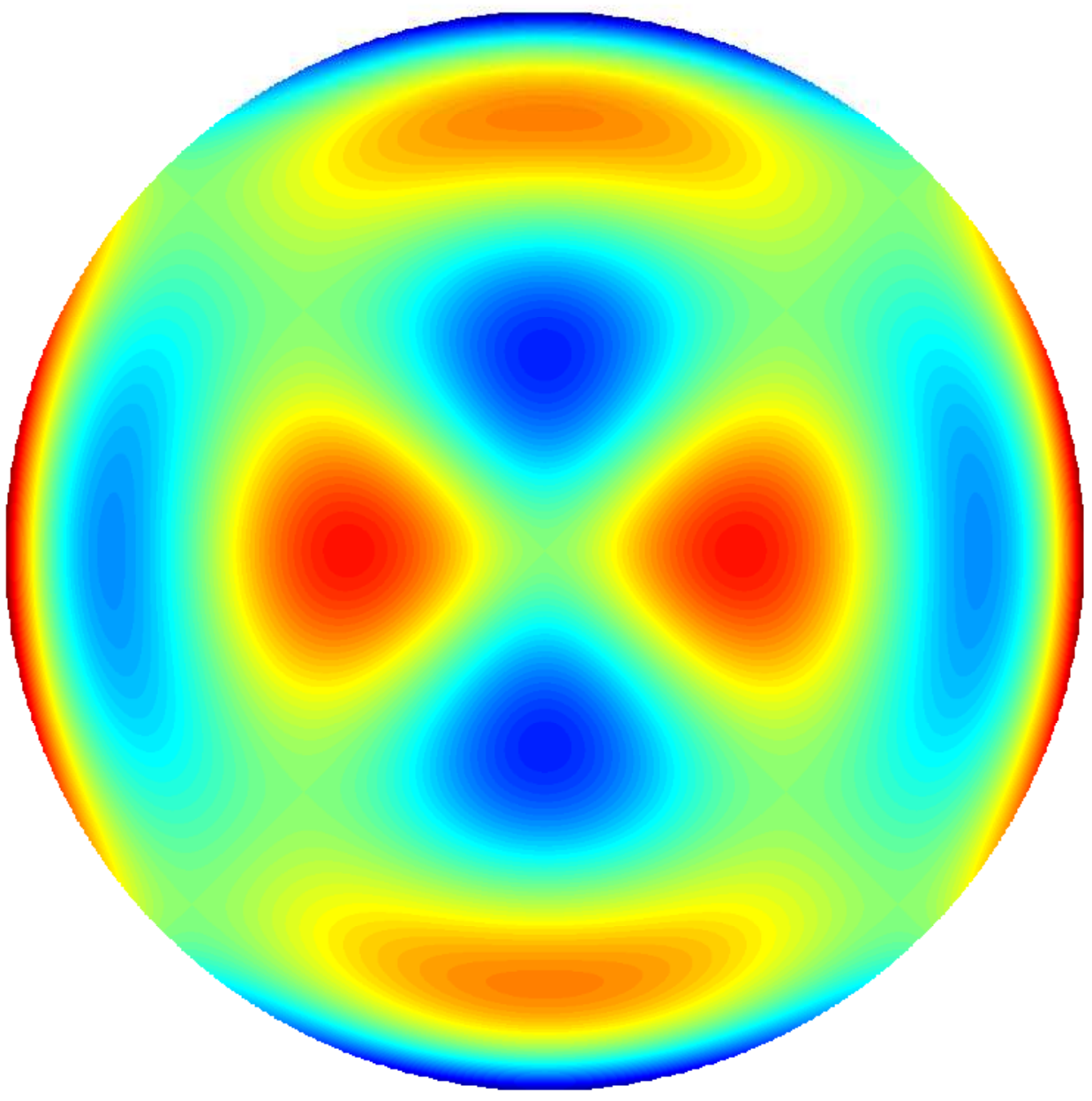}&\includegraphics[height=26pt]{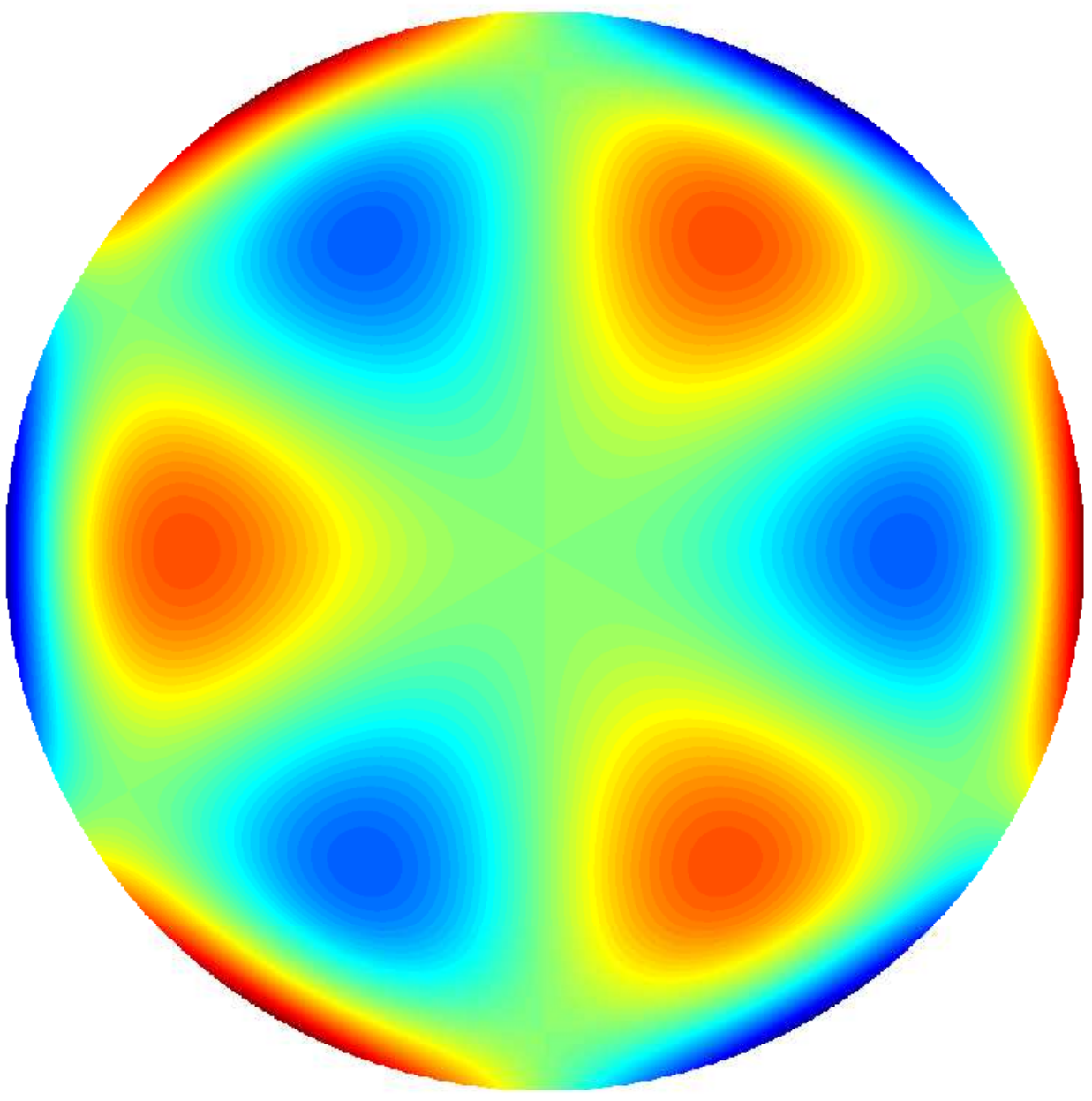}&\includegraphics[height=26pt]{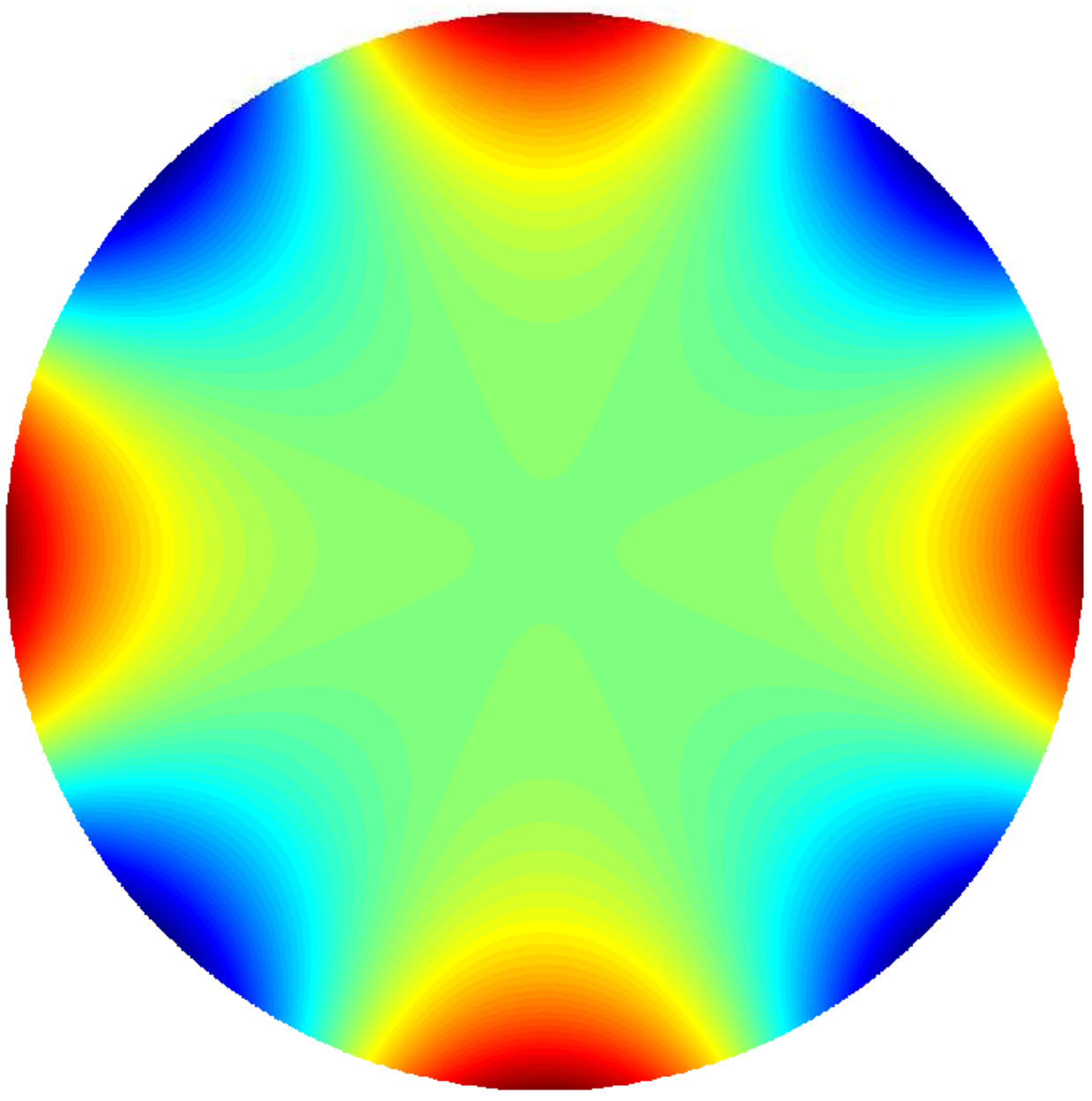}\\

\end{tabular}
\caption{Visualization of pseudo-Zernike polynomials for $n=1\ldots 4$.}
\label{fig:zernfilters}
\end{figure}

Zernike polynomials were originally introduced by Zernike~\cite{Zern34} and are traditionally used in optics and also as shape descriptors~\cite{KH90}\cite{RLB09}.
They form a set of orthogonal basis functions defined on the unit disc. 
We focus on the use of pseudo-Zernike polynomials~\cite{BW54} which have proven to be more robust to image noise~\cite{TC88}.
They are defined as
\begin{equation}
v_n^l= R_{nl}(r)e^{il\theta}, 
\label{equ:zernike}
\end{equation}
with
\begin{equation}
R_{nl}(r) = \sum_{s=0}^{n-|l|}(-1)^s \frac{(2n+1-s)!}{s!(n-|l|-s)!(n+|l|+1-s)!} r^{n-s},
\end{equation}

where $n$ is the order and $l$ is the repetition.
In Figure~\ref{fig:zernfilters} we present pseudo-Zernike polynomials of order 1 up to order 4. Starting from low frequency patterns they end up being more complex patterns and more localized.

We normalize pseudo-Zernike functions to a 2D rectangular patch (patch width equal to $11$ in all our experiments) and convolve the input image using them.
Each pseudo-Zernike function is used as a filter and this response constitutes the interestingness measure for detection of point locations. 
Local maxima and local minima are detected for each filter independently.
We now define the maximum number of patches to be extracted per image as $N_\mathrm{z}$. 
This is used to define the available capacity per filter, in order to retain only the strongest detections.
Capacity is uniformly shared among different filters, and also among maxima and minima. 
We apply the same procedure in $n_\mathrm{\sigma}=5$ scales, similarly to regular sampling. The ratio of capacities of two consecutive scales is $2$. That is equal to the ratio of the down-sampled image areas.
Filter responses are ranked per filter, and point locations are selected until each filter capacity is filled or until there are no more local extrema left.

\begin{figure*}[t]
    \centering
    \begin{tabular}{c c c c}
        \subfigure[$v_1^{-1}$]{\includegraphics[width=0.22\textwidth]{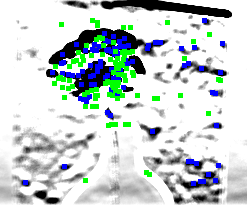}}
        &
        \subfigure[$v_1^{0}$]{\includegraphics[width=0.22\textwidth]{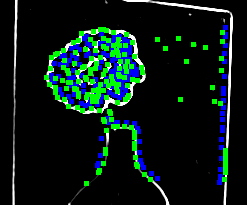}}
        &
        \subfigure[$v_1^{1}$]{\includegraphics[width=0.22\textwidth]{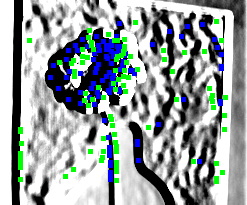}}
        &
        \subfigure[$v_2^{-2}$]{\includegraphics[width=0.22\textwidth]{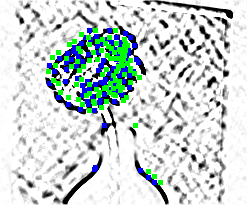}}
        \\
        \subfigure[$v_2^{2}$]{\includegraphics[width=0.22\textwidth]{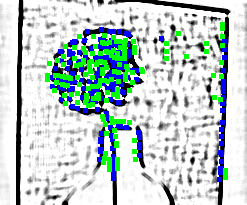}}
        &
        \subfigure[$v_3^{-3}$]{\includegraphics[width=0.22\textwidth]{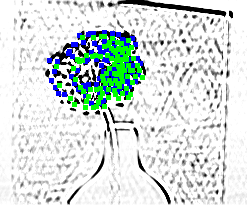}}
        &
        \subfigure[$v_3^{-1}$]{\includegraphics[width=0.22\textwidth]{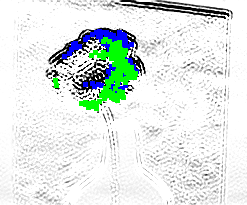}}
        &
        \subfigure[$v_4^{2}$]{\includegraphics[width=0.22\textwidth]{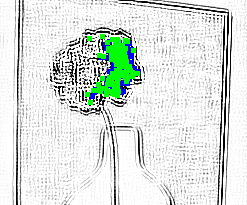}}

    \end{tabular}
\caption{The convolution response for some of the pseudo-Zernike polynomials. The response map is visualized only for the first scale. We use green (blue) patches to visualize the local maxima (minima). \emph{Best viewed in color.}}
\label{fig:zernresponse}
\end{figure*}

We employ $N_\mathrm{f}$ filters, which detect complementary structures as being orthogonal and produce high responses for different structures. 
In contrast to previous feature detectors, we do not focus particularly on detecting corners, blobs or edges and claim that all such structures are useful for image representation. For given capacity, the more filters we use the stronger features we select per filter. 
In Figure~\ref{fig:zernresponse}, we show the responses produced by various filters. Observe, from the responses and from the filters of Figure~\ref{fig:zernfilters}, that lower order polynomials will detect vertical edges, horizontal edges and blobs ($v_1^{-1}$,$v_1^{1}$ and $v_1^{0}$ respectively.), while filters of higher orders will detect more complex image structures.

\subsection{Edge maps}
\label{sec:edges}
We stress that it is advantageous to describe an image by patches centered on its edges. 
We do not focus on solving the edge detection problem, where breakthrough research have been done, such as the recent fast edge detection method based on structured forests~\cite{DZ13}. We use the existing techniques to sample dense points from images.

We initially consider MSER arbitrarily shaped regions to create an edge map. This 2D map is equal 1 on the region borders and 0 everywhere else. We compute the gradient magnitude on edge pixels and use it as interestingness measure. A local maxima selection is performed to select keypoint locations. Note that only edge pixels are selected in this manner.
We center $n_\mathrm{\sigma}$ patches at each detected location, one at each of the multiple scales used.
We typically use the same number of scales employed in regular dense sampling, as reported in Section~\ref{sec:background}. We will refer to this detection approach as MSER-edge.
Comparing MSER to MSER-edge is a way to compare performance based on region description by its interior or by its boundaries.
A similar prior attempt is to describe an MSER region by a set of Local Affine Frames extracted with multiple affine-covariant procedures~\cite{OM06}.

We now consider exactly the same procedure, but applied on the edge map derived from
the selective search algorithm of Uijlings's \etal~\cite{UVGS13}. Their starting point is  Felzenszwalb's segmentation algorithm~\cite{FH04} applied on multiple color spaces. Similar segments are merged in a hierarchical manner, and the resulting segments form candidate object regions. We tune it to produce smaller regions, via parameter $k$, potentially capturing object parts. A super-pixel~\cite{RSSL12} like segmentation is produced in this fashion.

Except for centering patches on selective search region (SSR) borders, we also fit an upright ellipse to each one. We refer to the latter as SSR, and to the former as SSR-edge.

Finally, we consider the fast edge detection algorithm of Doll{\'a}r and Zitnick~\cite{DZ13} and apply the same process. The only difference is that the interestingness measure now comes directly from the edge detection process, that is the edge strength. We further eliminate all pixels with strength less than $\tau$, in order to control the number of patches per image. We refer to this approach as \emph{fast-edge} in our experiments.

\subsection{Dense \l2-norm local maxima}
\label{sec:normdense}

\begin{figure*}
    \centering
    \begin{tabular}{c c c c}
        \subfigure[response map]{\includegraphics[width=0.23\textwidth]{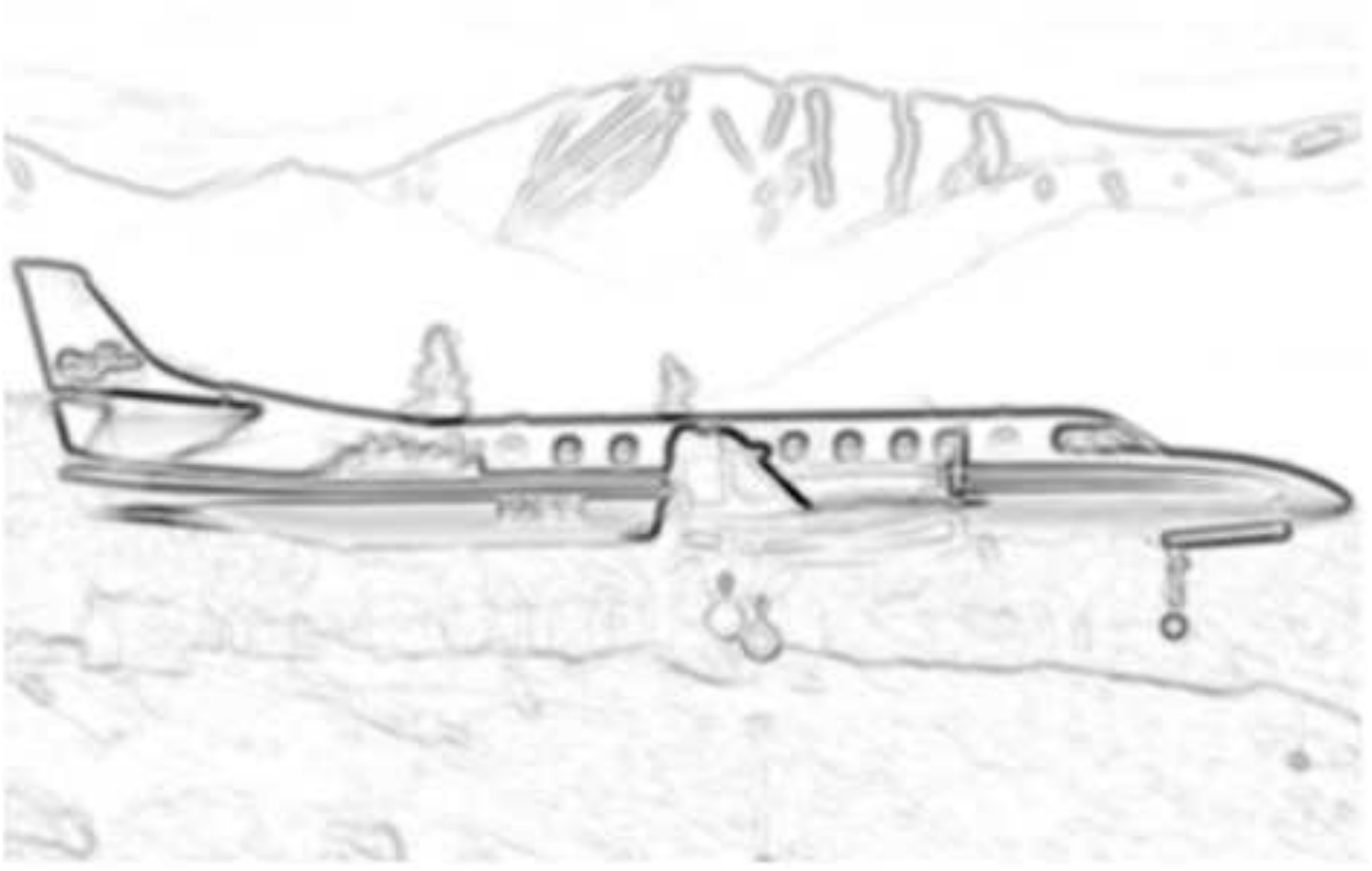}\label{capt:colormap}}
        &
        \subfigure[$\tau=0.1 \cdot 10^4$]{\includegraphics[width=0.23\textwidth]{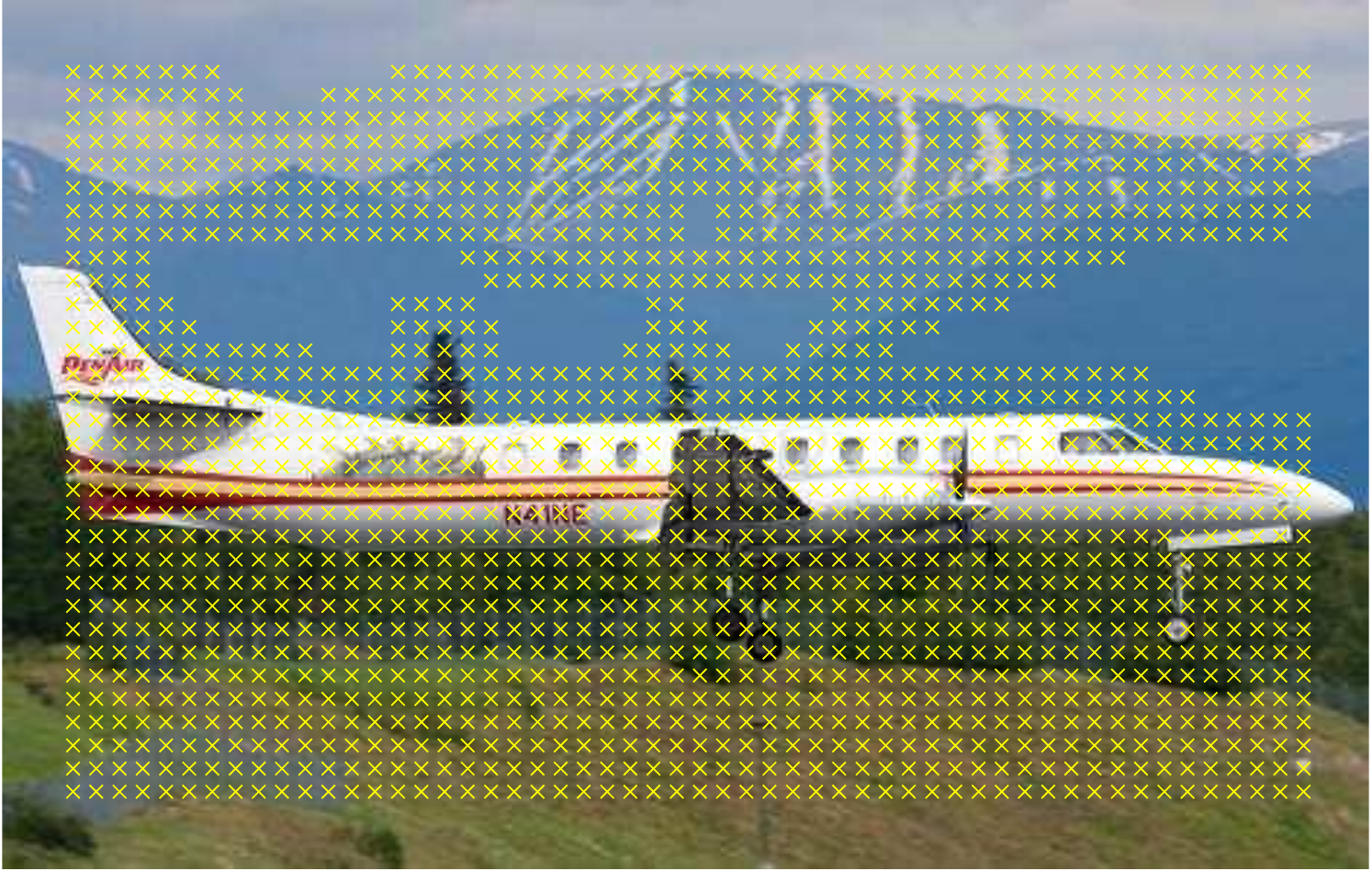}\label{capt:t1000}}
		&
        \subfigure[$\tau=0.5 \cdot 10^4$]{\includegraphics[width=0.23\textwidth]{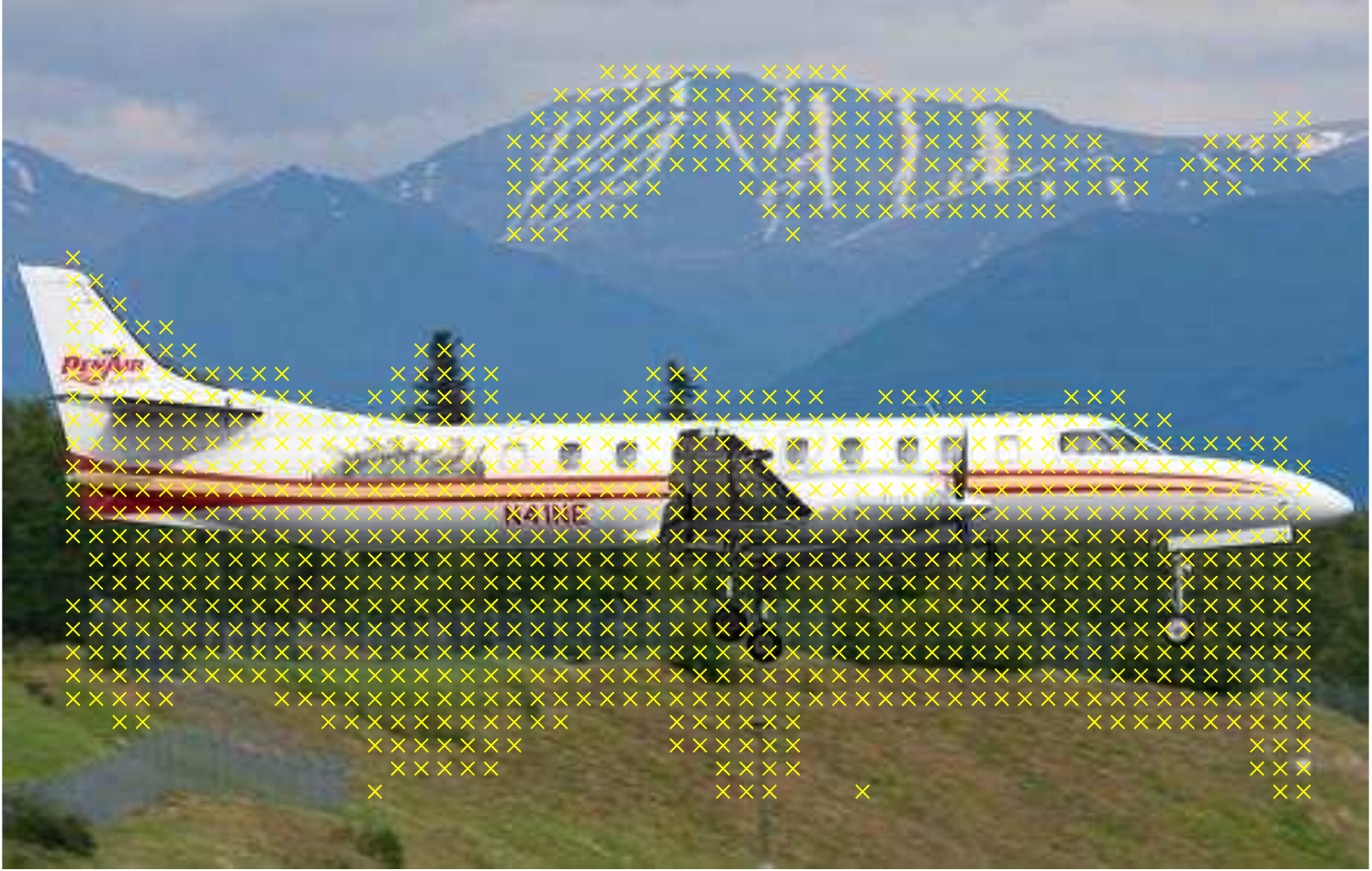}\label{capt:t5000}}
        &
        \subfigure[$\tau=10^4$]{\includegraphics[width=0.23\textwidth]{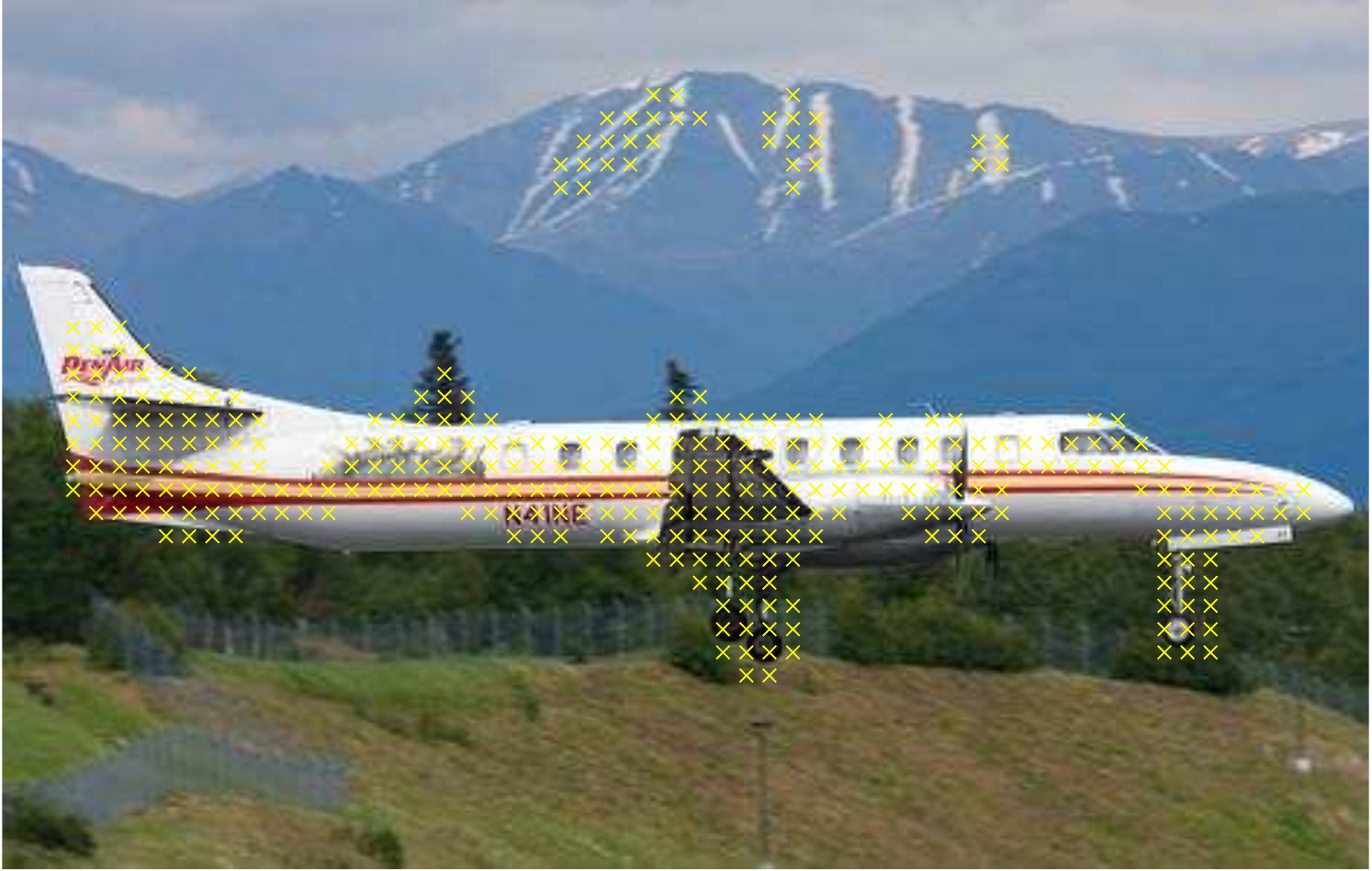}\label{capt:t10000}}
        
    \end{tabular}
\caption{The response map of \l2-norm of SIFT descriptors is shown in \subref{capt:colormap}. Dark pixels correspond to higher responses. Filtering with different threshold values are shown in \subref{capt:t1000}-\subref{capt:t10000}. We use the patch size of $41 \times 41$ pixels.}
\label{fig:l2filter}
\end{figure*}

Gosselin~\etal~\cite{GMJP14} show that the classification performance increases when SIFT vectors with low energy are filtered out. 
This reasoning is supported visually, as these features correspond to homogeneous regions with very little visual information.
However, finding a threshold value is not a feasible operation as it requires cross validation. The optimal threshold value may vary depending on the image and the size of sampled patches. 
The effect of this filtering strategy on regular dense is shown in Figure~\ref{fig:l2filter}, where the descriptors whose squared \l2-norm is lower than the threshold $\tau$ are removed. 

It appears that SIFT vectors and in particular their \l2-norm provide an interestingness information subsequent at feature detection.
In contrast to traditional detectors that localize interest points by low-level statistics, such as gradient changes, we propose to use the \l2-norm of each SIFT as an interestingness measure. 
Initially, we compute SIFT on a very dense uniform grid of step $\delta_{xy}=1$ and at 5 different scales.
The \l2-norm of SIFT descriptors constitutes a response map. That is the interestingness value per pixel.
Following the standard detector procedure, we select the local maxima of this response map.
Similar to interest point detectors, we introduce a threshold parameter $\tau$, to only retain the stronger points. 

As a consequence, patches from homogeneous regions are discarded and the ones with highest energy of SIFT descriptors and more structured regions are retained. 
As shown in Figure~\ref{fig:detectors} the retained points are mostly localized along edges and corners. 
It is the only proposed method that is based on the local descriptor to compute the interestingness measure.

%% file: experiments.tex
\section{Experiments}
\label{sec:experiments}

\begin{table}
\begin{center}
\begin{tabular}{|@{\ssp}l@{\ssp}|@{\ssp}c@{\ssp}|}
\hline
Detector & Section \\
\hline \hline
Dense & Section~\ref{sec:densesample} \\
Dense Interest Points (Dense-IP)~\cite{Tu10}& Section~\ref{sec:densesample} \\
Dense \l2-norm & Section~\ref{sec:normdense} \\
Harris-Laplace~\cite{MiS04}, +relaxed & Sections~\ref{sec:intpoints},~\ref{sec:harlapext} \\
Frobenius, +relaxed & Section~\ref{sec:harlapext} \\
Hessian Affine (HesAff)~\cite{MTSZMSKG05} & Section~\ref{sec:intpoints} \\
Difference of Gaussians (DoG)~\cite{L04} & Section~\ref{sec:intpoints} \\
Zernike & Section~\ref{sec:zernikefilt} \\
MSER~\cite{MCMP02}, +edge & Sections~\ref{sec:intpoints},\ref{sec:edges} \\
Selective Search~\cite{UVGS13} (SSR), +edge & Section~\ref{sec:edges} \\
Edge Detector (Fast-edge)~\cite{DZ13} & Section~\ref{sec:edges} \\
\hline
\end{tabular}
\vspace{1ex}
\caption{List of detectors used in experiments.\label{tab:detectorlist}}
\end{center}
\end{table}

\begin{figure*}
    \centering
    \begin{tabular}{c@{\hskip 0.01in}c@{\hskip 0.01in}c@{\hskip 0.01in}c}
        \subfigure{\includegraphics[height=1.25 in]{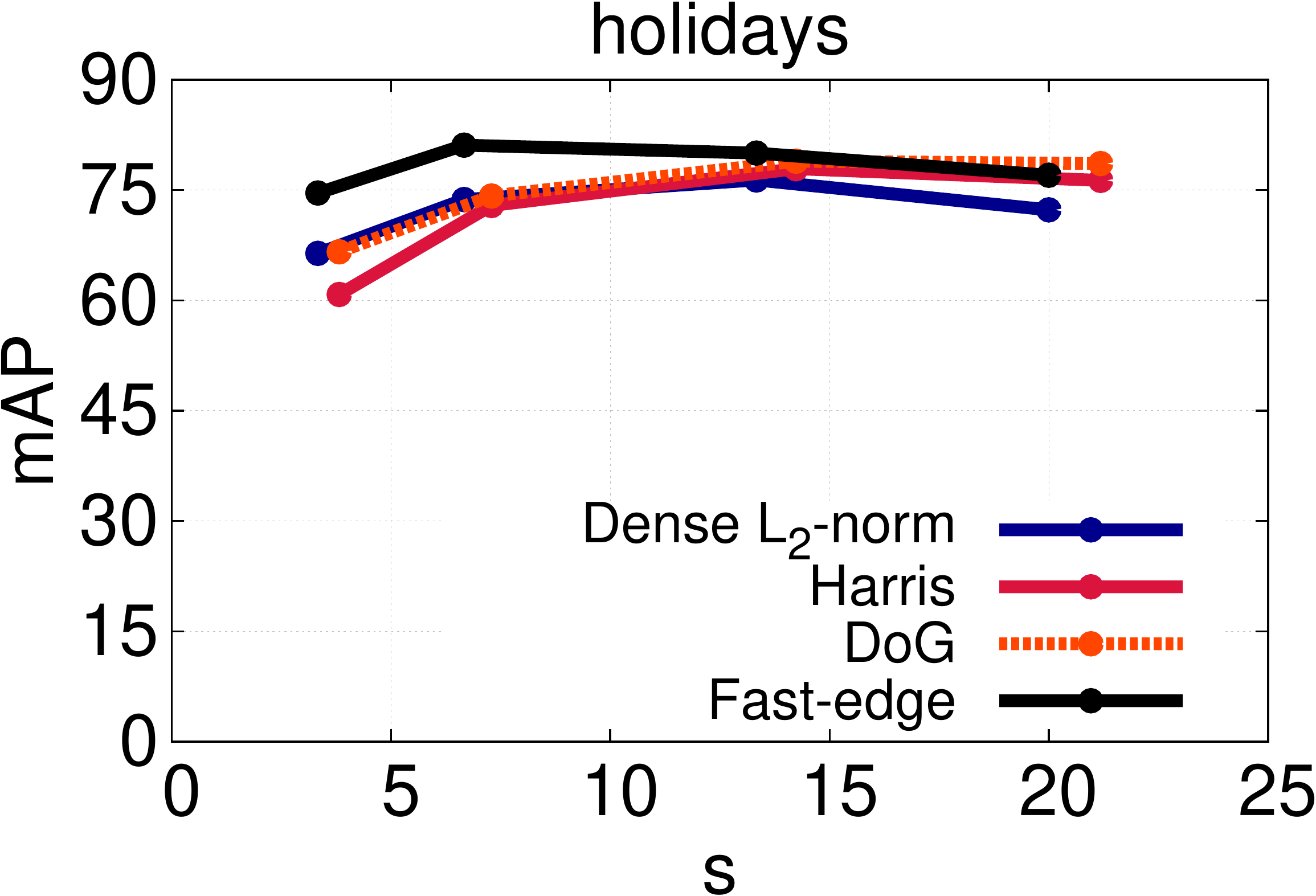}}
        &
        \subfigure{\includegraphics[height=1.25 in]{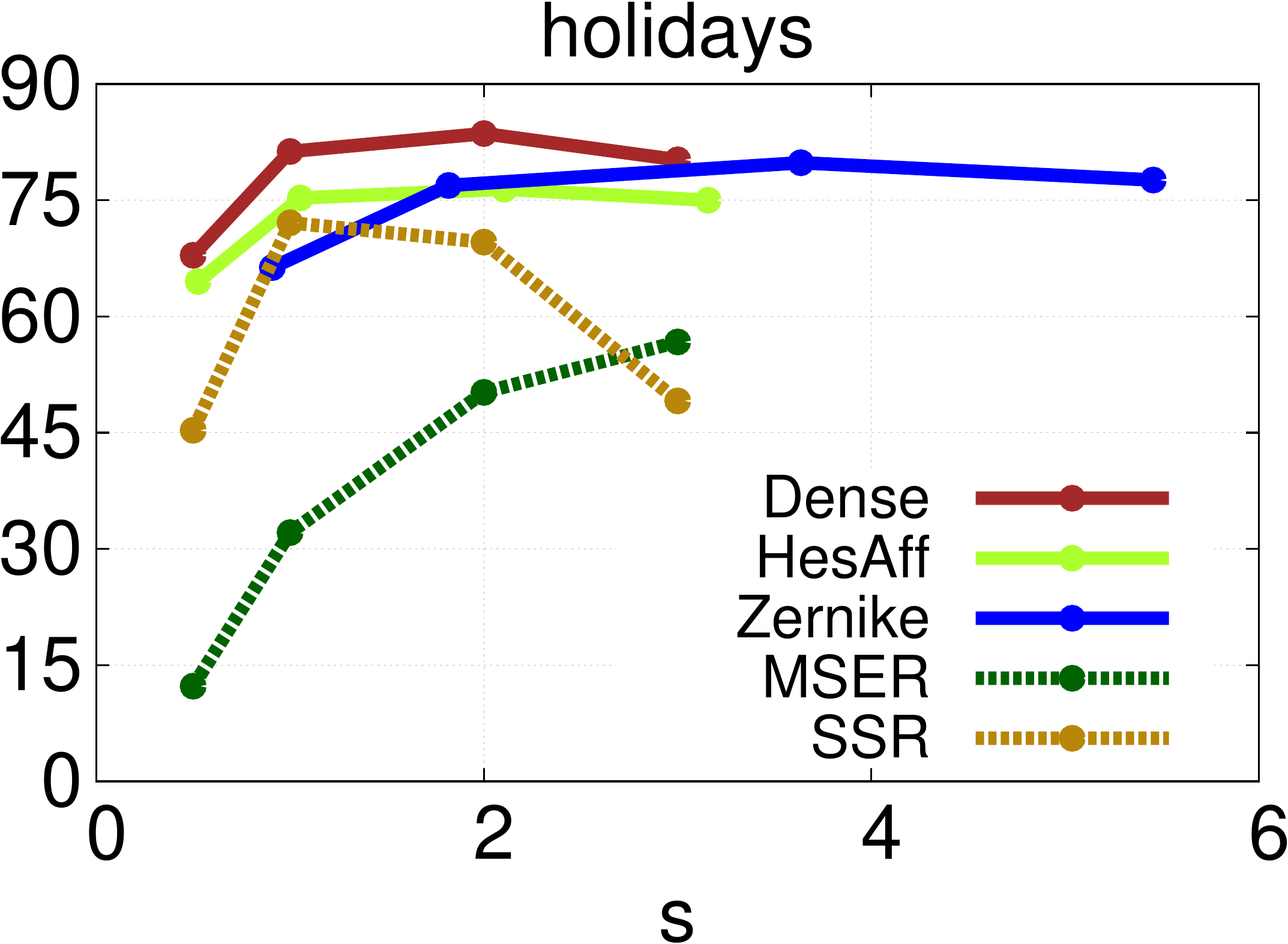}}
        &
        \subfigure{\includegraphics[height=1.25 in]{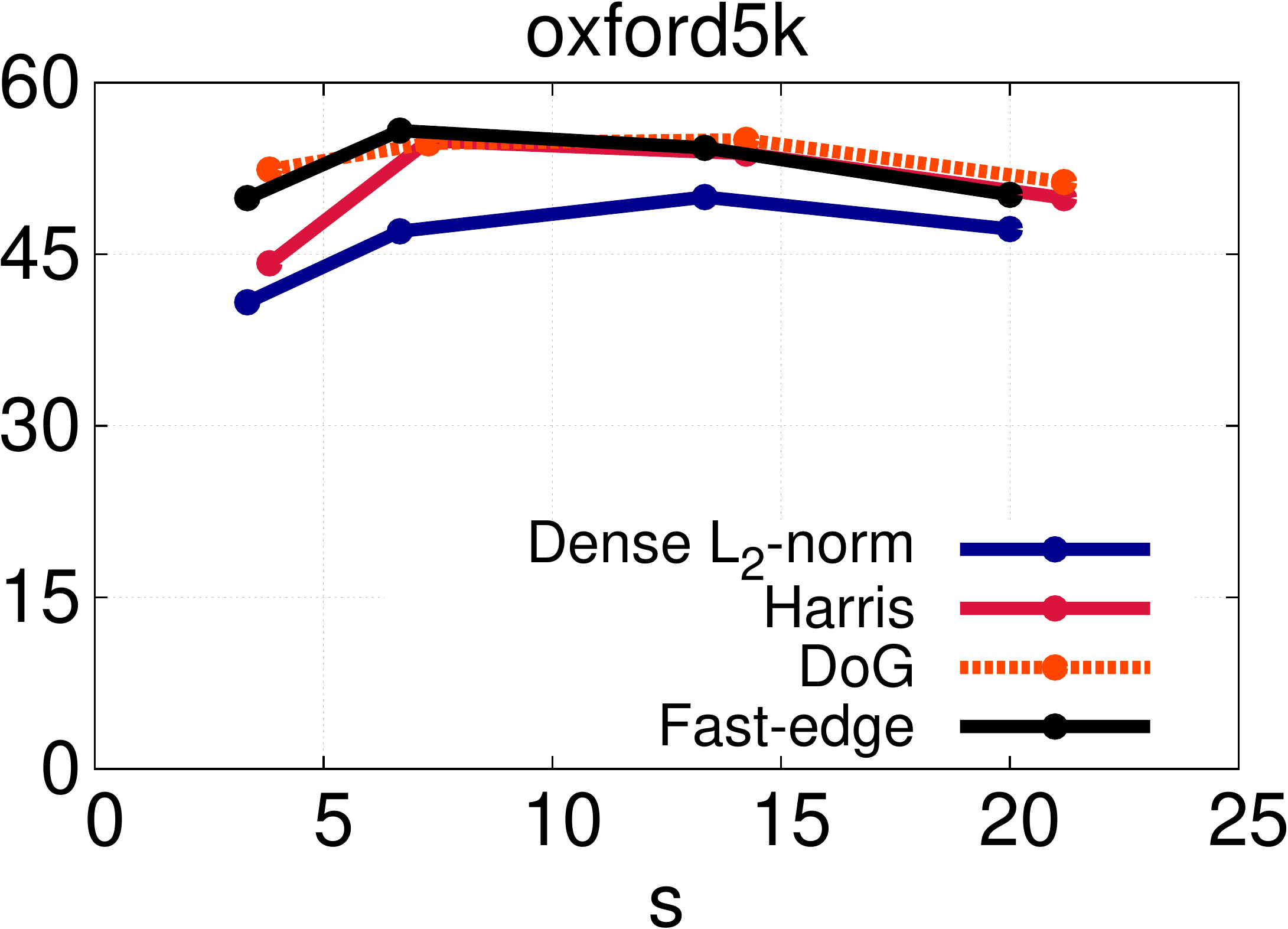}}
        &
        \subfigure{\includegraphics[height=1.25 in]{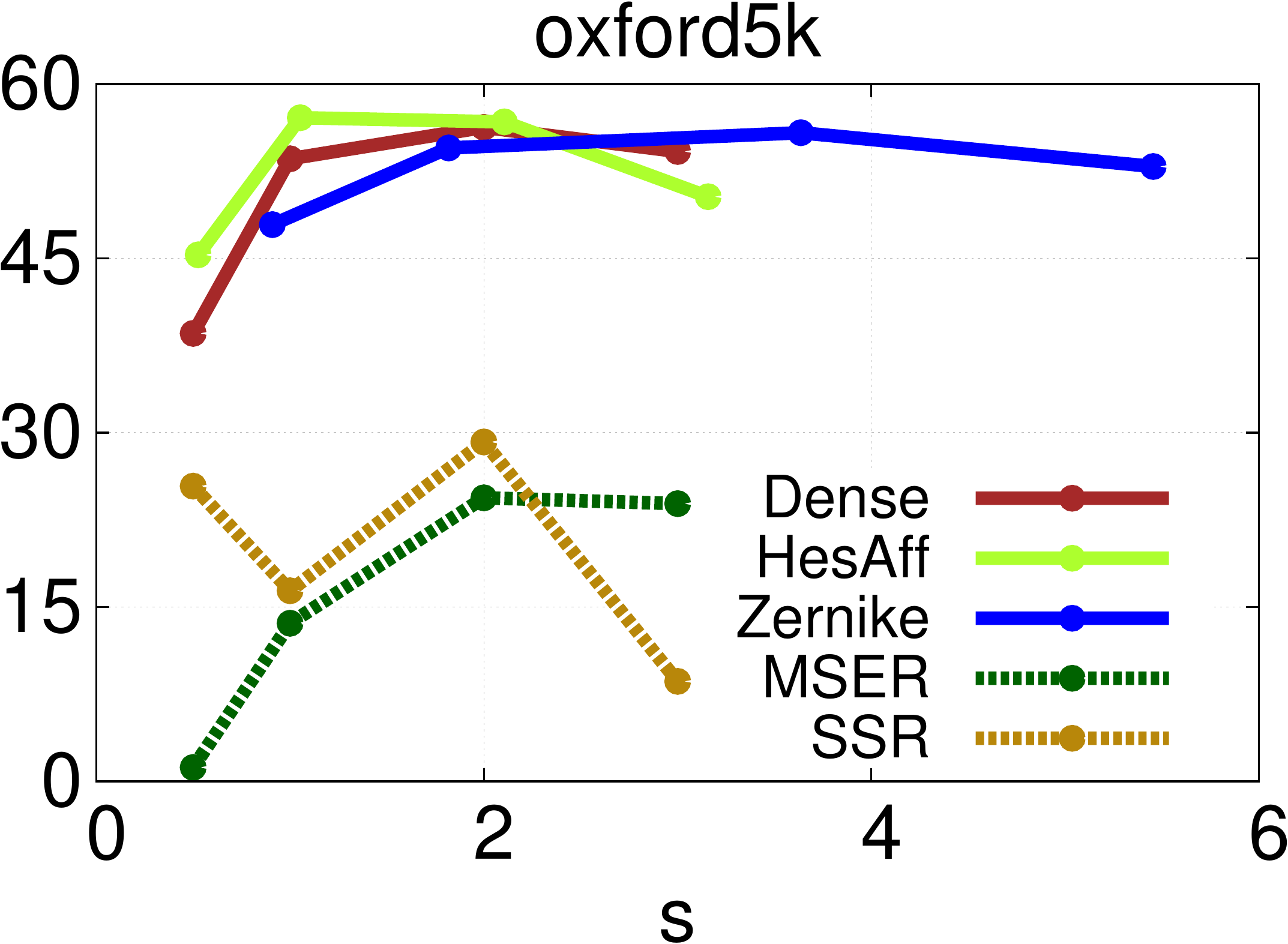}}
        
    \end{tabular}
\caption{Impact of scaling factor on retrieval performance.}
\label{fig:classScaling}
\vspace{-2ex}
\end{figure*}

\begin{figure*}
    \centering
    \begin{tabular}{c@{\hskip 0.01in}c@{\hskip 0.01in}c@{\hskip 0.01in}c}
        \subfigure{\includegraphics[height=1.25 in]{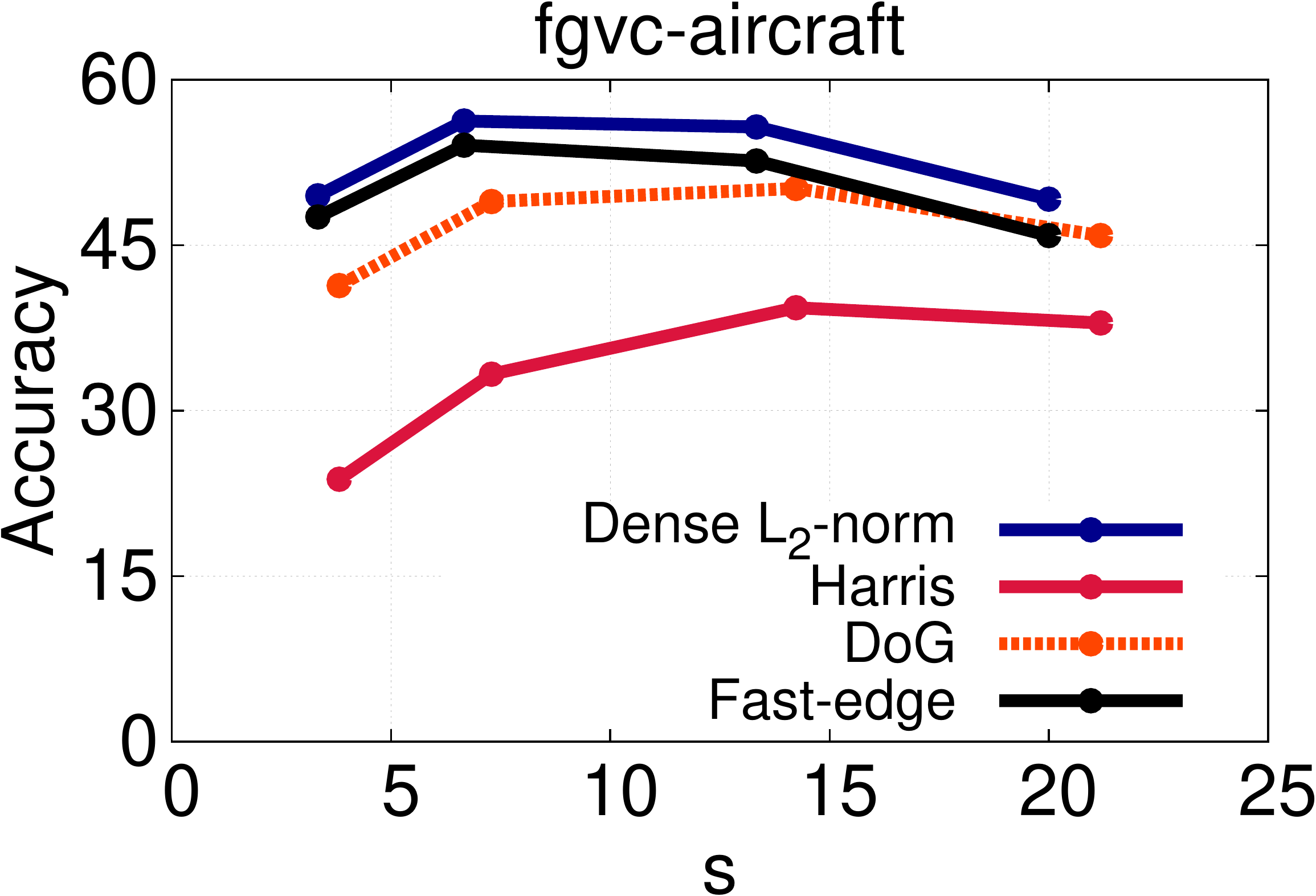}}
        &
        \subfigure{\includegraphics[height=1.25 in]{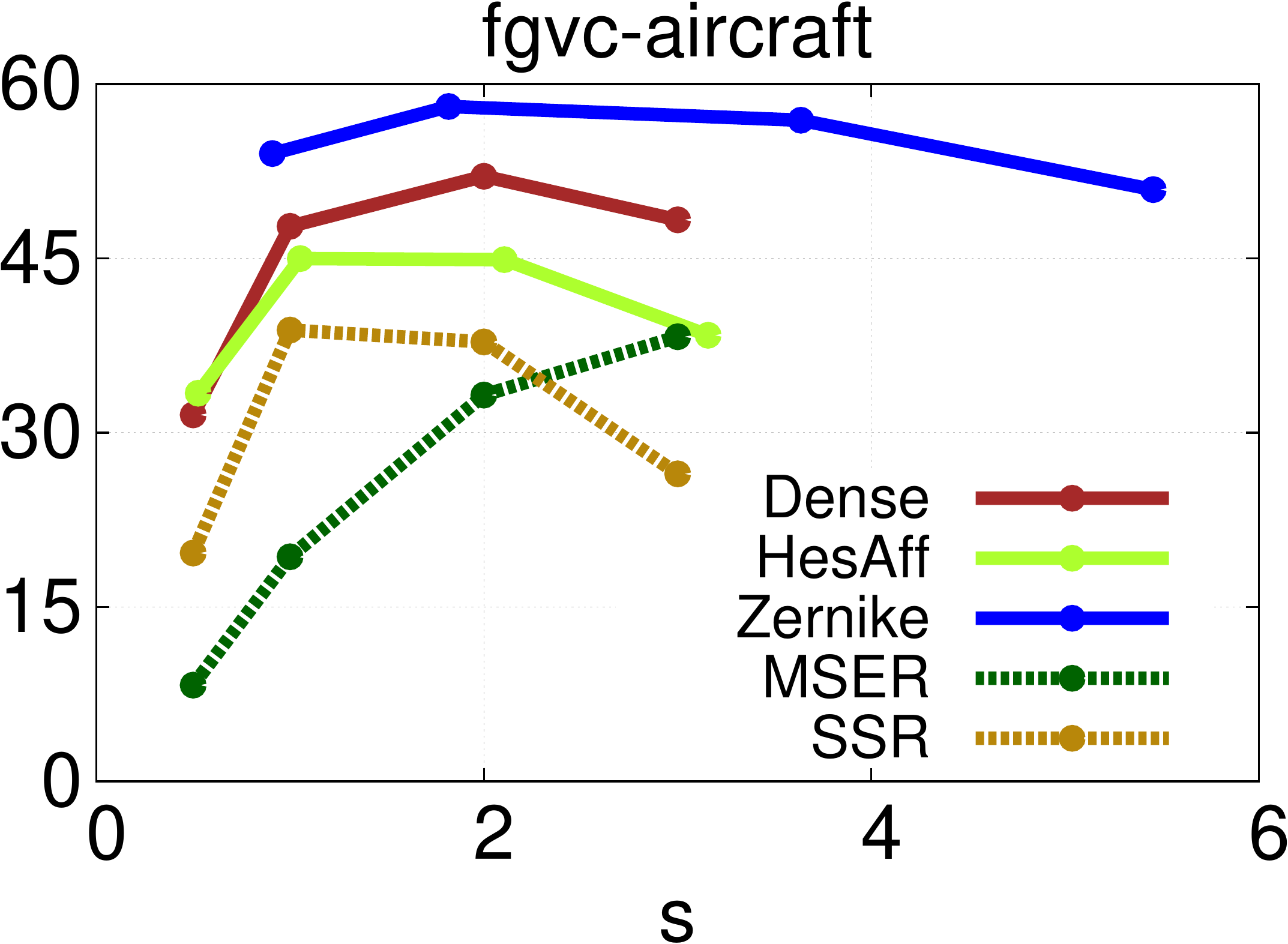}}
        &
        \subfigure{\includegraphics[height=1.25 in]{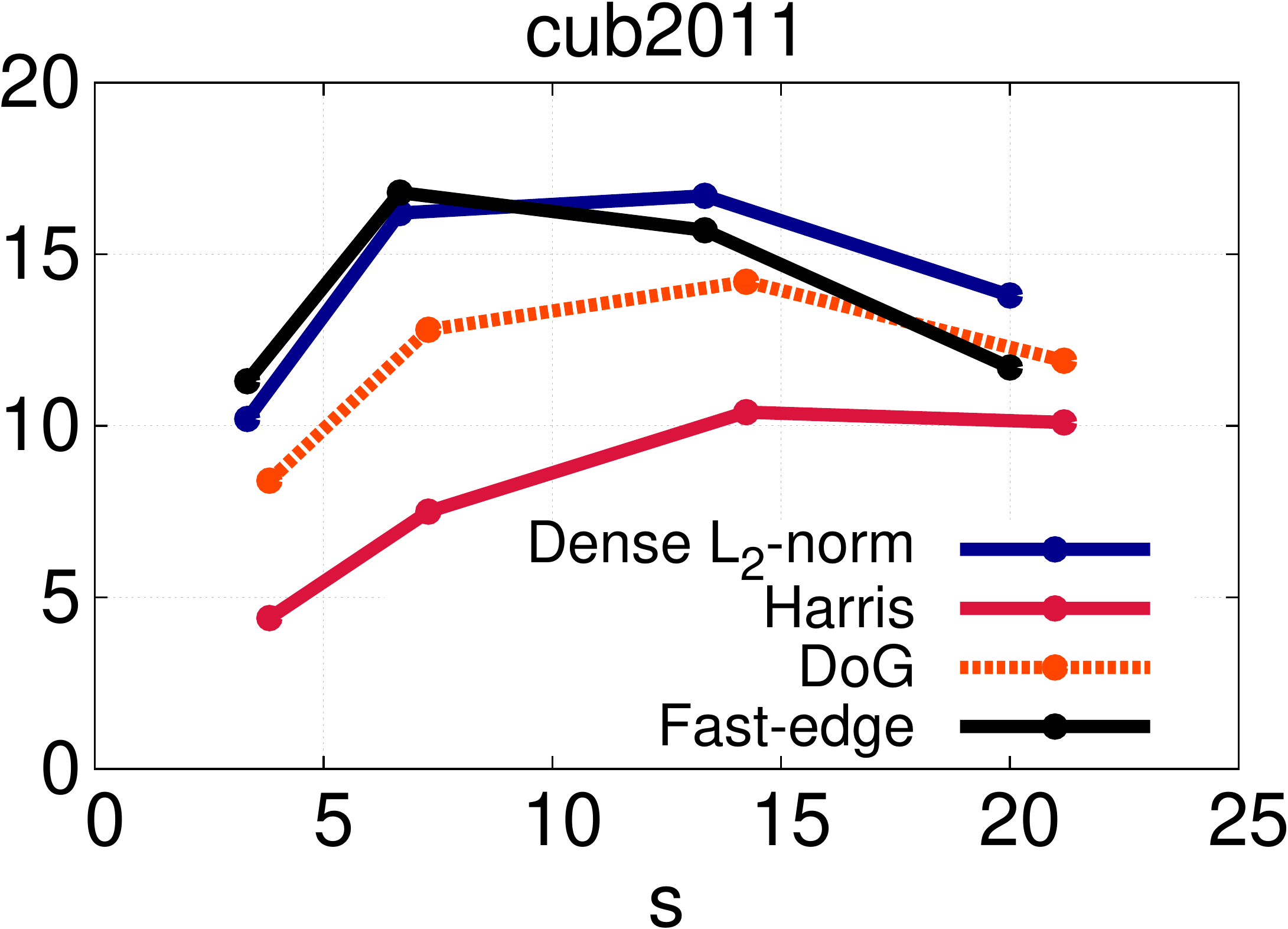}}
        &
        \subfigure{\includegraphics[height=1.25 in]{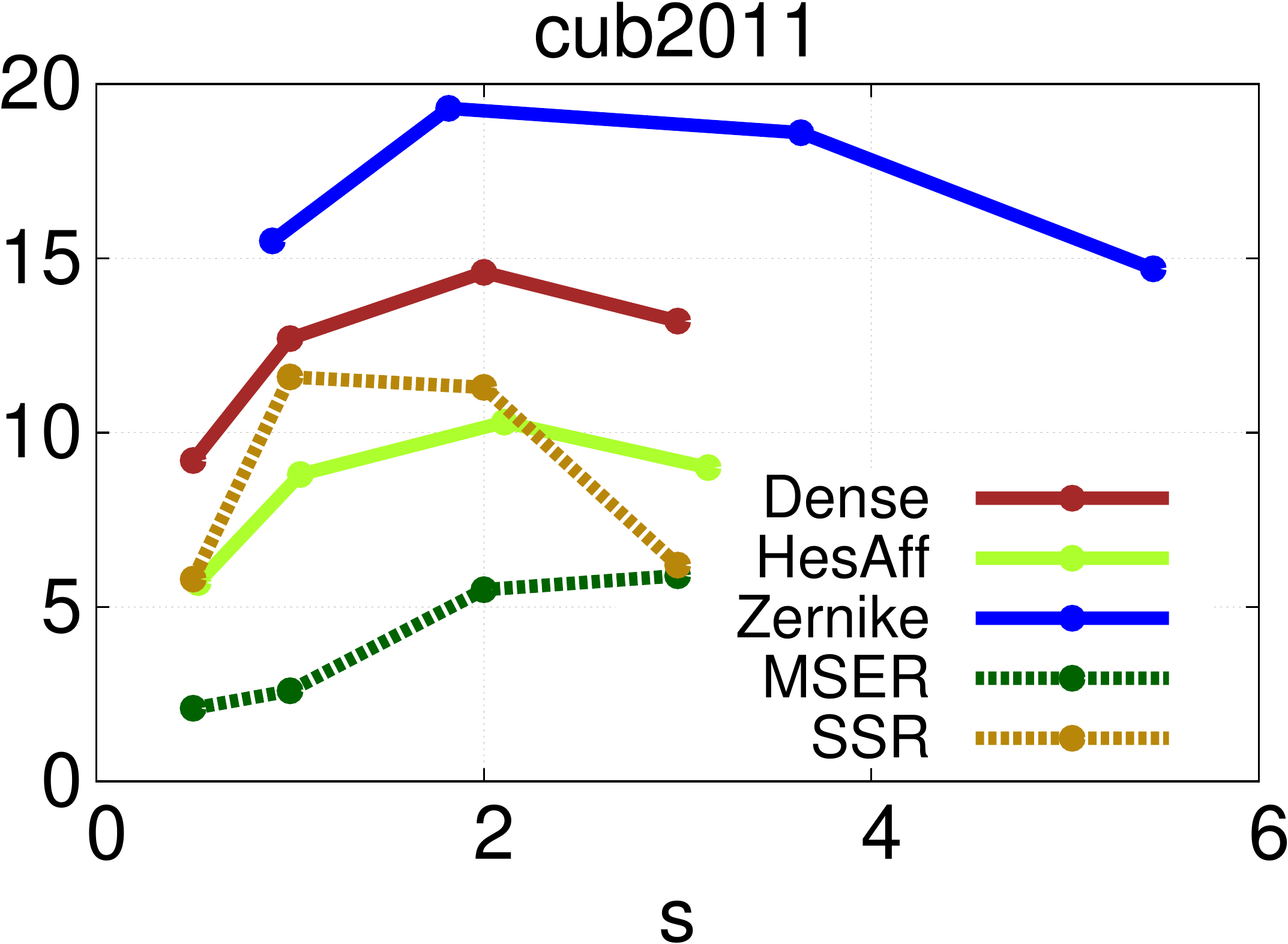}}
        
    \end{tabular}
\caption{Impact of scaling factor on fine-grained classification performance.}
\label{fig:retrScaling}
\vspace{-2ex}
\end{figure*}

We compare different detection strategies in two different applications, image retrieval and fine-grained image classification. In our experiments, we analyze aspects of feature detection and extraction separately and report our findings. List of detectors used in our experiments are shown in Table~\ref{tab:detectorlist}. Note that all the detectors we use are rotation variant~(up-right).

\subsection{Experimental Setup}
\label{sec:expsetup}
\paragraph{Image Retrieval datasets.} INRIA Holidays dataset~\cite{JDS08} consists of 1,491 personal holiday photographs. There are 500 image groups, with each group having a single query image. We use the up-right version of this dataset, where each image is rotated according to the natural orientation of objects depicted. Flickr60k dataset~\cite{JDS08} is used as a training set for this dataset.

Oxford5K Buildings dataset~\cite{Oxford5kurl} has 5,063 images of Oxford landmarks collected from Flickr. There are 55 query images. Training is performed using Paris 6k dataset~\cite{PCISZ08}.
\medskip

\paragraph{Fine-grained classification datasets.} 
We use the following evaluation benchmarks:
\begin{itemize}
\item FGVC-Aircraft dataset~\cite{MKR13} has 10,200 aircraft images. There are three different classification tasks, classification by 102 different variants, 70 different families, and 41 different manufacturers. We perform experiments on classification of variants, and use the provided train, validation and test sets.

\item Caltech-UCSD Birds 2011 dataset~\cite{WBMWSBP11} consists of images of bird species. The dataset has 200 categories, and 11,788 images. We randomly select $30\%$ of the training images as validation set~\cite{WBMWSBP11}.

\item Oxford-IIIT Pet dataset~\cite{PVZJ12} has 37 categories and 7,349 images of cats and dogs. We randomly select $50\%$ of the training images as validation set~\cite{PVZJ12}.

\item The 102 Category Flower dataset~\cite{NiZi08} consists of 102 flower categories, with each category having between 40 and 258 images. We use the provided train, validation and test sets.
\end{itemize}

We did not make any use of the provided objects regions or segments in any of the aforementioned datasets. Although using segmentation information is shown to improve results~\cite{WJY14}, we do not consider such an approach. Evaluating the performance on full images offers a direct and fair way to compare different detectors.
\medskip

\paragraph{Image retrieval pipeline.} All images are initially down-sampled to 150k pixels. We extract 128-dimensional SIFT descriptors from the detected set of keypoints. We apply RootSIFT~\cite{AZ12,JBGJ12}, center, rotate along the PCA-axis and \l2 normalize the local descriptors. Codebook of $256$ visual words is trained with k-means on an independent dataset. We encode the set of local descriptors with the VLAD~\cite{JDSP10} representation. Finally, power-law~\cite{PSM10} and \l2 normalization is applied to the VLAD vectors. During evaluation, nearest neighbors are found for each query vector, and the performance is measured with mean Average Precision (mAP).
\medskip

\paragraph{Fine-grained classification pipeline.} The down-sampling of images and post-processing of local descriptors is the same as in the retrieval task, except to the fact that we reduce to 80 dimensions using PCA, instead of only rotating and keeping all dimensions. A GMM of $256$ components is trained, and the Fisher vector~\cite{PD07} is adopted to represent an image. Power-law~\cite{PSM10} normalization is applied to the Fisher vectors, with its optimal parameter found through cross-validation. A linear classifier is trained using stochastic gradient descent~\cite{APHS13}, and the accuracy is reported.

\begin{table}
\begin{center}
\begin{tabular}{|@{\ssp}l@{\ssp}|@{\ssp}c@{\ssp}|@{\ssp}c@{\ssp}|@{\ssp}c@{\ssp}|@{\ssp}c@{\ssp}|@{\ssp}c@{\ssp}|}
\hline
Detector & conversion & $p$=11 & $p$=21 & $p$=41 & $p$=61 \\
\hline \hline
Dense & \multirow{2}{*}{$s = (p-1) / 20$} & \multirow{2}{*}{0.5} & \multirow{2}{*}{1} & \multirow{2}{*}{2} & \multirow{2}{*}{3}\\
Dense-IP~\cite{Tu10}& & & & &\\
\hline
HesAff~\cite{MTSZMSKG05} & \multirow{3}{*}{$s = (p-1) / 20$} & \multirow{3}{*}{0.5} & \multirow{3}{*}{1} &\multirow{3}{*}{2}&\multirow{3}{*}{3}\\
MSER~\cite{MCMP02} & &  &  &  &  \\
SSR~\cite{UVGS13} & &  &  &  &  \\
\hline
Harris-Laplace~\cite{MiS04} & \multirow{2}{*}{$s = p / 2.88$}& \multirow{2}{*}{3.82} &  \multirow{2}{*}{7.29} & \multirow{2}{*}{14.24} & \multirow{2}{*}{21.18} \\
DoG~\cite{L04} && & &  & \\ \hline
Dense \l2-norm & \multirow{4}{*}{$s = (p-1) / 3$} & \multirow{4}{*}{3.33} &  \multirow{4}{*}{6.67} & \multirow{4}{*}{13.33} & \multirow{4}{*}{20} \\
MSER-edge && & & & \\
SSR-edge && & & & \\
Fast-edge~\cite{DZ13} && & & & \\
\hline
Zernike & $s = (p-1) / 11$ & 1 & 1.91 & 3.73 & 5.55 \\
\hline
\end{tabular}
\vspace{1ex}
\caption{Relation of scaling factor to patch size. We linearly relate scaling factor to patchsize.\label{tab:scalepatch}}
\end{center}
\end{table}

\subsection{Impact of the scaling factor}

Traditionally, a detected region of interest is mapped to a rectangular patch of width equal to $p$ pixels and a local descriptor is extracted from it. It is beneficial to enlarge this measurement region by a scaling factor~\cite{SVZ14}, denoted by $s$. The ratio between the descriptor measurement region (used to extract local descriptors) and the detection measurement region (used to compute the interestingness measure to detect keypoints) is exactly the scaling factor $s$. When $s=1$, these two are identical.

We investigate the impact of the scaling factor for a variety of detection processes. We linearly relate the scaling factor to the patch size, such that larger regions will be normalized to patches of larger resolution. This  exact relation is presented in Table~\ref{tab:scalepatch} for all the examined detectors.

Affine regions are isotropically enlarged, and when no enlargement is involved, the patch size is equal to 21. For Zernike, no scaling factor corresponds to descriptor measurement region equal to 11, since this is the filter size we use (see Section~\ref{sec:methods}). For regular dense sampling, where there is no detector measurement region, we have arbitrarily defined the scaling factor to be equal to 1 when the descriptor measurement region size is 21 pixels.

We perform experiments to investigate the impact of the scaling factor and present results in Figures~\ref{fig:classScaling} and \ref{fig:retrScaling}. It is observed that such kind of region enlargement is beneficial for all detectors. Performance increases by increasing the descriptor measurement regions, while for large increase it saturates or even drops. Our conclusions agree with the conclusions of Simonyan \etal~\cite{SVZ14}, but we evaluate the scaling factor for a larger number of detectors and both classification and retrieval tasks. We adopt the scaling factor that corresponds to a patch size of $41$ for all detectors, as it seems to be a good choice overall. The same stands for detectors not included in the corresponding figures. Note that we do not compare different detectors with each other in this experiment. We investigate the impact of scaling factor for each detector individually. Each detector produces different number of features per image in this experiment. We consider this a crucial parameter for performance and investigate it in Section~\ref{sec:perf}.

\begin{figure*}
    \centering
    \begin{tabular}{c c c c}
        \subfigure{\includegraphics[height=0.165\textwidth]{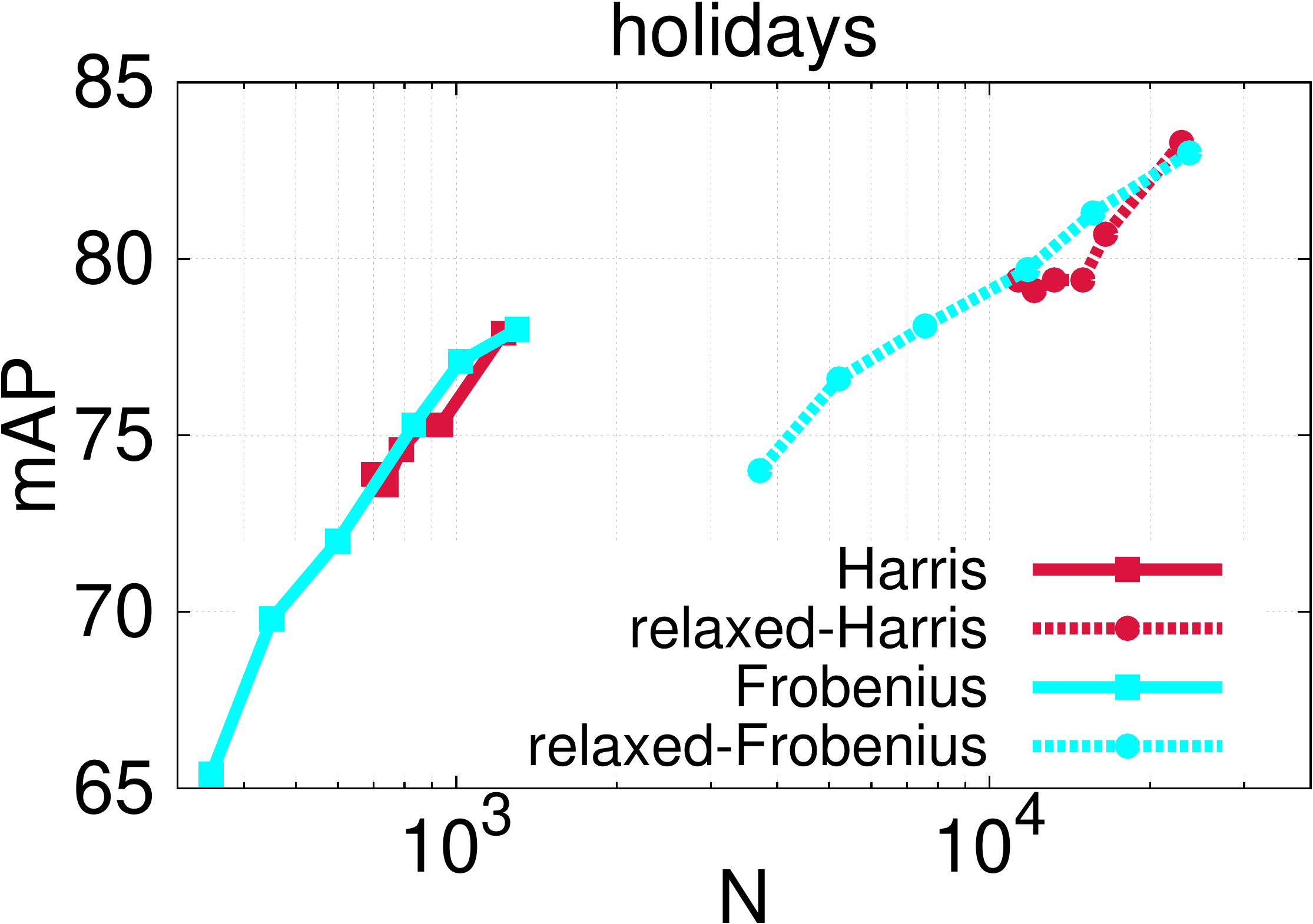}}
        &
        \subfigure{\includegraphics[height=0.165\textwidth]{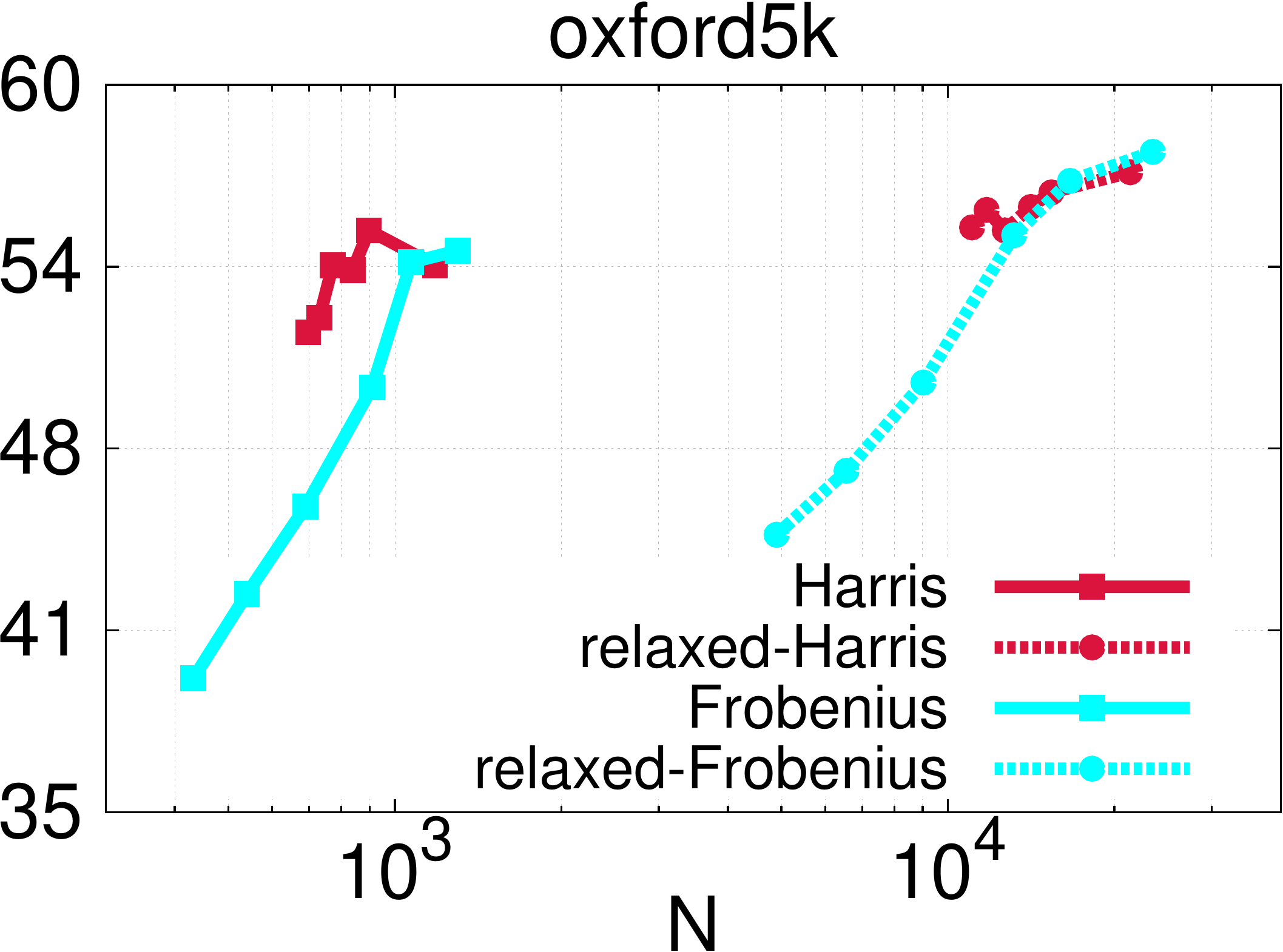}}
        &
        \subfigure{\includegraphics[height=0.165\textwidth]{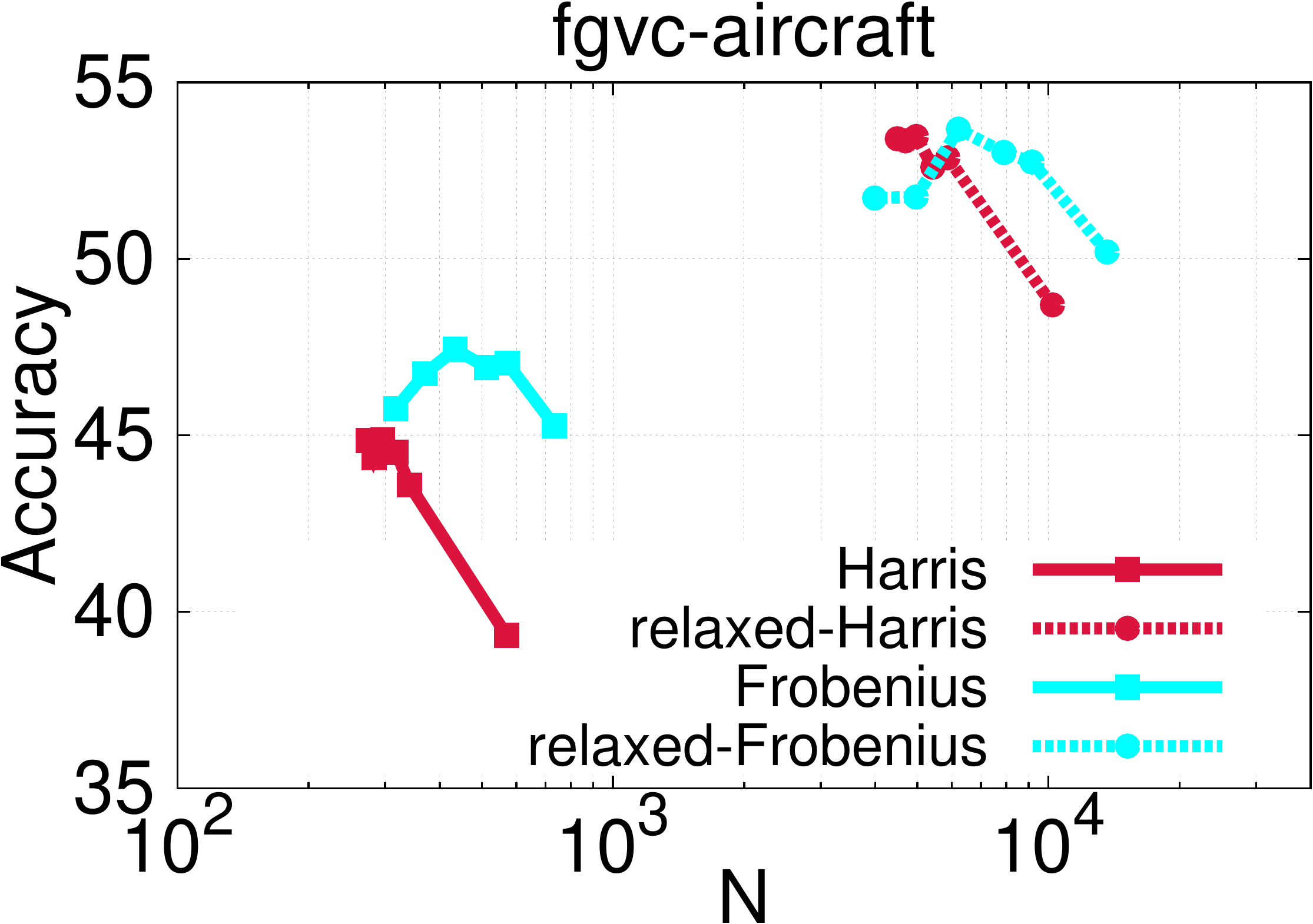}}
        &
        \subfigure{\includegraphics[height=0.165\textwidth]{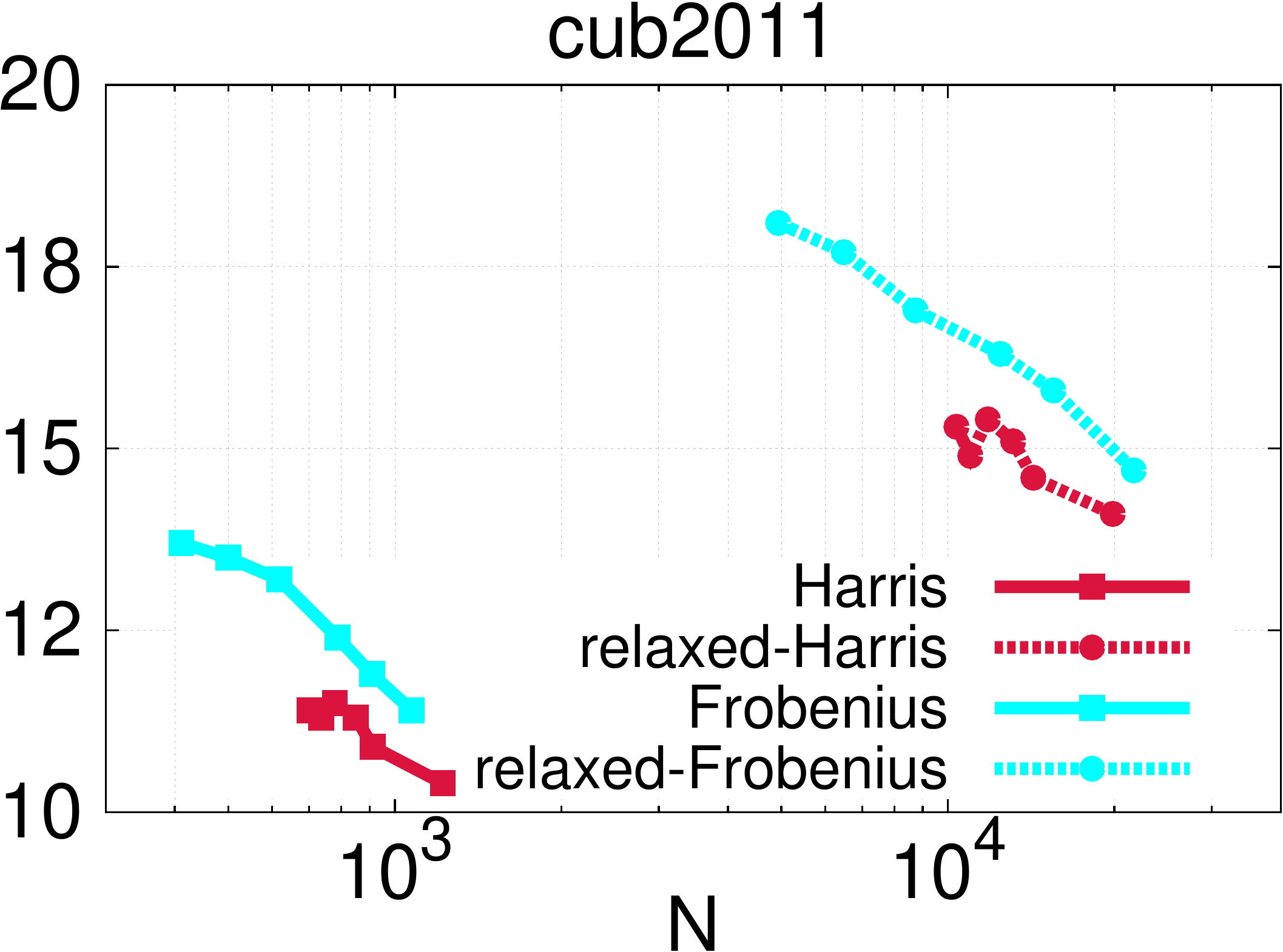}}

    \end{tabular}
\caption{Performance comparison between Harris-Laplace and its relaxed modifications that we propose. We show performance versus average number of descriptors per image $N$.}
\label{fig:retrLocalHarris}
\end{figure*}

\begin{figure*}
    \centering
    \begin{tabular}{c c c c}
        \subfigure{\includegraphics[height=0.165\textwidth]{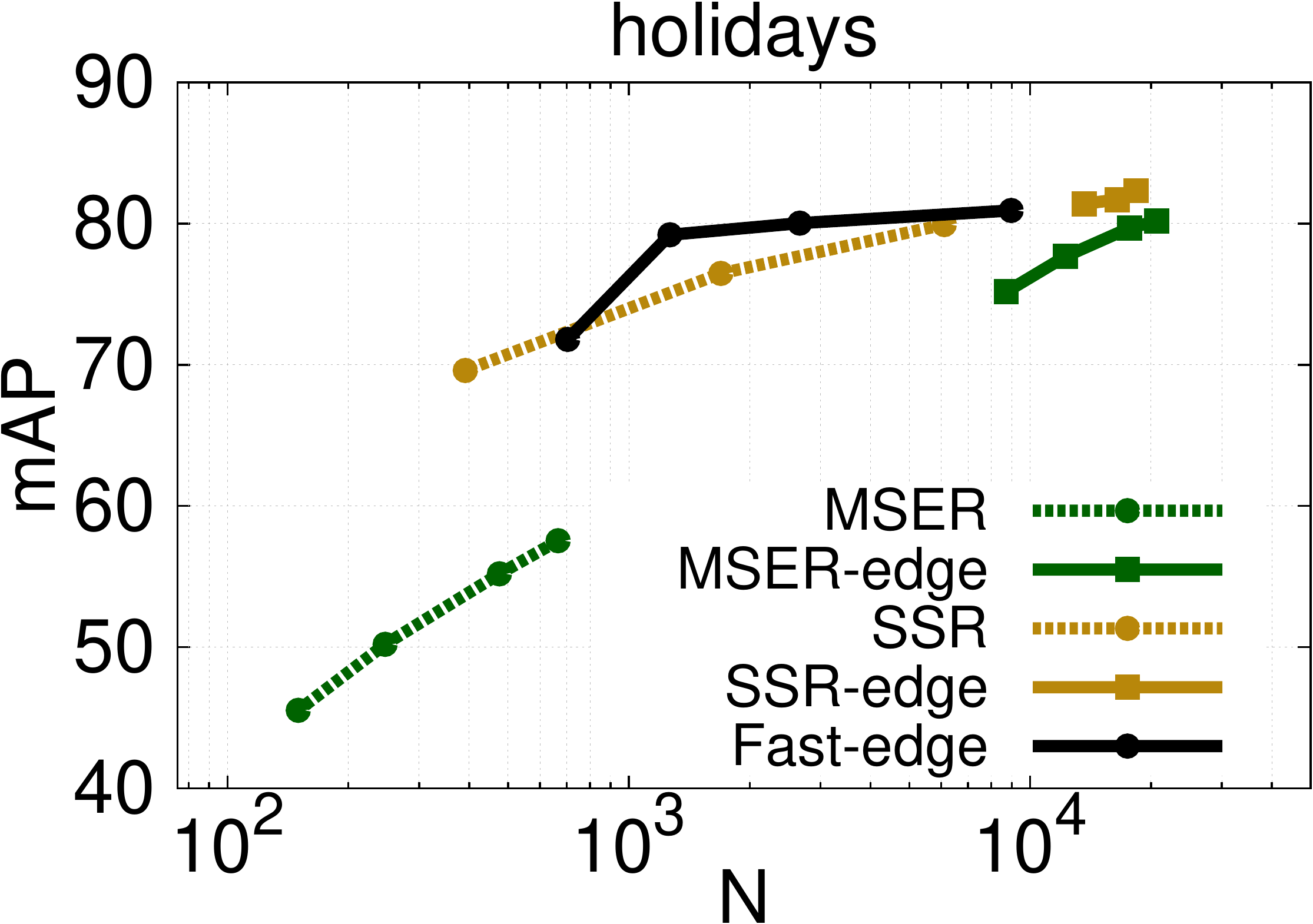}}
        &
        \subfigure{\includegraphics[height=0.165\textwidth]{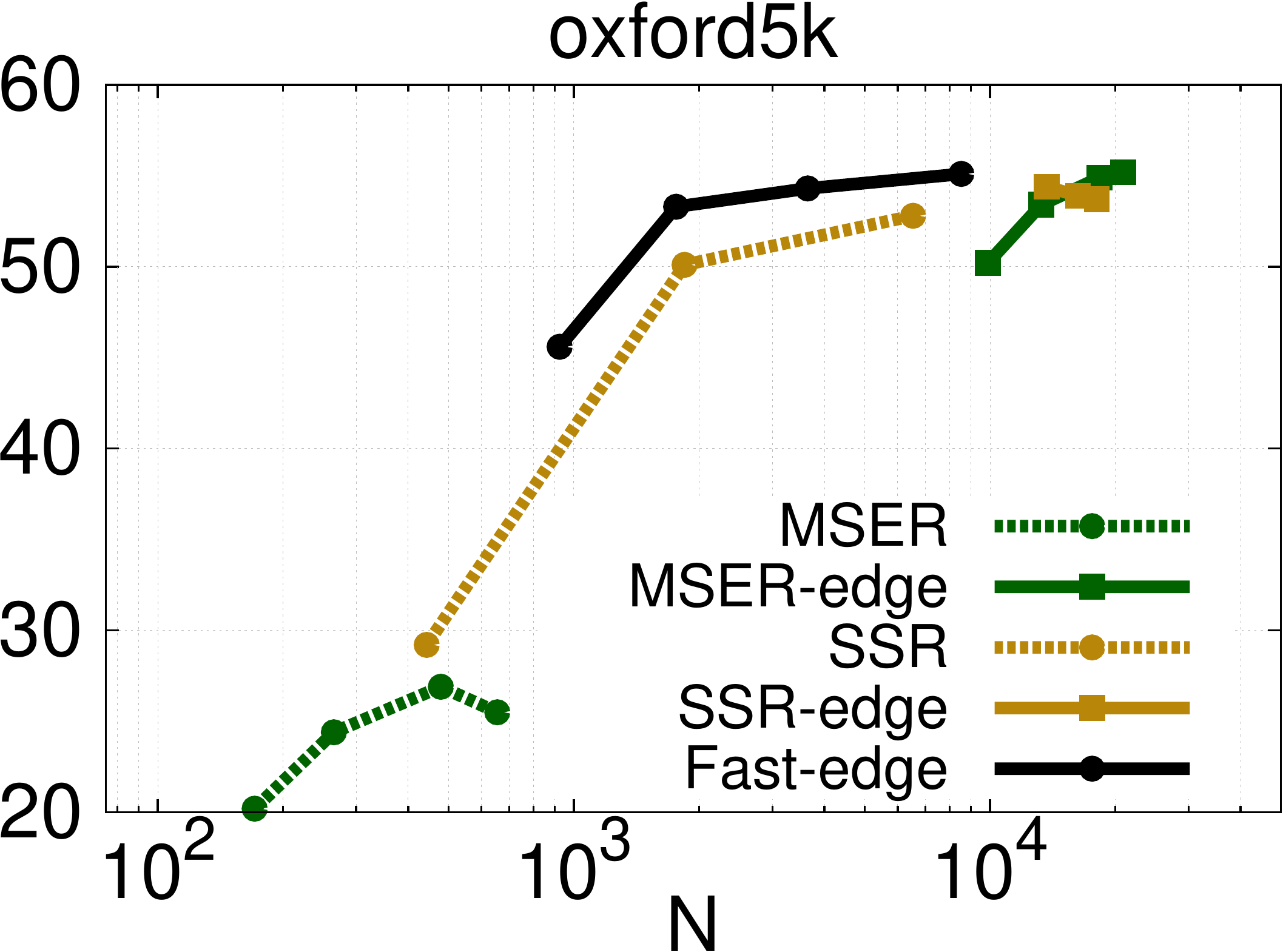}}
        &
                \subfigure{\includegraphics[height=0.165\textwidth]{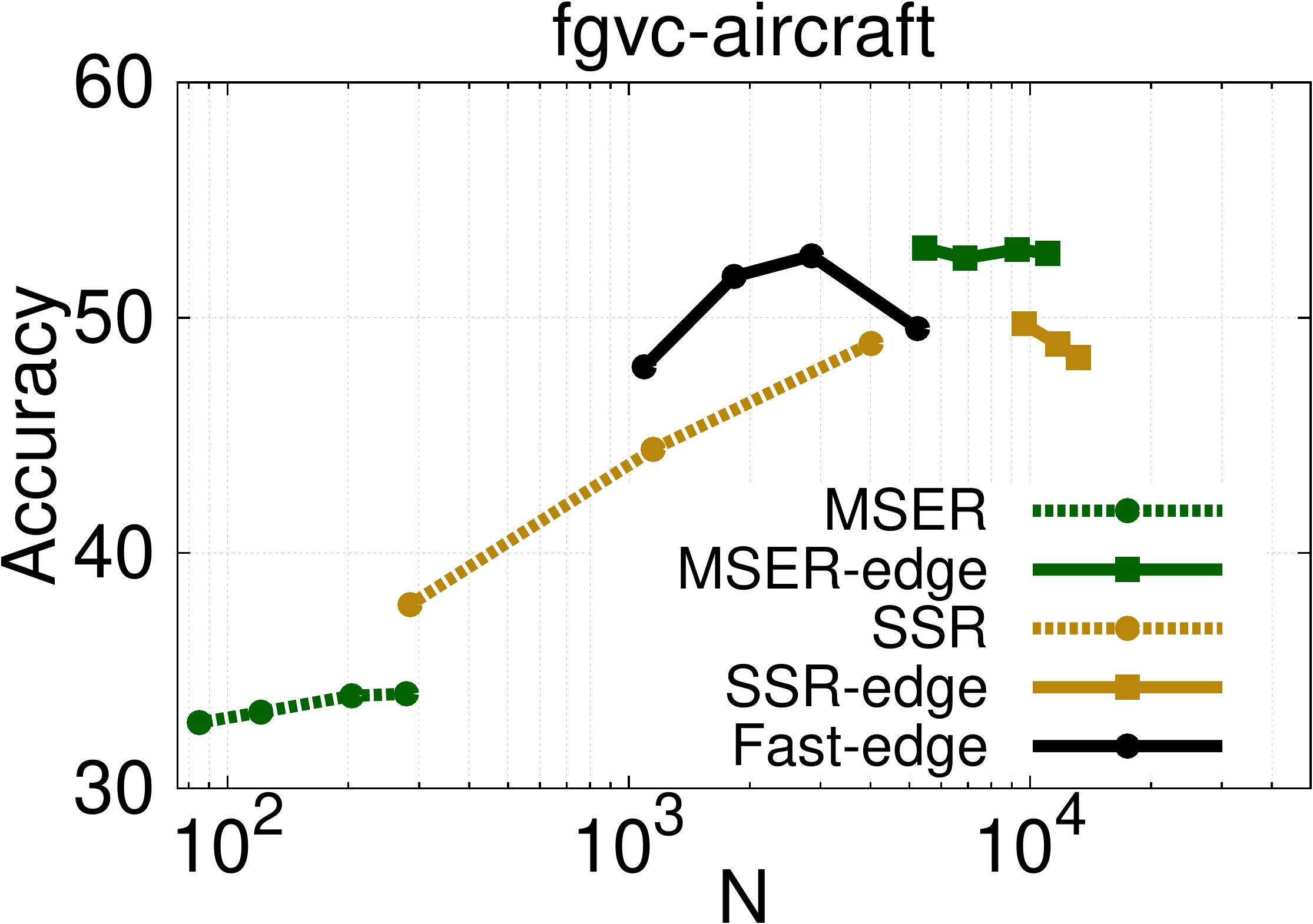}}
        &
        \subfigure{\includegraphics[height=0.165\textwidth]{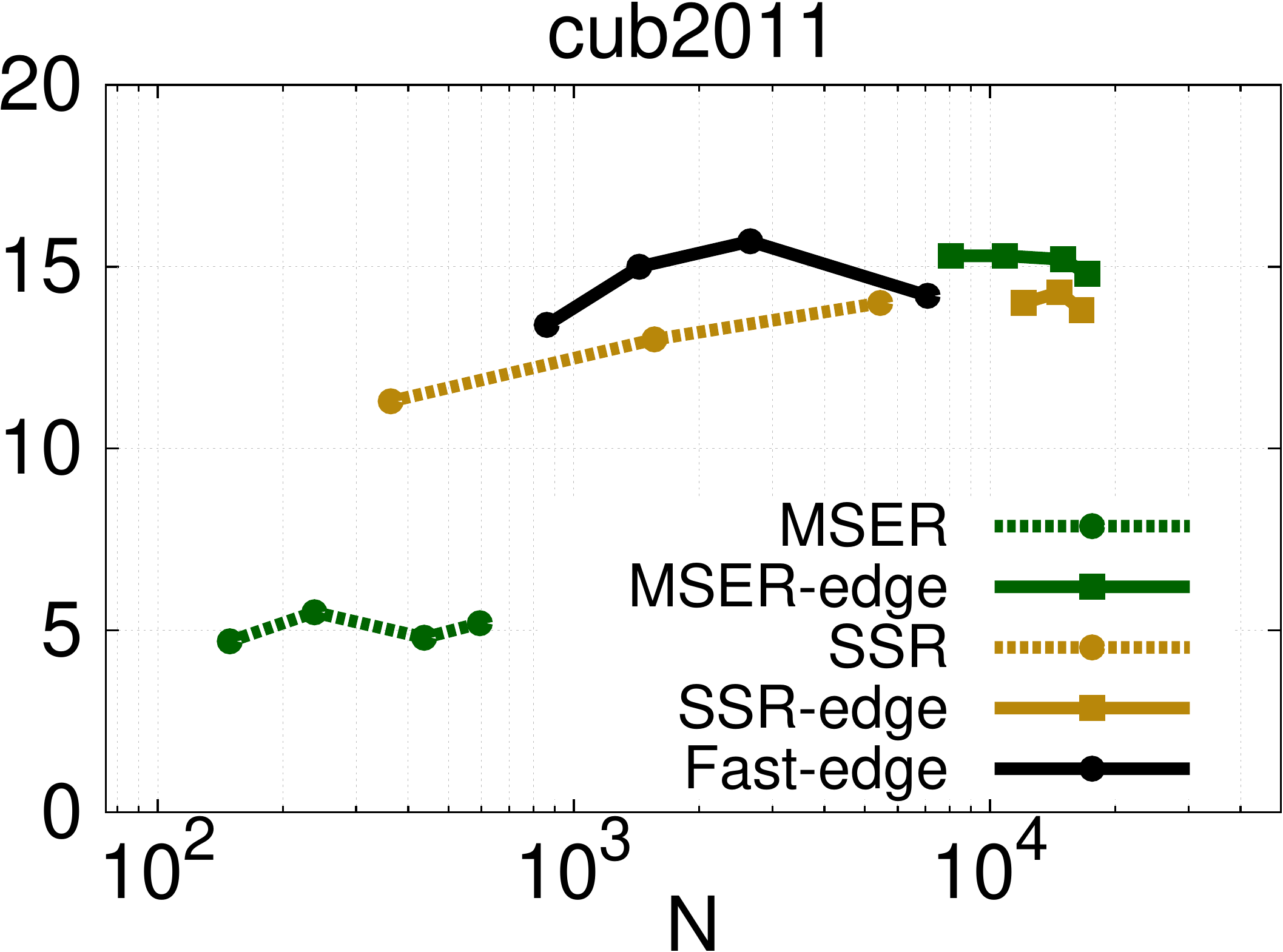}}

    \end{tabular}
\caption{Performance comparison between approaches that describe a detected region by fitting a single ellipse and approaches that sample patches along the region edges. Fast-edge is another detector that samples patches on edges. We show performance versus average number of descriptors per image $N$.}
\label{fig:expedges}
\end{figure*}

\begin{figure*}
    \centering
    \begin{tabular}{c c c c}
        \subfigure{\includegraphics[height=0.165\textwidth]{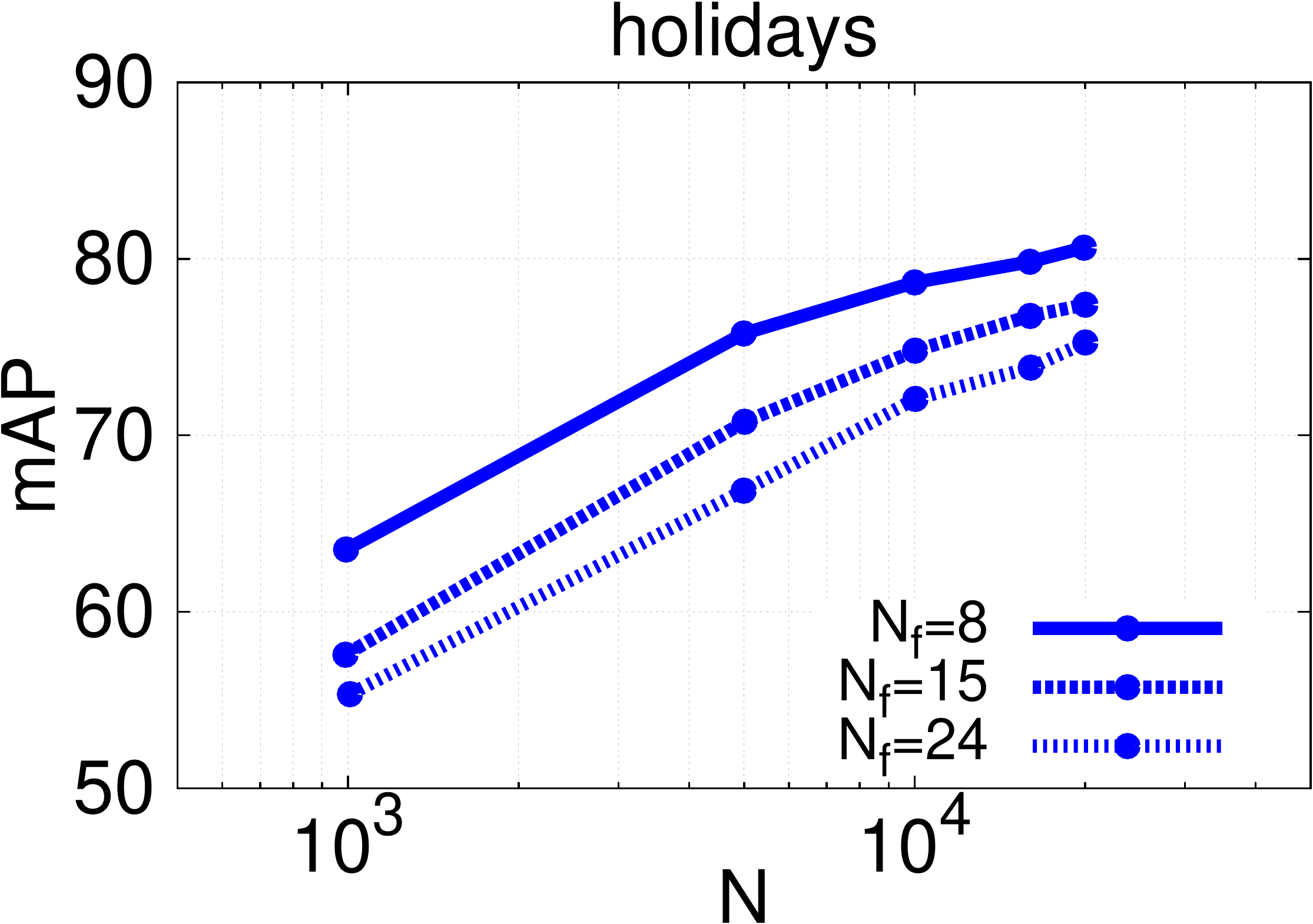}}
        &
        \subfigure{\includegraphics[height=0.165\textwidth]{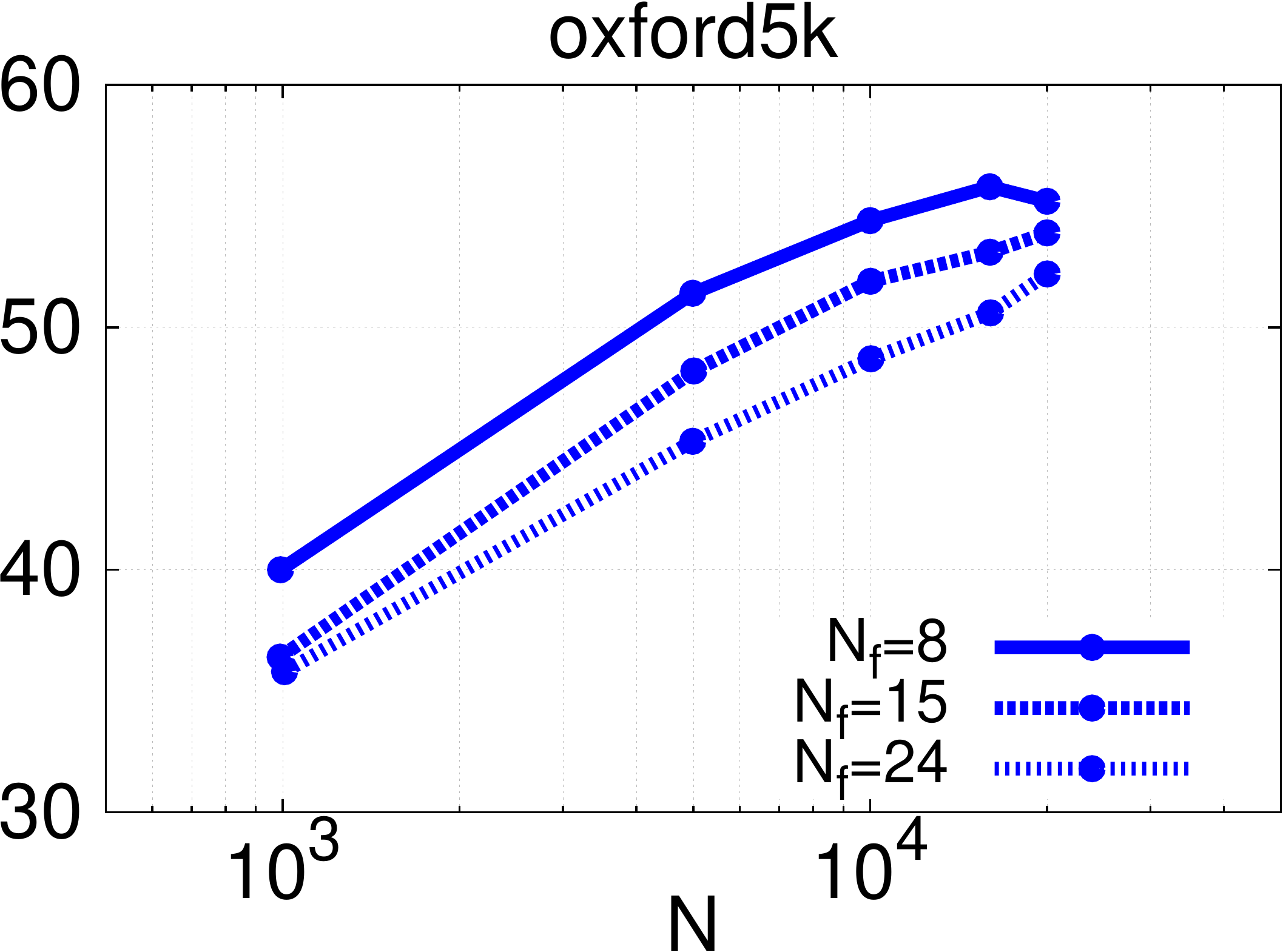}}
        &
        \subfigure{\includegraphics[height=0.165\textwidth]{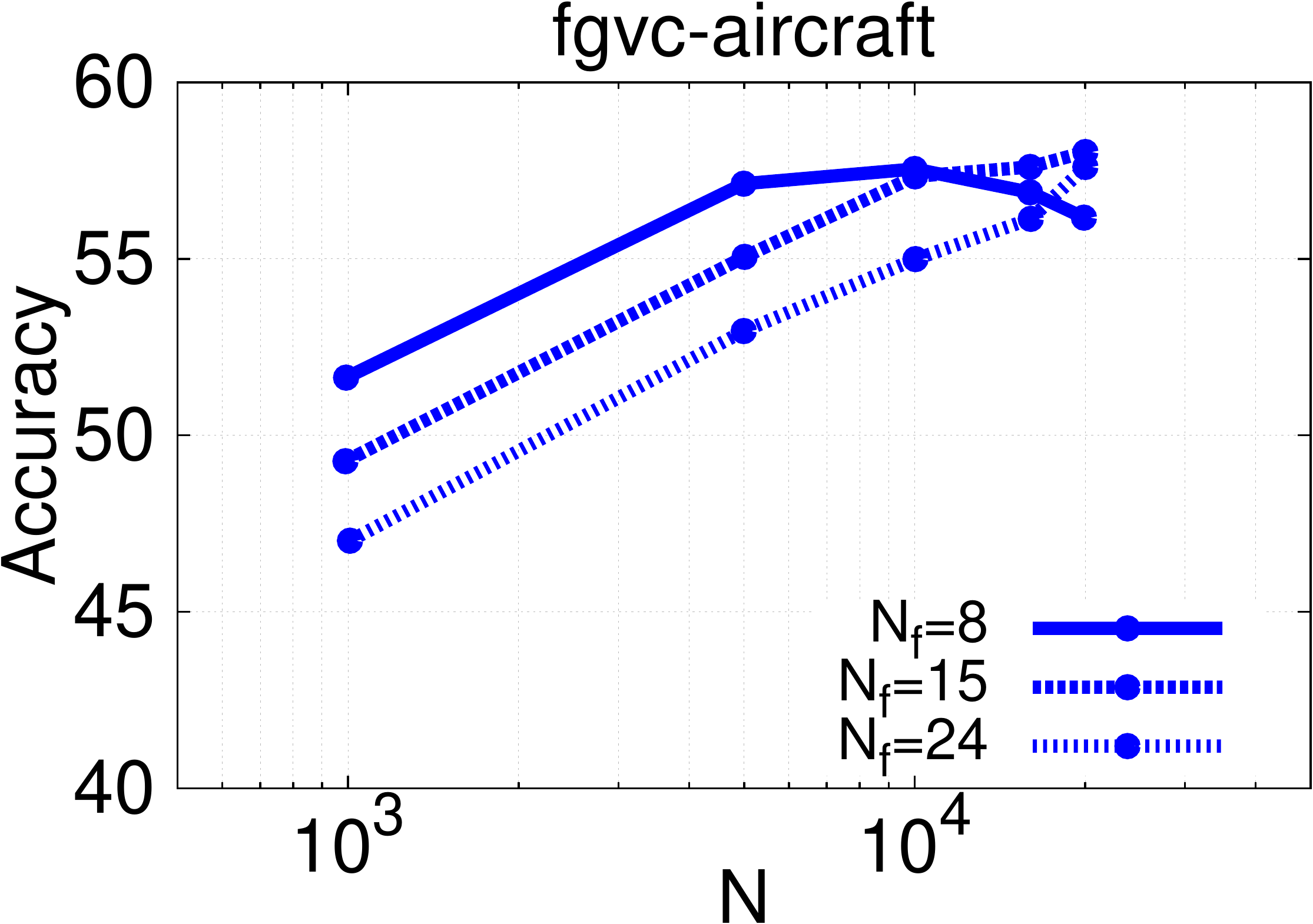}}
        &
        \subfigure{\includegraphics[height=0.165\textwidth]{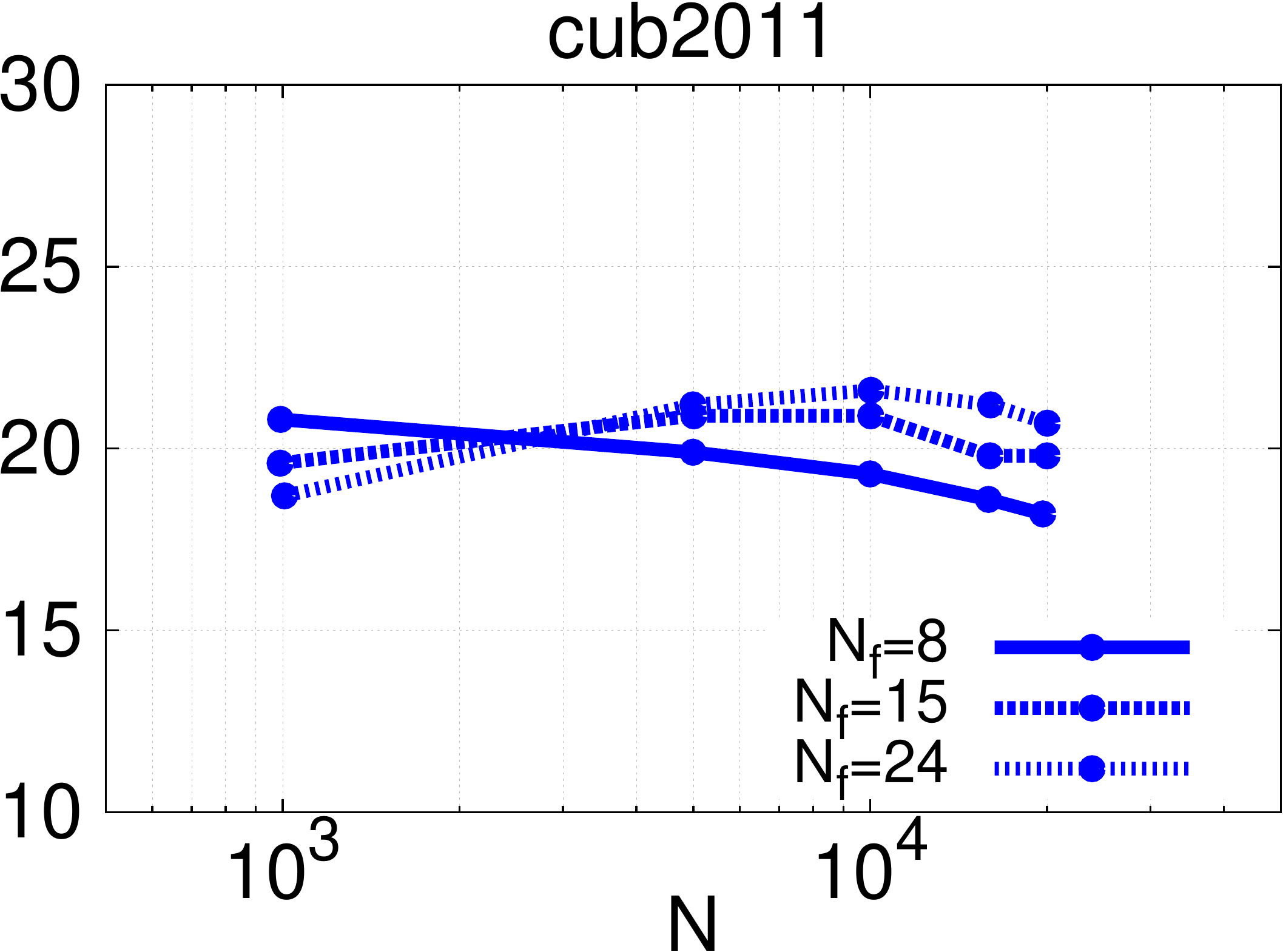}}
    \end{tabular}
\caption{Performance versus average number of descriptors for Zernike detector with different number of filters.}
\label{fig:zernikenf}
\end{figure*}

\subsection{Impact of focusing on edges}
\label{sec:localization}

In Section~\ref{sec:methods} we proposed detectors that have different localization properties than the existing ones. In particular, we try to focus the local representation on image edges.

In Figure~\ref{fig:retrLocalHarris}, we present the performance measured for Harris-Laplace detector and the relaxed modifications that we proposed. Relaxed-Harris produces more keypoints and consistently improves performance. The relaxation based on Frobenius norm also seems to improve. An exception is the Oxford5K Buildings dataset, where due to particular object matching, original Harris keypoints are localized to have higher repeatability and perform better.

We argue that the typical choice of fitting an ellipse and trying to describe a region by its interior is not necessarily the optimal option for the tasks we consider. Figure~\ref{fig:expedges} shows the results of comparing such a traditional approach to our proposal of extracting local descriptors from patches centered on the region borders. We perform this comparison for MSER regions and selective search regions (SSR). Interestingly, the standard choice does not appear to be the optimal. A larger number of features is obtained when focused on edges, which generally results in better performance. Finally, the region detector based on the fast edge detection algorithm~\cite{DZ13} (fast-edge) performs rather well, especially for intermediate number of descriptors per image.

Additionally, we conduct localization experiments for the proposed Zernike detector by directly controlling the number of keypoints via parameter $N_\mathrm{z}$. We also capture more complementary information by increasing the number of filters $N_\mathrm{f}$. We evaluate performance using all polynomials up to order 2, 3 and 4. These correspond to 8, 15 and 24 filters respectively. 
We present the results in Figure~\ref{fig:zernikenf}. Less filters seem to perform better on retrieval datasets. On the contrary, more filters are needed for fine-grained classification, possibly to capture larger intra class variations that exist. We set $N_\mathrm{f}$ equal to 8 for retrieval and 15 for fine-grained classification.

\begin{figure*}
    \centering
    \begin{tabular}{c c}
        \subfigure{\includegraphics[width=0.475\textwidth]{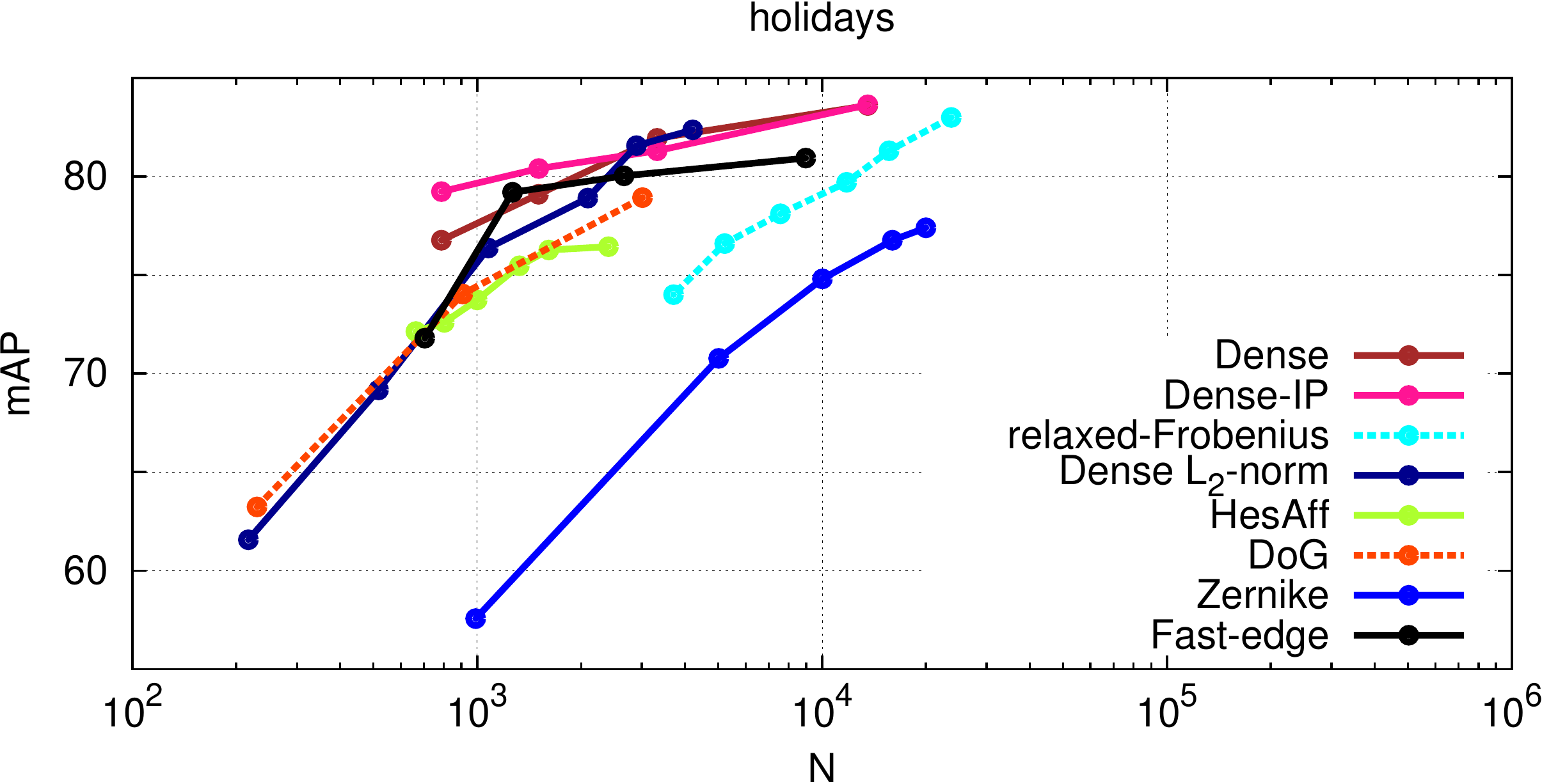}}
        &
        \subfigure{\includegraphics[width=0.475\textwidth]{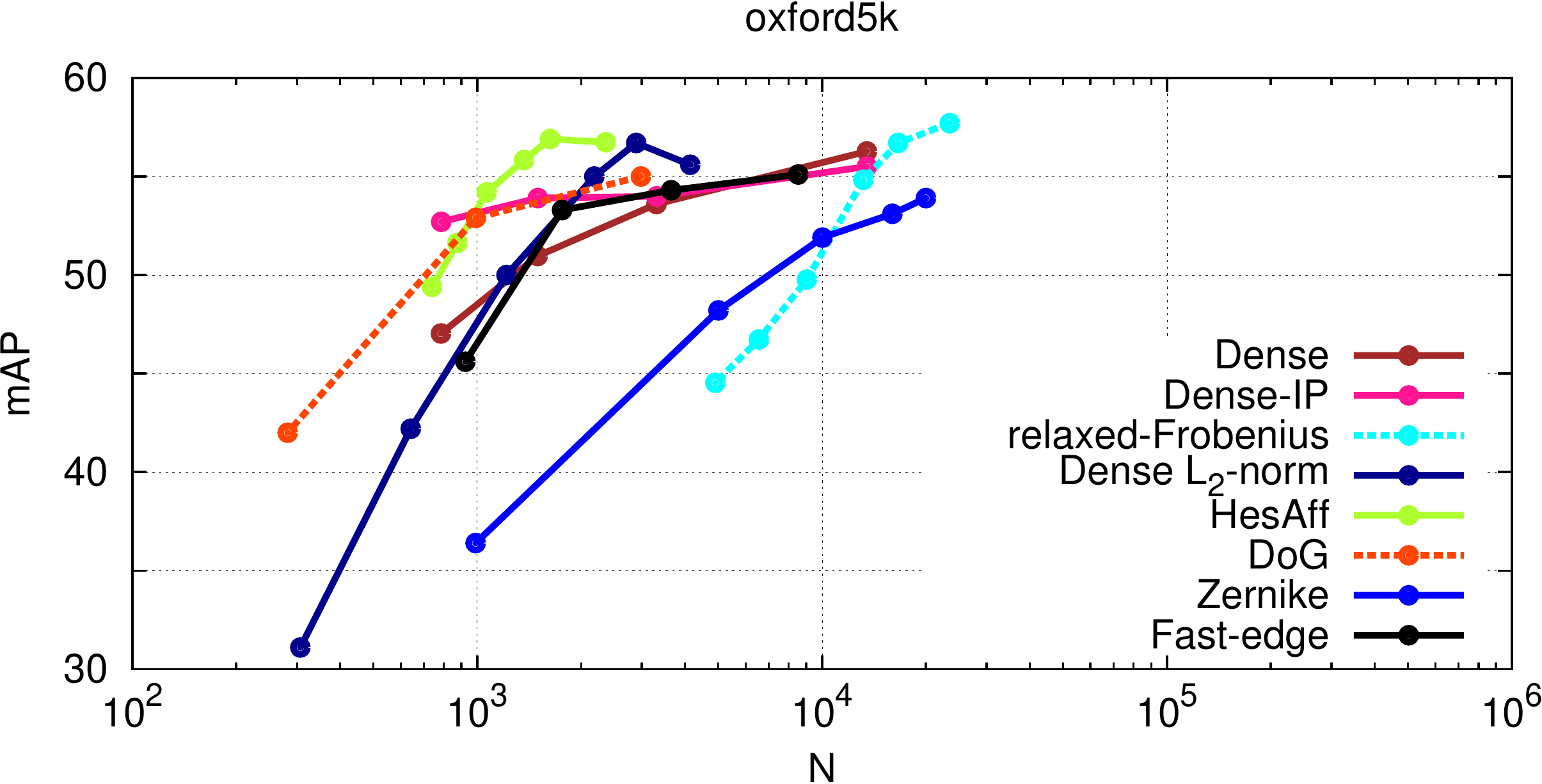}}
        
    \end{tabular}
\vspace{1ex}
\caption{Performance comparison on image retrieval versus number of descriptors per image $N$. The parameter values shown in Table~\ref{tab:numfeatparam} are used.}
\label{fig:retrcomp}
\end{figure*}

\begin{figure*}
    \centering
    \begin{tabular}{c c}
        \subfigure{\includegraphics[width=0.475\textwidth]{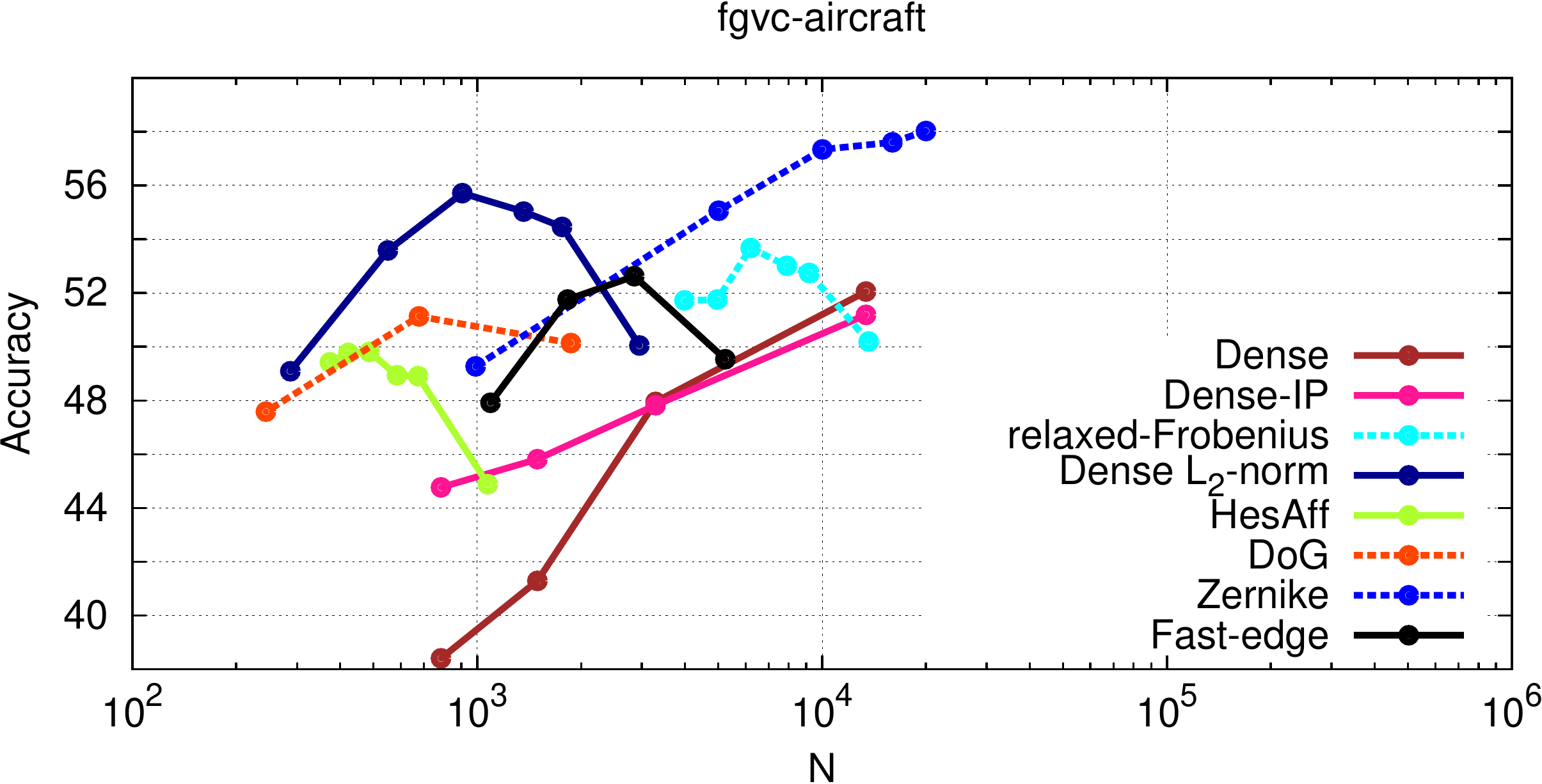}}
        &
        \subfigure{\includegraphics[width=0.475\textwidth]{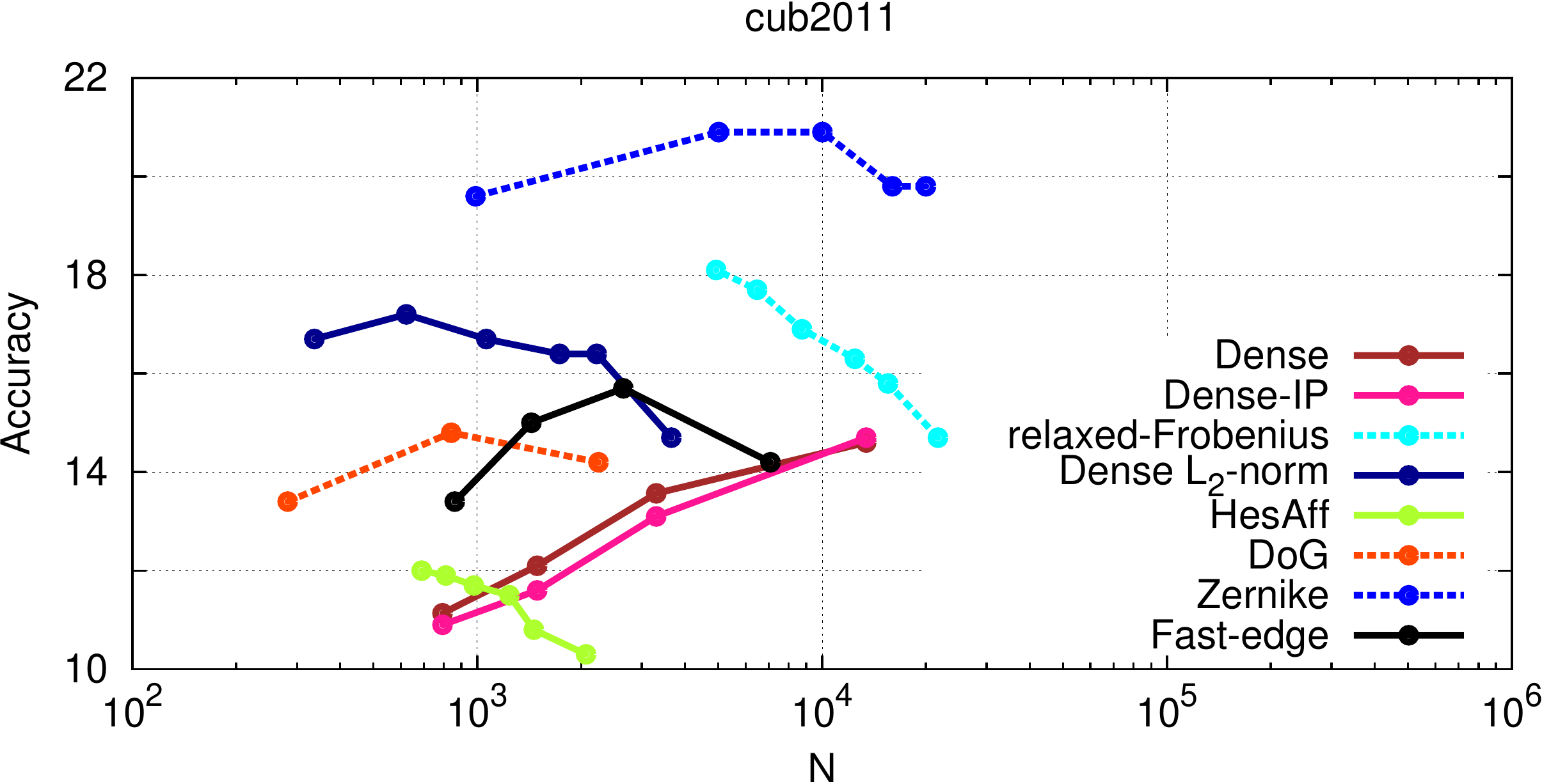}} \\
        
        \subfigure{\includegraphics[width=0.475\textwidth]{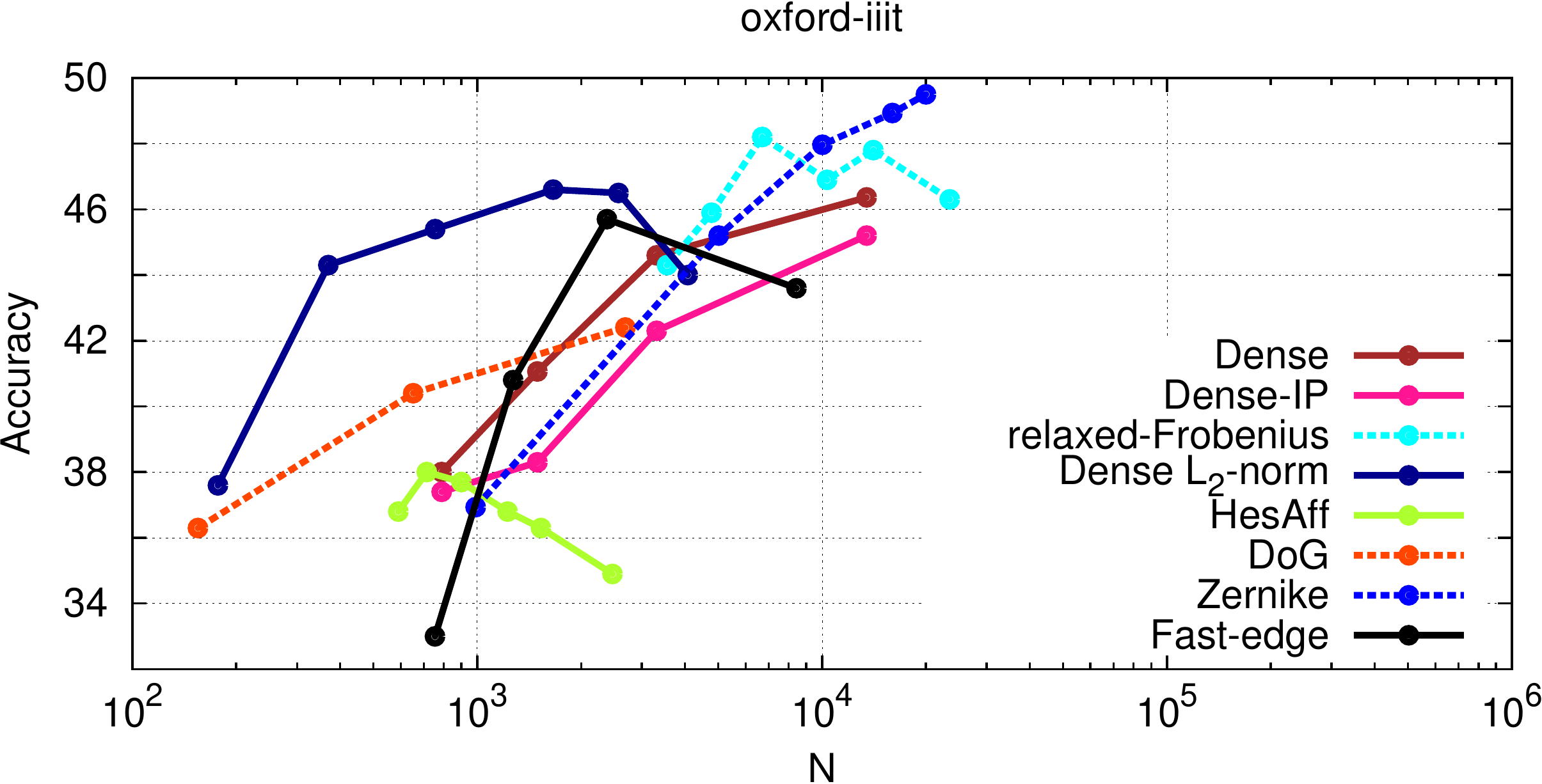}}
        &
        \subfigure{\includegraphics[width=0.475\textwidth]{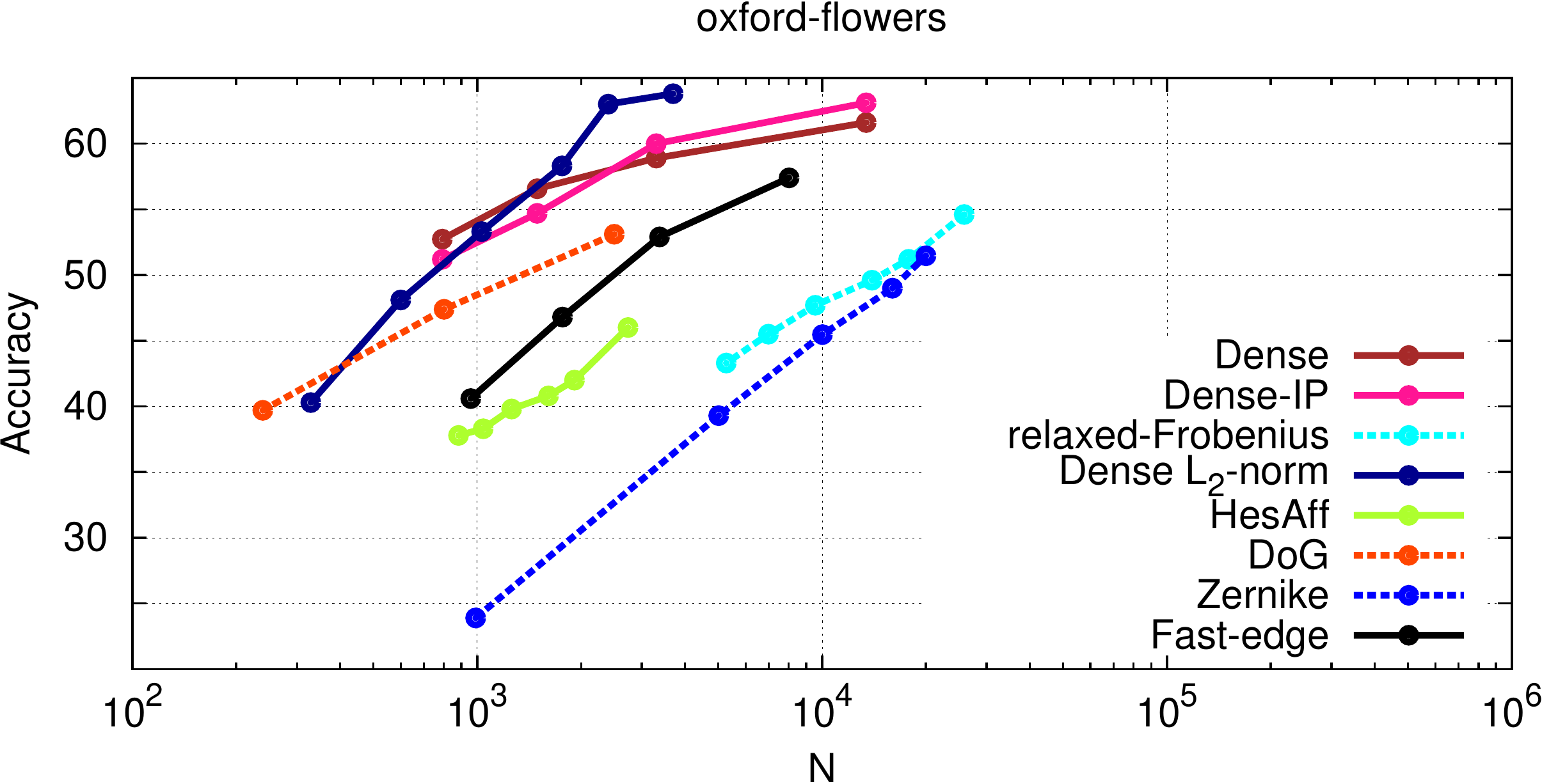}}
        
    \end{tabular}
\vspace{1ex}
\caption{Performance comparison on fine-grained classification versus number of descriptors  per image $N$. The parameter values shown in Table~\ref{tab:numfeatparam} are used.}
\label{fig:classcomp}
\end{figure*}

\vspace{-1ex}

\begin{table}
\begin{center}

\begin{tabular}{|@{\ssp}l@{\ssp}|@{\ssp}l@{\ssp}|@{\ssp}l@{\ssp}|@{\ssp}l@{\ssp}|}
\hline
Detector & param & explanation & values \\
\hline \hline
Dense &\multirow{2}{*}{$\delta_{xy}$} &\multirow{2}{*}{step size} &\multirow{2}{*}{4,8,12,16}\\
Dense-IP~\cite{Tu10} & & & \\
\hline
Dense \l2-norm &\multirow{4}{*}{$\tau$} &\multirow{4}{*}{threshold} & 0,50,100,200 \\
Harris-Laplace~\cite{MiS04} & & & {0,50,100,200,300,400}\\
HesAff~\cite{MTSZMSKG05} & & & {0,50,100,200,300,400}\\
DoG~\cite{L04} & & & 0.1,0.2,0.3 \\
\hline
Zernike & $N$ & number of features & 1k,5k,10k,16k,20k \\
\hline
MSER~\cite{MCMP02} &\multirow{2}{*}{$\Delta$}&\multirow{2}{*}{stable region parameter}&\multirow{2}{*}{3,5,10,15} \\
MSER-edge & &  &  \\
\hline
SSR~\cite{UVGS13} & \multirow{2}{*}{$k$} & \multirow{2}{*}{initial segment size} & \multirow{2}{*}{25,50,100} \\
SSR-edge & & & \\
\hline
Fast-edge~\cite{DZ13} & $\tau$ & edge strength threshold & 0,0.1,0.2,0.3 \\
\hline
\end{tabular}
\vspace{1ex}
\caption{Parameters controlling number of points for each detector and their corresponding values.\label{tab:numfeatparam}}
\end{center}
\end{table}

\subsection{Performance comparison versus number of descriptors}
\label{sec:perf}
A single parameter for each detector controls the number of keypoints per image. These are shown in Table~\ref{tab:numfeatparam}. We evaluate performance for multiple values. This allows a fair comparison of all detectors with respect to the number of descriptors used to describe an image. The descriptor set cardinality is directly related to the complexity of encoding stage.

Results of performance comparison between different detectors versus the average number of features per image $N$ are shown in Figures~\ref{fig:retrcomp} and \ref{fig:classcomp}, for retrieval and classification respectively. Only the best performing variants of the previous experiments are included in this comparison.

On retrieval it appears that the more descriptors the better, while for fine-grained classification this is not always the case. However, such a behavior seems to be consistent among all detectors for a particular dataset. Thus, it seems to be related with the nature of images and objects depicted in them.

\begin{figure}
    \centering
    \begin{tabular}{c c}
    \subfigure{\includegraphics[height=0.31\columnwidth]{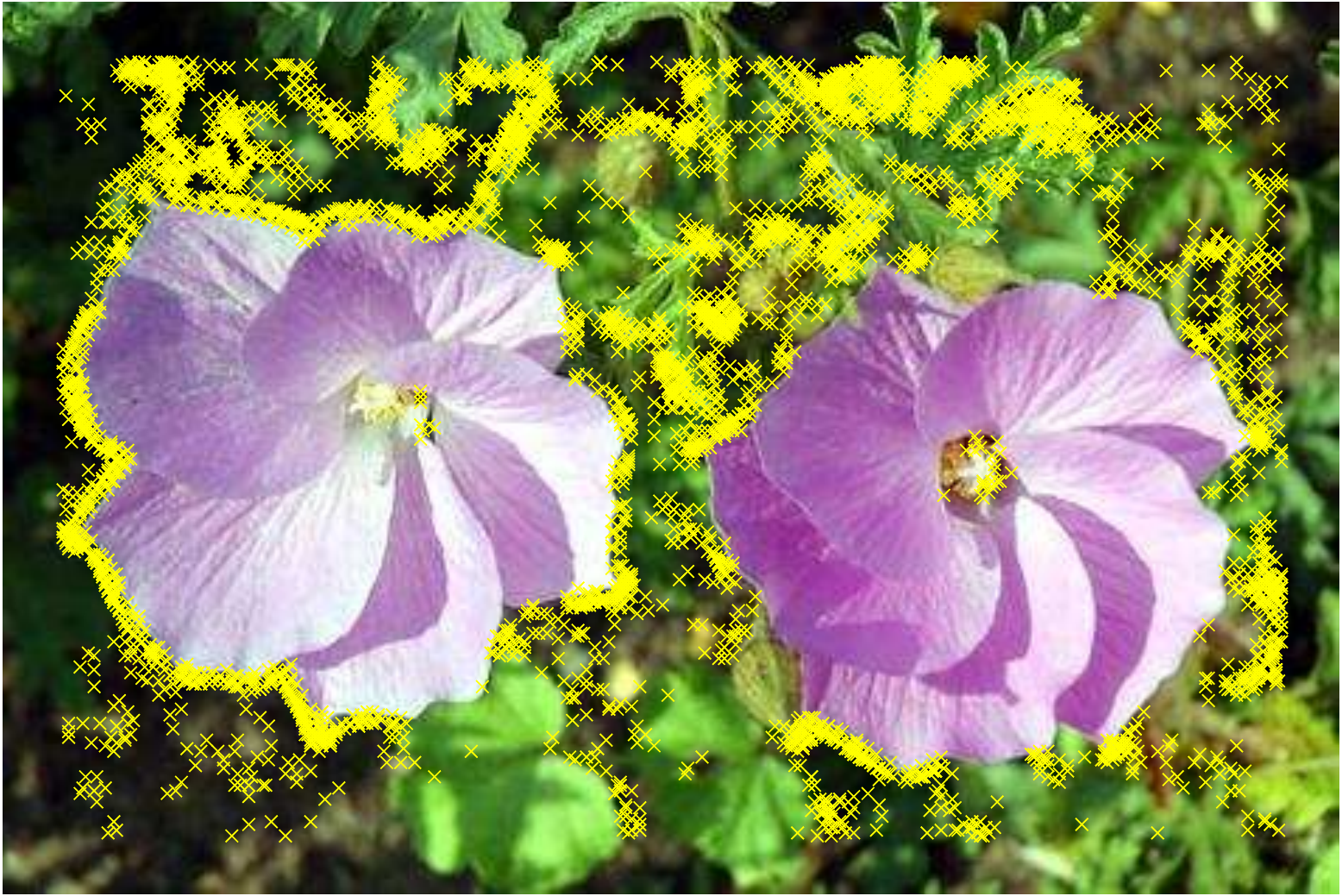}
}
        &
        \subfigure{\includegraphics[height=0.31\columnwidth]{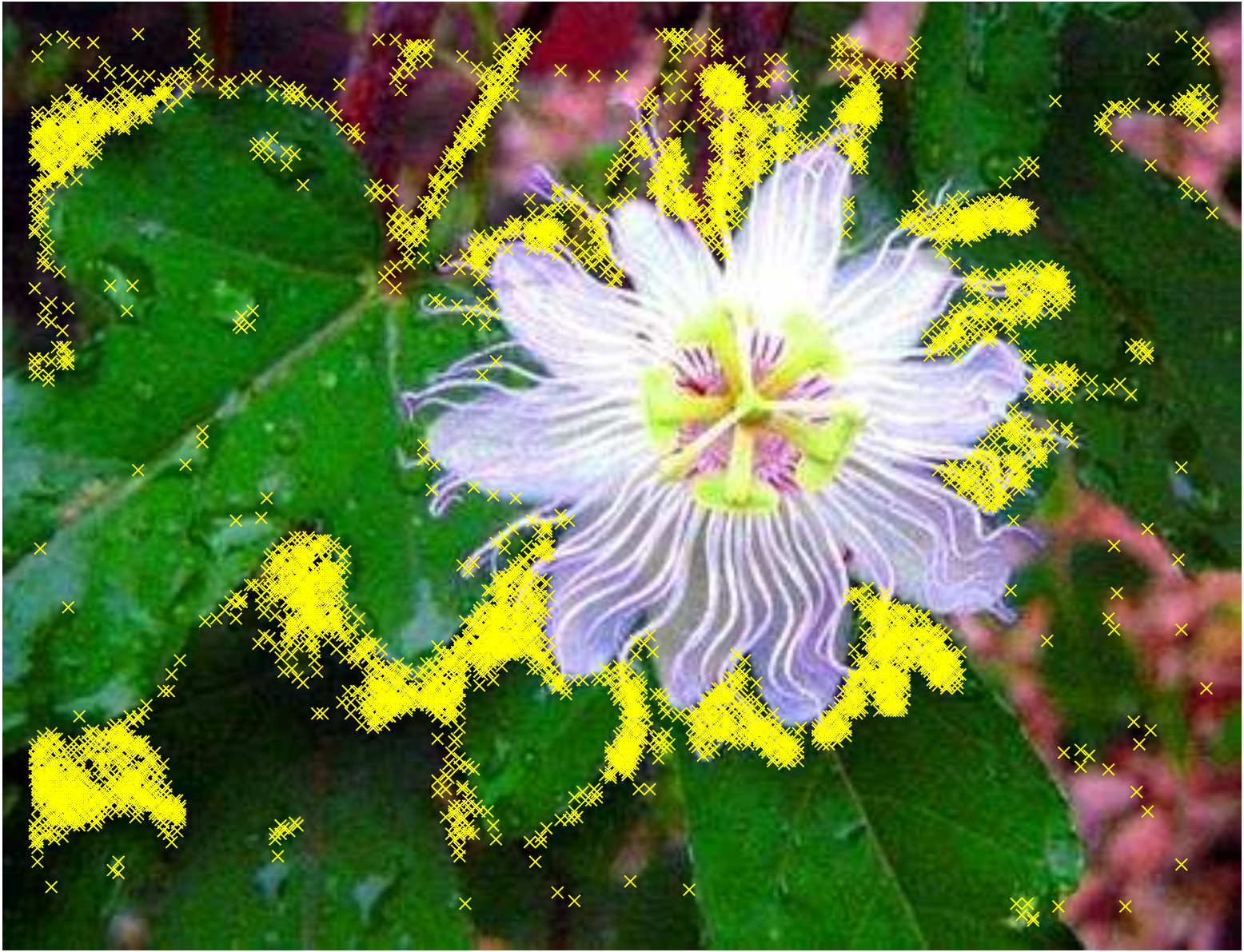}
}
    \end{tabular}
    \vspace{1ex}
\caption{Detection example showing the failure of Zernike detector on images of the Oxford-Flowers dataset.\label{fig:flowersdense}}
\vspace{5ex}
\end{figure}

The relaxed Frobenius detector achieves the best performance on Oxford5k, however Hessian-Affine also performs very well with less number of features, due to its better localization properties. However, we observe that the standard dense and relaxed-Frobenius detectors give the best performance using larger number of features on Holidays. This fact is related to the nature of the dataset; it contains more scenes and not particular objects. On fine-grained classification, the Zernike detector outperforms all other detectors by far, with an exception on Oxford-Flowers dataset. The latter is attributed to many points localized on background texture, as shown in the example of Figure~\ref{fig:flowersdense}.
Interestingly, over both tasks our dense \l2-norm local maxima detector achieves best performance for an intermediate amount of points.

Finally, note that detectors which extract descriptors from patches of 5 constant scales perform better than the detectors that detect local maxima in the scale space. It appears that the former attains enough tolerance to scale changes, although it is not scale invariant.

\subsection{Effect of \l2-norm based filtering}

\begin{figure}
    \centering
		\includegraphics[width=0.95\columnwidth]{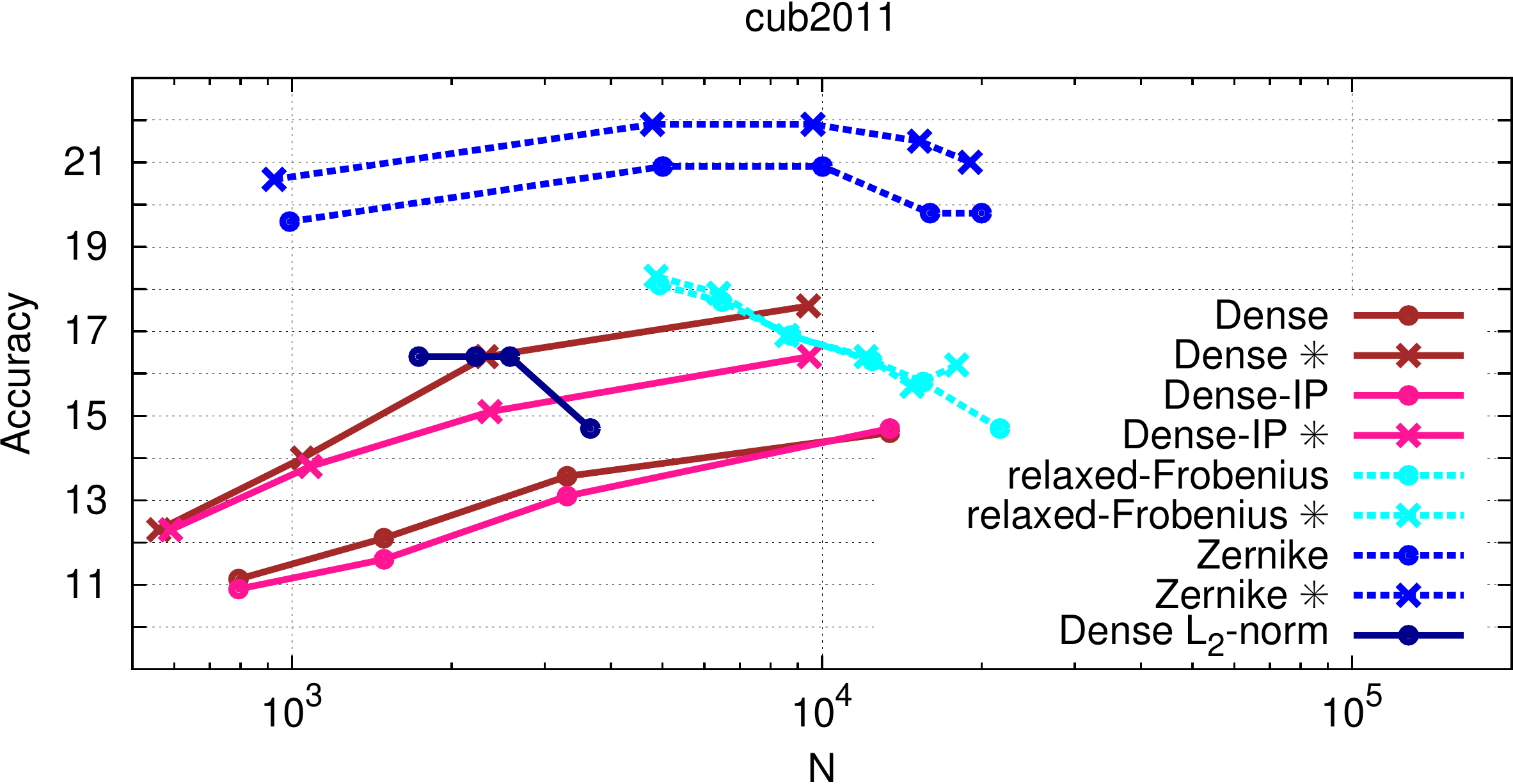}
\caption{Impact of \l2-norm based filtering when applied on top of a variety of detectors. We show performance versus average number of descriptors per image $N$. $\Asterisk$ corresponds to detection with \l2-norm filtering.}
\label{fig:expnormfilter}
\end{figure}

Gosselin \textit{et al.}~\cite{GMJP14} show that the  accuracy on fine-grained classification based on regular dense sampling significantly increases when the SIFT vectors with low \l2-norm are filtered out. We apply the same strategy and show results on Figure~\ref{fig:expnormfilter}. We compute squared \l2-norm to avoid the calculation of square root, and choose $T = 5000$.
As expected, this strategy which aims at removing descriptors extracted from too uniform regions improves performance for regular dense sampling, but the improvement is not significant for detectors that are better localized.

\subsection{State of the art in fine-grained classification}

We compare accuracy of Zernike detector, which is best performing among our contributions, with the state of the art results on fine-grained classification. Fisher vectors are employed once more to encode an image based on Zernike patches. This time, we also apply horizontal mirroring to images, in order to obtain another Fisher vector representation per image. As recently shown, encoding spatial information can boost performance~\cite{GMJP14}. We adopt the choice of spatial coordinate coding (SCC)~\cite{ML12}\cite{KYM13} that allows the use of much larger codebooks, without increase of dimensionality, in contrast to spatial pyramid matching. We choose SCC instead of spatial pyramid matching following the work of Gosselin~\cite{GMJP14}, which shows that SCC performs better than spatial pyramid matching for fine-grained image classification. It allows the use of much larger codebooks. In particular, we append regularized 2D coordinates of patches to RootSIFT descriptors before quantizing them. Regularization weights are learned via cross validation. We have set $N_\mathrm{z}=10$k for the Zernike detector.

Several works focus on a single domain of fine-grained classification, \eg birds, and optimize their methods only for that particular domain. Since this is not the case with our study, we compare with methods that are similar. We present our results in Table~\ref{tab:soaTable}, and compare them with the work of Murray and Perronnin~\cite{MP14}. 
Using the same codebook, with no additional features, we outperform their results in 2 out of 3 datasets. They also employ color descriptors~\cite{CCPR07} which further improves the accuracy. This can complement also for Zernike patches. We also increase the codebook size and combine with SCC that drastically boosts performance.

To our knowledge, there are not any published results for the FGVC-Aircraft dataset yet, except for the initial paper that achieved accuracy equal to 48.69\%~\cite{MKR13}. For the other datasets, it is possible to improve the results using the provided annotations of bounding boxes or object parts. We do not consider such an option. However, the best reported result on CUB2011 is 85.40\%~\cite{BVBP14}, which trains convolutional neural networks (CNN) using such annotations. On the other hand, we are superior to methods which use segmentation information on Oxford-IIIT dataset, such as Wang \etal~\cite{WJY14} that report 59.29\%. In the case of Oxford-Flowers dataset, the highest reported score is 86.6\%~\cite{RASC14}, where the authors also use CNNs without the provided segmentation information.

\begin{table}
\begin{center}
\begin{tabular}{|l|r|*{4}{@{\sssp}c@{\sssp}|}}
   \hline
   Method & $\#C$  & FGVC-Air. & CUB2011 & Ox.-IIIT & Ox.-Flowers \\
   \hline \hline
   Zernike & 256 & 56.7 & 19.3 & 49.6 & 51.1 \\
   Zernike & 4096 & 66.2 & 29.6 & 57.1 & 51.2 \\
   Zernike+SCC & 4096 & 69.9 & 31.9 & 59.5 & 56.1 \\
   GMP~\cite{MP14} & 256 & - & 17.0 & 49.2 & 73.3 \\
   \hline
   GMP+XColor~\cite{MP14} & 256 & - & 33.3 & 56.8 & 84.6 \\
   \hline
\end{tabular}
\end{center}
\caption{Performance comparison with the state of art methods on fine-grained image classification. SCC denotes spatial coordinate coding and $\#C$ is the codebook size. All methods use SIFT descriptors, except for the last one which additionally makes use of color descriptors.\label{tab:soaTable}}
\end{table}

%% file: bios.tex
\vspace{-2 cm.}
\begin{IEEEbiography}[{\includegraphics[width=1in,height=1.25in,clip,keepaspectratio]{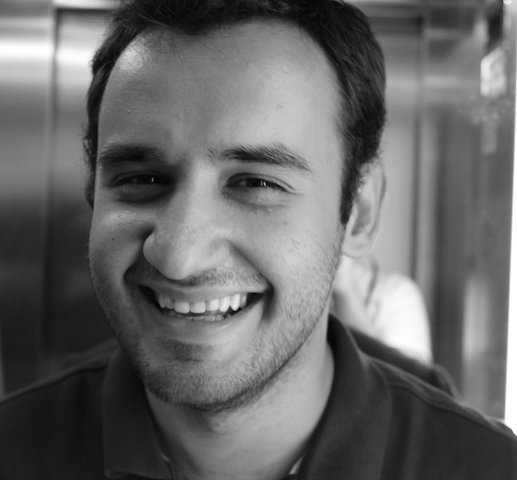}}]{Ahmet~Iscen}
received his B.S. degree from The State University of New York at Binghamton, and M.S from Bilkent University. He is a PhD student at Inria Rennes and University of Rennes I, working on large-scale image retrieval.
\end{IEEEbiography}

\vspace{-2 cm.}

\begin{IEEEbiography}[{\includegraphics[width=1in,height=1.25in,clip,keepaspectratio]{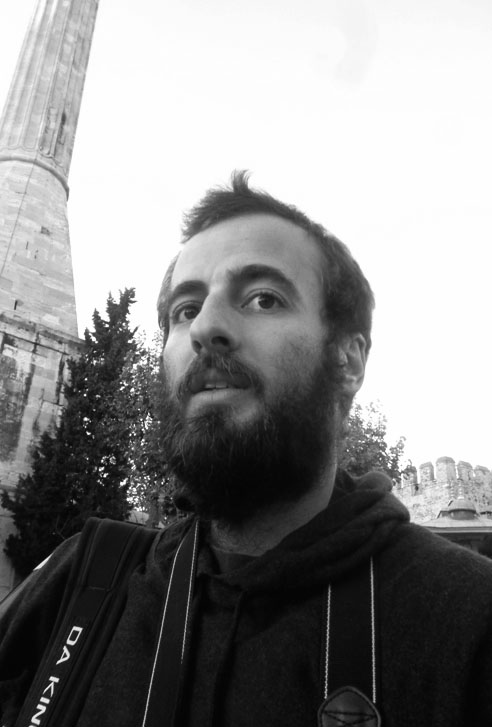}}]{Giorgos~Tolias}
holds a Ph.D. in Electrical and Computer Engineering from the National Technical University of Athens. He is a post-doctoral researcher at Inria Rennes, working mainly in the area of Computer Vision with focus on image search. 
\end{IEEEbiography}

\vspace{-2 cm.}

\begin{IEEEbiography}[{\includegraphics[width=1in,height=1.25in,clip,keepaspectratio]{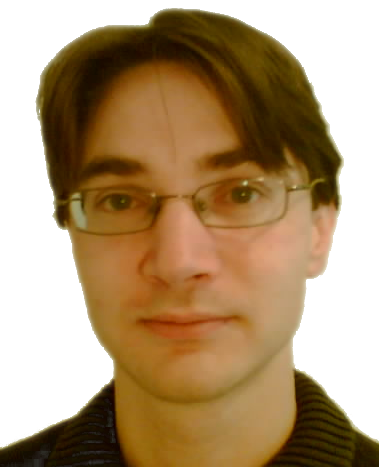}}]{Philippe-Henri~Gosselin}
received the PhD degree in image and signal processing in 2005 (Cergy, France). After 2 years of post-doctoral positions at the LIP6 Lab. (Paris, France) and at the ETIS Lab. (Cergy, France), he joined the MIDI Team in the ETIS Lab as an assistant professor, and then promoted to full professor in 2012. His research focuses on machine learning for online multimedia retrieval. He developed several statistical tools for dealing with the special characteristics of content-based multimedia retrieval. This includes studies on kernel functions on histograms, bags and graphs of features, but also weakly supervised semantic learning methods. He is involved in several international research projects, with applications to image, video and 3D objects databases.
\end{IEEEbiography}

\vspace{-2 cm.}

\begin{IEEEbiography}[{\includegraphics[width=1in,height=1.25in,clip,keepaspectratio]{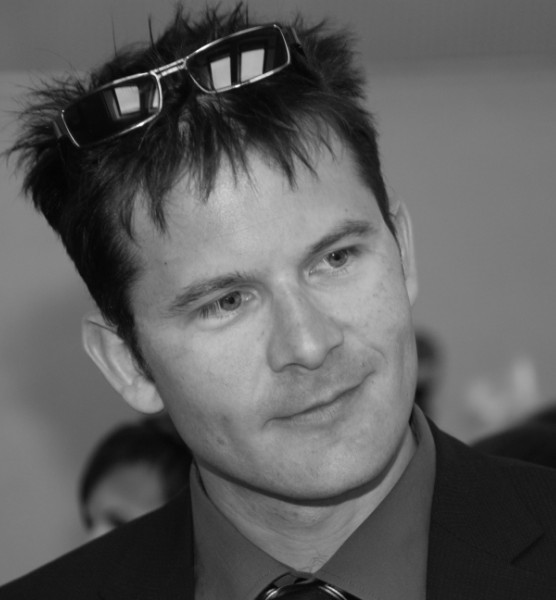}}]{Herv\'e~J\'egou}
holds a Ph.D in Computer Science from the University of Rennes, defended in 2005 on joint source channel coding.
He joined INRIA as a permanent researcher in 2006, in the LEAR team,
and moved to INRIA Rennes in 2009.
His research interests include large scale indexing, essentially for large image and video collections. 
\end{IEEEbiography}